\documentclass{article}

\usepackage[main, final]{neurips_2026}

\usepackage[utf8]{inputenc} 
\usepackage[T1]{fontenc}    
\usepackage{hyperref}       
\usepackage{url}            
\usepackage{booktabs}       
\usepackage{amsfonts}       
\usepackage{nicefrac}       
\usepackage{microtype}      
\usepackage{xcolor}         

\usepackage{graphicx}
\usepackage{amsmath}
\usepackage{amssymb}
\usepackage[utf8]{inputenc} 
\usepackage{colortbl}
\usepackage{amsmath}
\usepackage{makecell}
\usepackage{multirow}
\usepackage{subcaption}
\usepackage{float}
\usepackage{pifont}
\usepackage{arydshln} 

\usepackage{times}
\usepackage{epsfig}
\usepackage{graphicx}
\usepackage{amsmath}
\usepackage{amssymb}
\usepackage{colortbl}
\usepackage{booktabs} 
\usepackage{multirow}
\usepackage{makecell}
\usepackage{float}
\usepackage{enumitem} 
\usepackage{arydshln}
\usepackage{bbding}
\usepackage{animate}
\usepackage{grffile}
\usepackage{siunitx} %
\usepackage[table, usenames, dvipsnames]{xcolor}
\usepackage{tabularx}
\usepackage{bbm}
\usepackage{caption} 
\usepackage{subcaption}
\usepackage{pifont}     %
\usepackage{amsfonts}
\usepackage{tikz}
\usepackage{gensymb}
\usepackage{siunitx}
\usepackage{float}
\usepackage[T1]{fontenc}
\usepackage[scaled=0.9]{inconsolata}
\usepackage{wrapfig}

\newcommand{\project}{\textbf{CLoSeR (Ours)}}

\colorlet{colorFst}{Green!25}       
\colorlet{colorSnd}{SpringGreen!45} 
\colorlet{colorTrd}{Yellow!30}      
\colorlet{colorLow}{darkgray!30}    
\newcommand{\fs}{\cellcolor{colorFst}\bf}   
\newcommand{\nd}{\cellcolor{colorSnd}}      
\newcommand{\rd}{\cellcolor{colorTrd}}      
\newcommand{\gr}[1]{\textcolor{gray!75}{#1}}

\title{CLoSeR: Closing the Loop for Long-Context Streaming Reconstruction}

\author{Moyang Li$^{1 *}$\qquad Zihan Zhu$^{1}$\thanks{Authors contributed equally to this work. Author order is interchangeable.}\qquad Wei Zhang$^{2}$\qquad Marc Pollefeys$^{1,3}$\qquad Daniel Barath$^{1}$\\
$^{1}$ETH Zurich \qquad $^{2}$University of Stuttgart \qquad $^{3}$Microsoft\\
}

\newcommand{\bd}{\mathbf{d}}

\newcommand{\bI}{\mathbf{I}}

\newcommand{\bP}{\mathbf{P}}

\newcommand{\bT}{\mathbf{T}}

\newcommand{\figref}[1]{Fig.~\ref{#1}}
\newcommand{\secref}[1]{Sec.~\ref{#1}}

\newcommand{\tabref}[1]{Table~\ref{#1}}

\makeatletter
\DeclareRobustCommand\onedot{\futurelet\@let@token\@onedot}
\def\@onedot{\ifx\@let@token.\else.\null\fi\xspace}

\makeatother

\newcommand{\boldparagraph}[1]{\vspace{0.2em}\noindent{\bf #1.}}

\renewcommand{\paragraph}[1]{\boldparagraph{#1}}

\definecolor{darkgreen}{rgb}{0,0.7,0}
\definecolor{newyellow}{rgb}{1,0.8,0.05}
\definecolor{newgreen}{rgb}{0.2,0.8,0.2}
\definecolor{brickred}{RGB}{180,60,30}    

\definecolor{lightyellow}{RGB}{255,255,180} 

\definecolor{coral}{RGB}{232,154,142}
\definecolor{grayblue}{RGB}{107,127,158}

\makeatletter
\def\adl@drawiv#1#2#3{%
        \hskip.5\tabcolsep
        \xleaders#3{#2.5\@tempdimb #1{1}#2.5\@tempdimb}%
                #2\z@ plus1fil minus1fil\relax
        \hskip.5\tabcolsep}
\newcommand{\cdashlinelr}[1]{%
  \noalign{\vskip\aboverulesep
           \global\let\@dashdrawstore\adl@draw
           \global\let\adl@draw\adl@drawiv}
  \cdashline{#1}
  \noalign{\global\let\adl@draw\@dashdrawstore
           \vskip\belowrulesep}}
\makeatother

\begin{document}
\maketitle
\begin{center}
    \includegraphics[width=0.99\textwidth]{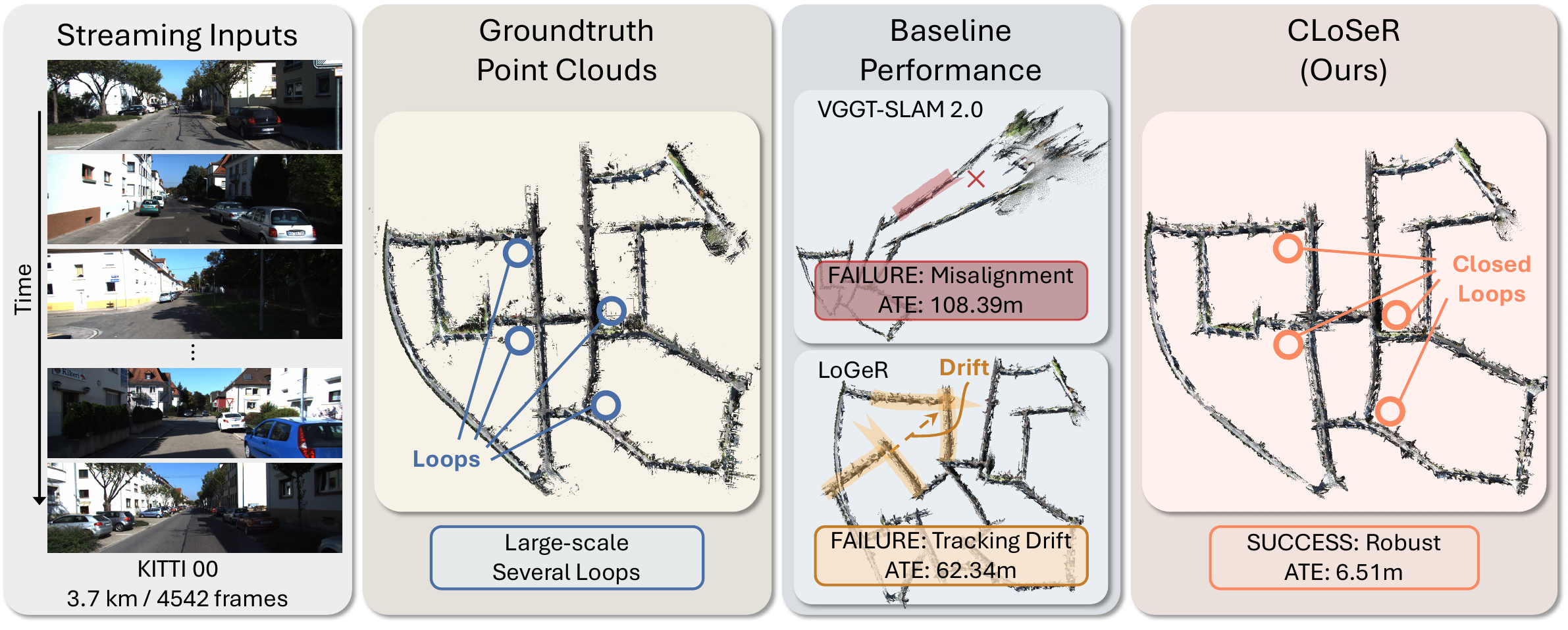}
    \captionof{figure}{\textbf{CLoSeR.} For very long sequences, the submap-based approach VGGT-SLAM 2.0 exhibits erroneous point cloud alignment, while the feedforward method LoGeR suffers from tracking drift due to the absence of explicit constraints when revisiting regions. In contrast, our approach consistently produces accurate reconstructions at kilometer scale.}
    \label{fig:teaser}
\end{center}

\begin{abstract}
\label{sec:abs}
Feedforward foundation models have recently shown remarkable 3D reconstruction capabilities. However, existing models exhibit large tracking drift in long-context streaming reconstruction due to error accumulation. In this paper, we revisit loop closure with streaming reconstruction foundation models to enable accurate, drift-free, kilometer-scale reconstruction. Specifically, our method detects loop candidates through global descriptor retrieval, and constructs loop-conditioned windows to estimate the relative poses between looped frames. Given the observation that our adopted streaming reconstruction backbone produces a globally consistent scale, we optimize all frame poses on the SE(3) manifold with sequential and loop closure constraints, avoiding the pose graph optimization on the Sim(3) or higher-dimensional SL(4) manifolds employed in prior works. Extensive experiments show that our method reduces drift and produces consistent geometry on kilometer-scale sequences, significantly outperforming the state of the art. Code is available at \href{https://github.com/MoyangLi00/CLoSeR.git}{https://github.com/MoyangLi00/CLoSeR.git}.
\end{abstract}

\vspace{-2mm}
\section{Introduction}
\label{sec:intro}
\vspace{-2mm}
Accurate geometry and pose estimation from a sequence of images are fundamental yet challenging tasks in 3D vision, with broad applications in robotics~\citep{liu2024slideslam, arm2023scientific, zhou2022swarm} and autonomous driving~\citep{badue2021self, geiger2013kitti}. Traditional approaches address this problem by leveraging multi-view geometric constraints, ranging from feature-based pipelines~\citep{Mur2015TRO, Mur2017orb2, campos2021orbslam3} that triangulate sparse keypoints across frames, to direct methods~\citep{Engel2017DSO, engel2014lsd} that minimize photometric residuals over image intensities. To achieve more robust estimation under challenging conditions, additional sensors such as inertial measurement units (IMU)~\citep{qin2018vins-mono, leutenegger2022okvis2} and LiDAR~\citep{zhang2014loam, shan2020liosam, xu2021fastlio} are commonly fused into the pipeline. Moreover, some systems have extensively studied loop closure detection~\citep{galvez2012dbow, arandjelovic2016netvlad} and pose graph optimization~\citep{kummerle2011g2o, kaess2012isam2}, to correct the tracking drift over long trajectories and large-scale scenes.

Recently, feedforward foundation models have shifted the paradigm in 3D vision. Early works~\citep{wang2024dust3r, leroy2024mast3r, chen2025easi3r} estimate dense two-view geometry and camera poses with transformers, and rely on pairwise alignment to handle multiple frames. VGGT~\citep{wang2025vggt} extends this paradigm to multi-frame inputs, jointly predicting camera poses, depth maps, dense point clouds, and camera intrinsics in a single forward pass. However, the $O(n^2)$ computational complexity of self-attention layers and its non-causal architecture prevent VGGT from scaling to long sequences or supporting streaming inference.

Several approaches extend feedforward backbones to long-sequence and sequential reconstruction. Some streaming approaches adopt online updates of global memory representation~\citep{wang2025cut3r, chen2025ttt3r} or Test-Time Training (TTT) layers \citep{zhuo2025streamvggt, elflein2026vggt3, jin2026zipmap, zhang2026loger} to propagate information across windows to enable long-context reconstruction. However, they still suffer from tracking drift over long trajectories due to the lack of explicit constraints imposed on revisited regions. Submap-based approaches~\citep{maggio2025vggt-slam, maggio2026vggt-slam2, deng2025vggt-long} instead partition the sequence into chunks, independently infer the geometry and camera poses of each submap, and align them through point clouds registration. These approaches further incorporate loop closure to mitigate drift. However, since each submap is produced by an independent forward pass of VGGT, scale inconsistency inevitably arises across submaps. As a consequence, the resulting pose graph has to be optimized on the Sim(3)~\citep{deng2025vggt-long} or even higher-dimensional SL(4)~\citep{maggio2025vggt-slam, maggio2026vggt-slam2} manifold to absorb scale or projective ambiguity. However, the relative poses and scales recovered through point cloud registration remain noisy and unreliable, as shown in \figref{fig:teaser}, which substantially limits their performance in real-world complex scenarios.

This raises a natural question: \emph{can loop closure be made simpler and more reliable when built upon a streaming backbone whose global scale is already consistent?} 
In this paper, we present \textbf{CLoSeR}, a streaming reconstruction system that augments a prior streaming backbone LoGeR~\citep{zhang2026loger} with loop closure for accurate kilometer-scale reconstruction from uncalibrated monocular videos. 
Our approach exploits two key properties of LoGeR: (a) its pose predictions across windows share a globally consistent scale, rendering pose graph optimization on the SE(3) manifold sufficient and avoiding the Sim(3) or SL(4) manifolds used in prior submap-based methods; (b) its window-wise inference does not require temporal contiguity of input frames, allowing us to append a loop-conditioned window comprising the current and revisited frames after the regular streaming windows, and to directly recover their relative poses in the global coordinate frame.

In summary, our main contributions are as follows:
\begin{itemize}[nosep]
    \item We present a system that integrates loop closure into streaming feedforward reconstruction models, enabling accurate kilometer-scale reconstruction from monocular videos.
    \item We propose a simple yet effective loop-conditioned window mechanism that exploits the streaming nature of the backbone to recover relative poses between looped frames in a globally consistent coordinate frame.
    \item We show that, owing to the globally consistent scale, pose graph optimization on the SE(3) manifold is sufficient for accurate large-scale reconstruction, obviating the Sim(3) or SL(4) optimization and the point-cloud alignment adopted by prior submap-based methods.
\end{itemize}

\vspace{-2mm}
\section{Related Works}
\label{sec:related_works}
\vspace{-2mm}

\textbf{Structure-from-Motion (SfM)} aims to estimate sparse point clouds and camera poses from a collection of images.
Traditional SfM pipelines~\citep{snavely2008modeling, Colmap2016, pan2024glomap} typically decompose this problem into several stages: keypoint detection and correspondence matching, camera pose estimation, and joint refinement of structure and poses.
To improve the robustness and accuracy of SfM, recent works incorporate deep learning at different stages, including deep feature extraction~\citep{detone2018superpoint, dusmanu2019d2, yi2016lift} and learning-based feature matching~\citep{sarlin2020superglue, edstedt2024roma, pautrat2023gluestick, lindenberger2023lightglue}.
Beyond this, another line of work explores end-to-end differentiable SfM~\citep{teed2018deepv2d, wei2020deepsfm, brachmann2024scene, wang2024vggsfm}, which jointly optimizes geometry and camera poses within a unified learning framework.

\textbf{Simultaneous Localization and Mapping (SLAM)} aims to estimate camera poses and reconstruct the scene online from sequential observations.
Traditional approaches are commonly categorized into feature-based methods~\citep{Mur2015TRO, Mur2017orb2, campos2021orbslam3} and direct methods~\citep{Engel2017DSO, engel2014lsd, newcombe2011dtam}.
Feature-based methods track sparse keypoints across frames and triangulate them into a sparse map, while direct methods minimize photometric residuals over image intensities to jointly recover poses and geometry.
To improve robustness under challenging conditions like motion blur, low texture and large-scale scenes, some pipelines introduce additional sensors (IMU~\citep{qin2018vins-mono, leutenegger2022okvis2, ZHU2025VIGS-SLAM} and LiDAR~\citep{zhang2014loam, shan2020liosam, xu2021fastlio}).
To mitigate drift over long trajectories, many systems further incorporate loop closure~\citep{Mur2015TRO, Mur2017orb2, campos2021orbslam3, engel2014lsd, leutenegger2022okvis2}.
More recently, learning-based visual SLAM and VO systems~\citep{teed2021droid, teed2023dpvo, lipson2024dpv} have demonstrated superior accuracy and robustness over their classical counterparts.
In parallel, another line of work integrates neural scene representations such as NeRF~\citep{Zhu2022CVPR, zhu2024nicer} and 3D Gaussian Splatting~\citep{yan2024gs-slam, keetha2024splatam, matsuki2024monogs} into SLAM, enabling photorealistic dense mapping. Despite maintaining consistency over long sequences, these carefully designed SLAM systems typically rely on iterative optimization and sophisticated pipelines. 
In contrast, recent feedforward reconstruction models jointly predict scene geometry and camera poses in a single forward pass, eliminating the need for iterative optimization.

\textbf{Feedforward 3D Reconstruction Models} predict dense 3D geometry directly from images without requiring known camera parameters.
DUSt3R~\citep{wang2024dust3r} and MASt3R~\citep{leroy2024mast3r} predict dense point clouds from an image pair within a common coordinate frame.
MonST3R~\citep{zhang2024monst3r} and Easi3R~\citep{chen2025easi3r} further extend DUSt3R to dynamic environments.
VGGT~\citep{wang2025vggt} and ${\pi^3}$~\citep{wang2025pi3} support arbitrary numbers of input frames and achieve more accurate predictions.
However, their quadratic attention limits scalability to very long sequences and prohibits online streaming inference.
To address this limitation, VGGT-SLAM~\citep{maggio2025vggt-slam, maggio2026vggt-slam2} and VGGT-Long~\citep{deng2025vggt-long} extend VGGT to long sequences through submap-based alignment.
In contrast, CUT3R~\citep{wang2025cut3r} and TTT3R~\citep{chen2025ttt3r} handle long sequences by continuously updating a global state representation in an online manner.
Several other streaming reconstruction works~\citep{elflein2026vggt3, jin2026zipmap, zhang2026loger} instead leverage Test-Time Training (TTT) layers~\citep{sun2024learning, zhang2025lact} to propagate information across windows with linear complexity.
Despite the rapid progress of feedforward reconstruction, existing approaches still struggle with drift accumulation in long-context streaming reconstruction.
In contrast, our method incorporates effective loop closure within the feedforward reconstruction model, enabling accurate, drift-free streaming reconstruction at the kilometer scale.

\vspace{-2mm}
\section{Method}
\label{sec:method}
\vspace{-2mm}

\begin{figure}
    \centering
    \includegraphics[width=1.0\linewidth]{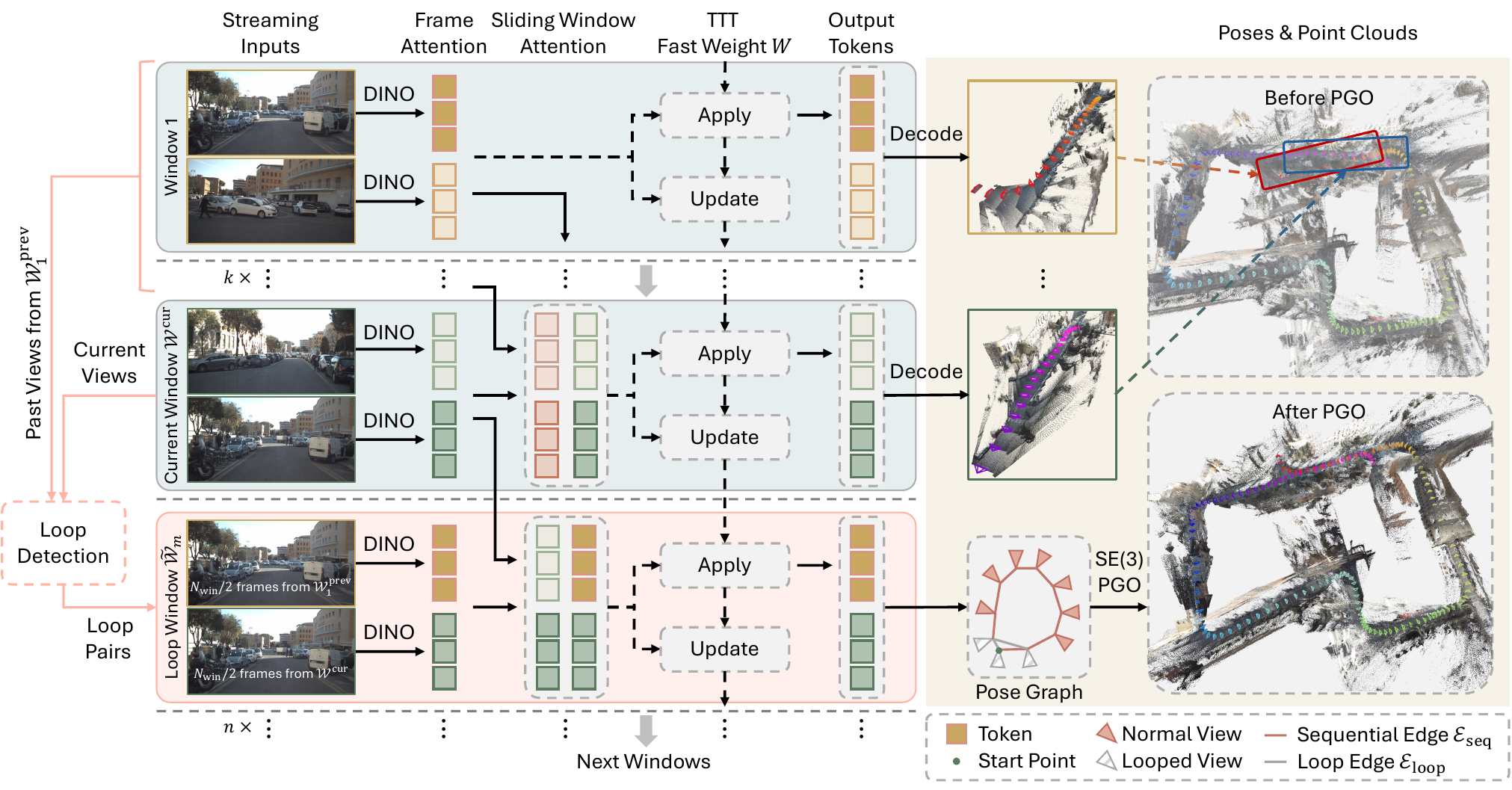}
    \vspace{-2mm}
    \caption{\textbf{System overview.} Given a streaming monocular sequence, our method processes it in a sliding-window manner. For each window, we first tokenize the frames with a DINO-based encoder and then apply a stack of attention layers, comprising frame attention, sliding-window attention, and TTT layers, to maintain global consistency across the sequence. Upon the arrival of each new window, we additionally extract SALAD descriptors to identify potential loop closures against prior windows. When such candidates are detected, we construct a loop-conditioned window by pairing $N_{\text{win}}/2$ frames from the current window with $N_{\text{win}}/2$ frames from the corresponding prior window, predict the relative poses between loop edges from a single forward pass, and perform pose graph optimization on the $\mathrm{SE}(3)$ manifold to correct tracking drift.}
    \label{fig:pipeline}
    \vspace{-4mm}
\end{figure}

Our proposed method, \textbf{CLoSeR}, enables accurate kilometer-scale streaming reconstruction from monocular sequences by augmenting a feedforward backbone~\citep{zhang2026loger} with a loop closure module. In \secref{sec:preliminary}, we introduce the preliminaries of the streaming reconstruction backbone that our framework builds upon. In \secref{sec:loop_detect}, we present loop detection and relative pose estimation between looped frames. Finally, in \secref{sec:pgo}, we describe the global pose graph optimization over all frames with sequential and loop closure constraints. The overview of CLoSeR is shown in \figref{fig:pipeline}.

\vspace{-1mm}
\subsection{Preliminaries}
\label{sec:preliminary}
\vspace{-1mm}

Given a long image sequence $\mathcal{I} = \{\bI_1, \dots, \bI_N\}$ as inputs, LoGeR~\citep{zhang2026loger} partitions it into $K$ overlapping chunks of size $N_{\text{win}}$ with stride $s = N_{\text{win}} - N_{\text{overlap}}$. The $k$-th chunk ($k = 1, \dots, K$) is defined as
\begin{equation}
    \mathcal{W}_k = \{\, \bI_t \mid (k-1)s + 1 \leq t \leq (k-1)s + N_{\text{win}} \,\}.
\end{equation}

During inference, LoGeR \citep{zhang2026loger} first tokenizes the current window of images with a DINO-based \citep{oquab2023dinov2} encoder, and then applies multiple attention layers to produce per-frame camera-to-world poses $\{\bT_t\}_{t \in \mathcal{W}_k} \subset \mathrm{SE}(3)$ in a global world coordinate system and dense point maps $\bP_t \in \mathbb{R}^{H \times W \times 3}$ in local camera coordinates.

To maintain global consistency over long sequences, LoGeR employs two mechanisms for cross-chunk information propagation. Sliding Window Attention (SWA) preserves uncompressed features from recent chunks and enables precise alignment at chunk boundaries, whereas Test-Time Training (TTT) fast weights accumulate an evolving global state that anchors the trajectory to a consistent coordinate frame and mitigates scale drift over long horizons. As a result, all per-frame poses $\{\bT_t\}_{t=1}^{N}$ are expressed in a single global reference frame with consistent scale throughout inference.

Despite these properties, LoGeR still suffers from tracking drift over long trajectories, due to error accumulation caused by the lossy compression of global information in the TTT memory and the absence of explicit revisit constraints in the feedforward inference. This motivates us to build a loop closure framework on top of LoGeR. Two properties of the backbone make it particularly well suited as a frontend for loop closure: its chunk-based forward pass places no requirement on the temporal contiguity of input frames, and its pose outputs across chunks already share a common global scale. Exploiting these properties, we introduce a loop closure module on top of the LoGeR backbone, which estimates relative poses between looped frames under a consistent scale and enforces the resulting constraints through pose graph optimization.

\vspace{-1mm}
\subsection{Loop Detection and Alignment}
\label{sec:loop_detect}
\vspace{-1mm}
To correct the tracking drift caused by error accumulation, we perform loop detection and loop pair alignment in a sequential manner.

\paragraph{Loop Detection}
Let $\bI_j$ denote a frame in the current window $\mathcal{W}^{\text{cur}}$ and $\bI_i$ a frame from previous windows. Upon the arrival of a new window of $N_{\text{win}}$ frames, we extract the compact global descriptor $\bd_j^{\text{cur}}$ for each $\bI_j$ via SALAD~\citep{izquierdo2024salad}, and compute its cosine similarity with the descriptors $\bd_i^{\text{prev}}$ of all frames from previous windows. 
A pair $(\bI_j, \bI_i)$ is considered a valid loop candidate if its similarity exceeds $\tau_{\text{sim}}$ and its temporal separation satisfies $j - i > \tau_{\text{time}}$. For each frame $\bI_j$, we retain its top-$k$ candidates ranked by similarity, yielding the loop pair set
\begin{equation}
    \mathcal{C}_k = \bigcup_{j} \left\{ (\bI_j, \bI_i) \;\middle|\;
    i \in \mathrm{Top}_k\!\left( s(\bd_j^{\text{cur}}, \bd_i^{\text{prev}}) \right),\;
    s(\bd_j^{\text{cur}}, \bd_i^{\text{prev}}) > \tau_{\text{sim}},\;
    j - i > \tau_{\text{time}}
    \right\},
\end{equation}
where $s(\bd_j^{\text{cur}}, \bd_i^{\text{prev}}) = \dfrac{\bd_j^{\text{cur}} \cdot \bd_i^{\text{prev}}}{\|\bd_j^{\text{cur}}\|_2 \, \|\bd_i^{\text{prev}}\|_2}$ denotes the cosine similarity, and $\mathrm{Top}_k(\cdot)$ selects the indices of the $k$ frames with the largest similarity scores.

\paragraph{Loop-Conditioned Window Construction}
The detected loop frames in $\mathcal{C}_k$ may come from several distinct prior windows. Accordingly, the current window can be associated with multiple prior windows, each yielding a separate loop-conditioned window. Formally, let $\{\mathcal{W}^{\text{prev}}_m\}_{m=1}^{M}$ denote the set of prior windows connected to the current window $\mathcal{W}^{\text{cur}}$ through loop pairs. For each $\mathcal{W}^{\text{prev}}_m$, we denote the set of loop pairs linking it to $\mathcal{W}^{\text{cur}}$ as:
\begin{equation}
    \mathcal{L}_m = \{ (\bI_j, \bI_i) \in \mathcal{C}_k \mid \bI_j \in \mathcal{W}^{\text{cur}},\; \bI_i \in \mathcal{W}^{\text{prev}}_m \}.
\end{equation}

We then construct two subsets $\mathcal{S}^{\text{cur}}_m \subseteq \mathcal{W}^{\text{cur}}$ and $\mathcal{S}^{\text{prev}}_m \subseteq \mathcal{W}^{\text{prev}}_m$, each of size $N_{\text{win}}/2$, from the frames that participate in $\mathcal{L}_m$. If more than $N_{\text{win}}/2$ frames are available on either side, we keep the $N_{\text{win}}/2$ frames with the highest similarity scores; if fewer are available, we augment the subset with temporally adjacent frames from the respective window. The loop-conditioned window is then formed as $\tilde{\mathcal{W}}_m = \mathcal{S}^{\text{cur}}_m \cup \mathcal{S}^{\text{prev}}_m$.

This yields a collection $\{\tilde{\mathcal{W}}_m\}_{m=1}^{M}$ of loop-conditioned windows that preserve local temporal continuity while explicitly incorporating long-range loop constraints.

\paragraph{Relative Pose Estimation}
Each loop-conditioned window $\tilde{\mathcal{W}}_m$ is passed through the streaming inference model to obtain the poses. Unlike the standard sliding-window inference in Sec.~\ref{sec:preliminary}, which operates on temporally contiguous frames, a single forward pass here jointly infers poses for frames from $\mathcal{W}^{\text{cur}}$ and $\mathcal{W}^{\text{prev}}_m$. Let $\bT_j^{(m)}$ and $\bT_i^{(m)}$ denote the poses of $\bI_j \in \mathcal{W}^{\text{cur}}$ and $\bI_i \in \mathcal{W}^{\text{prev}}_m$ estimated from $\tilde{\mathcal{W}}_m$. The relative pose constraint for each loop pair $(\bI_j, \bI_i) \in \mathcal{L}_m$ is
\begin{equation}
    \bT_{ji}^{\text{loop}} = \bigl(\bT_j^{(m)}\bigr)^{-1} \, \bT_i^{(m)}.
\end{equation}
Collecting these constraints over all loop pairs and all $\{\tilde{\mathcal{W}}_m\}_{m=1}^{M}$ yields the inter-frame loop constraints used in the pose graph optimization (Sec.~\ref{sec:pgo}).

\vspace{-1mm}
\subsection{Pose Graph Optimization}
\label{sec:pgo}
\vspace{-1mm}
Given sequential constraints between each frame and its $N_{\text{seq}}$ temporal neighbors, and loop closure constraints $\{\bT_{ji}^{\text{loop}}\}$ provided by Sec.~\ref{sec:loop_detect}, we jointly refine all frame poses via pose graph optimization on $\mathrm{SE}(3)$, as LoGeR’s SWA and TTT layers ensure a globally consistent scale. We denote the corresponding edge sets by $\mathcal{E}_{\text{seq}}$ and $\mathcal{E}_{\text{loop}}$, and use $\bT_{qp}^{\text{seq}}$ to represent the relative pose on a sequential edge $(q, p) \in \mathcal{E}_{\text{seq}}$.
Thus, we formulate the following cost function to optimize the global poses $\{\bT_t\}_{t=1}^{N} \subset \mathrm{SE}(3)$:
\begin{eqnarray}
    \{\bT_t^{*}\} = \arg\min_{\{\bT_t\}}
    \sum_{(q, p) \in \mathcal{E}_{\text{seq}}}
    \left\| \text{Log}_{{\text{SE}}(3)}\!\left( (\bT_{qp}^{\text{seq}})^{-1} \, \bT_q^{-1} \, \bT_p \right) \right\|_2^2 \nonumber \\
    + \sum_{(j, i) \in \mathcal{E}_{\text{loop}}}
    \rho_{\delta}\!\left( \left\| \text{Log}_{\mathrm{SE}(3)}\!\left( (\bT_{ji}^{\text{loop}})^{-1} \, \bT_j^{-1} \, \bT_i \right) \right\| \right),
\end{eqnarray}
where $\text{Log}_{\mathrm{SE}(3)}(\cdot): \mathrm{SE}(3) \to \mathbb{R}^6$ maps a transformation to its 6-dimensional tangent vector, and $\rho_{\delta}(\cdot)$ is the Huber loss with threshold $\delta$, applied to the loop term to suppress the influence of outlier loop constraints as follows:
\begin{equation}
    \rho_{\delta}(r) =
    \begin{cases}
        \tfrac{1}{2} r^2, & |r| \leq \delta, \\
        \delta \left( |r| - \tfrac{1}{2}\delta \right), & |r| > \delta.
    \end{cases}
\end{equation}
We solve the problem with the Levenberg-Marquardt algorithm, and update the poses on the manifold via $\bT_t \leftarrow \bT_t \, \exp(\boldsymbol{\xi}_t^{\wedge})$ at each iteration, where $\boldsymbol{\xi}_t \in \mathbb{R}^6$ is the tangent-space increment and $(\cdot)^{\wedge}$ denotes the hat operator that lifts $\mathbb{R}^6$ to $\mathfrak{se}(3)$.

\vspace{-2mm}
\section{Experiments}
\label{sec:exp}
\vspace{-2mm}

\begin{figure}[t]
\vspace{-3mm}
\centering
\setlength{\tabcolsep}{2pt}
\renewcommand{\arraystretch}{1.2}
\begin{tabular}{c@{\hspace{4pt}} m{0.15\textwidth} m{0.40\textwidth} m{0.40\textwidth}}
    \addlinespace[2pt]
    \raisebox{-33pt}{\rotatebox{90}{{\texttt{DROID-W Downtown1}}}} & \includegraphics[width=0.15\textwidth]{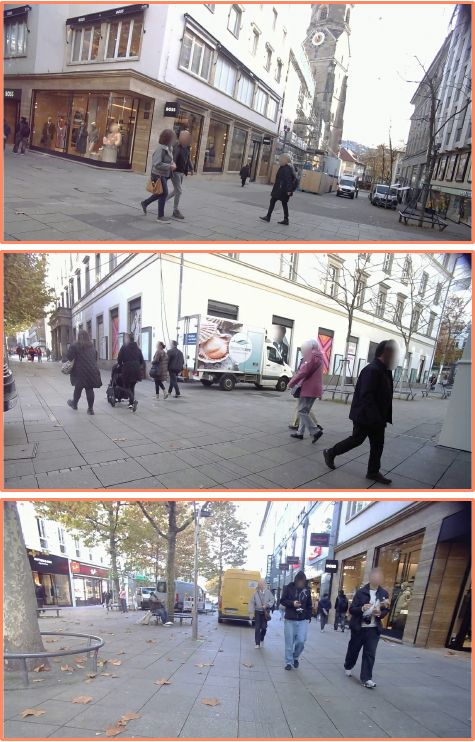} & \includegraphics[width=0.40\textwidth]{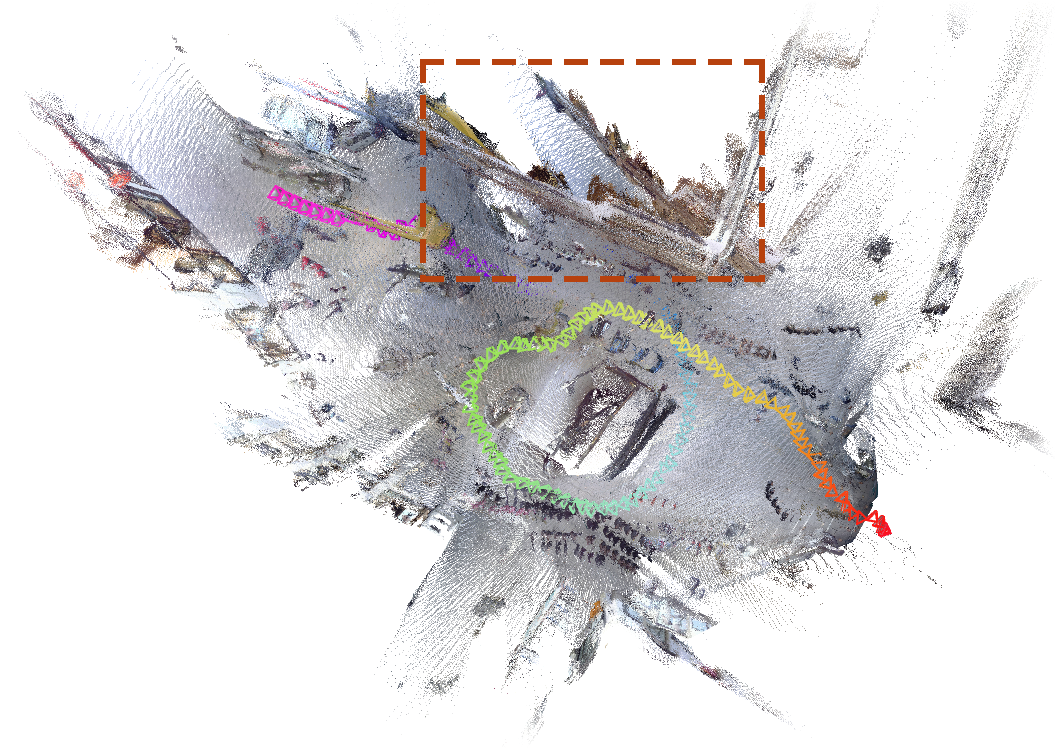} & \includegraphics[width=0.40\textwidth]{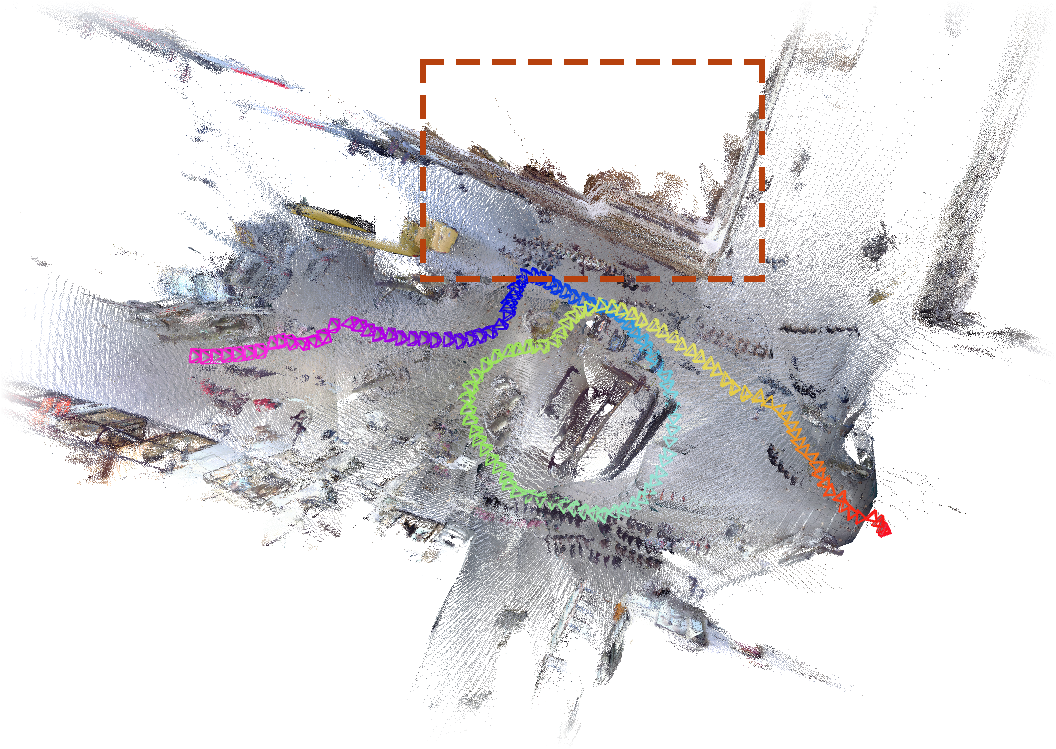} \\
    \raisebox{-27pt}{\rotatebox{90}{\texttt{VBR colosseo\_0}}} & \includegraphics[width=0.15\textwidth]{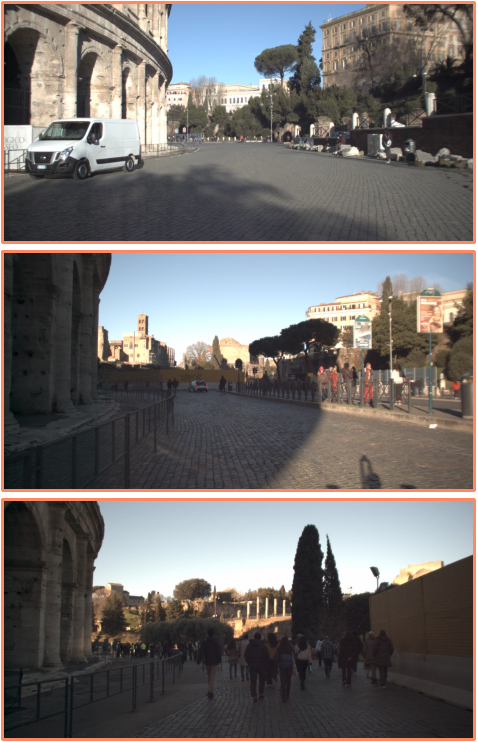} & \includegraphics[width=0.40\textwidth]{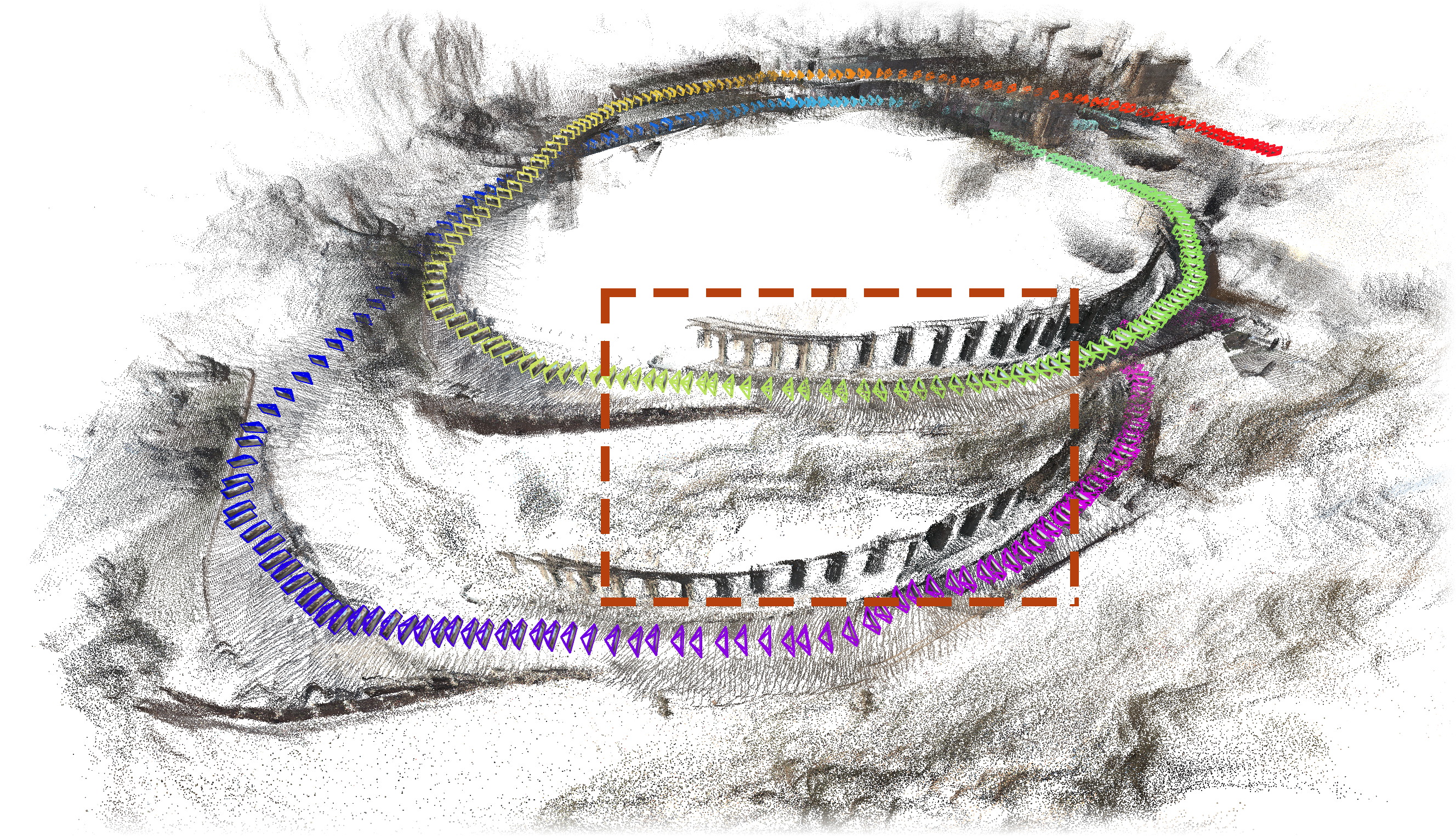} & \includegraphics[width=0.40\textwidth]{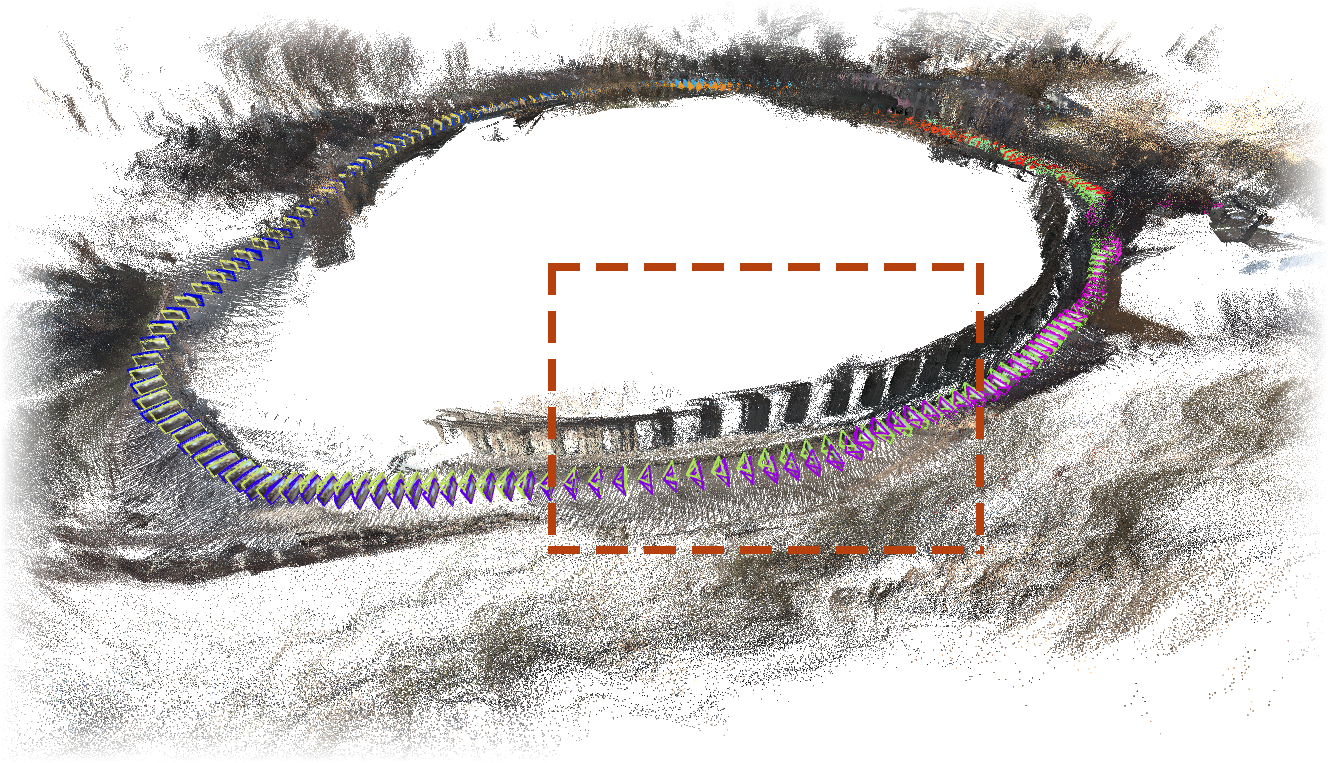} \\
     & \centering Inputs & \centering LoGeR \citep{zhang2026loger} & \centering\arraybackslash \project \\
\end{tabular}
\caption{\textbf{Reconstruction comparison.} LoGeR~\citep{zhang2026loger} fails to close loops and produces misaligned duplicate geometry, whereas our method recovers globally consistent reconstructions. Key regions are highlighted with {\color{brickred}red} dashed boxes.}
\label{fig:point_clouds}
\vspace{-6mm}
\end{figure}

\paragraph{Benchmarks and Metrics}
We evaluate our method on (a) three representative long-context reconstruction benchmarks: the VBR dataset~\citep{brizi2024vbr}, the autonomous driving benchmark KITTI Odometry~\citep{geiger2013kitti}, and the challenging Oxford Spires dataset~\citep{tao2024oxford}; and (b) one outdoor dynamic benchmark, DROID-W~\citep{li2026droidw}.
VBR is captured in Rome and provides RGB sequences, 3D point clouds, IMU measurements, and ground-truth trajectories obtained via bundle adjustment.
Following LoGeR, we evaluate on 7 VBR sequences ranging from 8{,}815 to 18{,}846 frames and 1.45 to 5.2\,km in trajectory length, all containing loops.
KITTI Odometry comprises 11 sequences with long, loop-containing trajectories of up to 4{,}661 frames and 5.1\,km.
Oxford Spires consists of 14 sequences spanning indoor and outdoor scenes with challenging loop closures, ranging from 3{,}961 to 15{,}803 frames and 294 to 925\,m in length.
DROID-W contains 7 challenging dynamic sequences captured outdoors, ranging from 1{,}427 to 2{,}200 frames and 62 to 130\,m in length.
We report Absolute Trajectory Error (ATE) to evaluate tracking accuracy, and align estimated trajectories with the groundtruth via Sim(3) Umeyama alignment~\citep{umeyama1991least}.

\begin{table}[t]
\vspace{-3mm}
\centering
\caption{\textbf{Comparison of absolute trajectory error} (ATE RMSE $\downarrow$ [m]) on VBR \citep{brizi2024vbr}. The top and bottom blocks denote \textit{optimization-based} and \textit{feedforward} methods, respectively. Optimization-based methods are shown in \textcolor{gray!75}{gray}, with the best result per sequence in \textbf{bold}. Among feedforward methods, best results are highlighted as \colorbox{colorFst}{\textbf{first}}, \colorbox{colorSnd}{second}, and \colorbox{colorTrd}{third}.}
\label{tab:vbr}
\resizebox{\textwidth}{!}{
\begin{tabular}{lrrrrrrr!{\smash{\tikz[baseline]{\draw[densely dashed, gray!80, line width=0.8pt] (0pt,-2pt)--(0pt,8pt);}}}r}
\toprule
Method 
& \texttt{colosseo\_0} & \texttt{campus\_0} & \texttt{campus\_1} & \texttt{pincio\_0} & \texttt{spagna\_0} & \texttt{diag\_0} & \texttt{ciampino\_1}
& \textbf{Avg.} \\
\midrule
\textit{\#Frames} & 8815 & 12042 & 11671 & 11142 & 14141 & 10021 & 18846 & 12383 \\
\textit{Scale}    & 1.45km & 2.73km & 2.95km & 1.27km & 1.56km & 1.02km & 5.20km & 2.31km \\
\textit{Contains Loop?} 
& $\checkmark$ & $\checkmark$ & $\checkmark$ & $\checkmark$ & $\checkmark$ & $\checkmark$ & $\checkmark$ & -- \\ 
\midrule
\rowcolor{gray!12}
\multicolumn{9}{c}{\textit{\textbf{Optimization-based (Calibrated)}}} \\
\gr{DROID-SLAM} & \gr{76.53} & \gr{128.77} & \gr{63.89} & \gr{\textbf{43.86}} & \gr{50.63} & \gr{\textbf{35.58}} & \gr{132.84} & \gr{76.01} \\  
\gr{DPV-SLAM} & \gr{107.55} & \gr{56.85} & \gr{88.46} & \gr{102.64} & \gr{\textbf{32.63}} & \gr{74.11} & \gr{60.95} & \gr{74.74} \\
\gr{DPV-SLAM++} & \gr{\textbf{55.47}} & \gr{\textbf{11.90}} & \gr{\textbf{13.11}} & \gr{101.84} & \gr{34.12} & \gr{74.69} & \gr{\textbf{59.47}} & \gr{\textbf{50.09}} \\
\midrule
\rowcolor{gray!12}
\multicolumn{9}{c}{\textit{\textbf{Feedforward (Uncalibrated)}}} \\
VGGT-Long 
& 45.73 & 132.68 & 115.65 & 64.40 & 58.49 & 33.66 & 187.97 & 91.23 \\
VGGT-SLAM 
& 102.92 & 110.77 & 89.69 & 72.98 & 62.67 & 35.62 & 144.02 & 88.38 \\
VGGT-SLAM 2.0
& 102.25 & 26.57 & \rd{22.42} & 47.05 & 59.91 & 36.00 & \rd{51.53} & 49.39 \\
InfiniteVGGT 
& 90.02 & 136.83 & 116.69 & 77.72 & 63.47 & 33.91 & 202.11 & 102.96 \\
CUT3R 
& 88.16 & 46.39 & 48.00 & 52.87 & 47.98 & 30.18 & 187.58 & 71.59 \\
TTT3R 
& 81.63 & 68.36 & 65.67 & 35.56 & 39.03 & \nd{19.84} & 185.59 & 70.81 \\
AMB3R 
& 62.60 & 110.60 & 119.76 & 61.69 & 62.04 & 35.02 & 185.44 & 91.02 \\
$\pi^3$-Chunk 
& 81.63 & 86.54 & 71.49 & 49.85 & 54.27 & 28.07 & 118.81 & 70.09 \\
LingBot-Map 
& \nd{19.22} & \nd{18.46} & \nd{11.40} & 41.89 & 36.56 & \rd{22.65} & 67.64 & \nd{31.12} \\
LoGeR 
& \rd{25.60} & \rd{19.53} & 28.90 & \rd{28.11} & \nd{23.04} & 35.01 & 92.94 & 36.16 \\
LoGeR* 
& 49.38 & 22.44 & 34.90 & \nd{11.08} & \rd{27.05} & 32.88 & \nd{44.51} & \rd{31.75} \\
\project & \fs{14.23} & \fs{7.43} & \fs{3.71} & \fs{9.78} & \fs{13.04} & \fs{12.92} & \fs{16.93} & \fs{11.15} \\
\bottomrule
\end{tabular}}
\vspace{-6mm}
\end{table}
\begin{table}[t]
\centering
\caption{\textbf{Comparison of absolute trajectory error} (ATE RMSE $\downarrow$ [m]) on KITTI \citep{geiger2013kitti}.  Traditional deep visual SLAM systems suffer from scale drift under KITTI's forward-dominant motion. Our approach outperforms all baselines.}
\resizebox{\textwidth}{!}{
\begin{tabular}{lrrrrrrrrrrr!{\smash{\tikz[baseline]{\draw[densely dashed, gray!80, line width=0.8pt] (0pt,-2pt)--(0pt,8pt);}}}r}
\toprule
Method & \texttt{00} & \texttt{01} & \texttt{02} & \texttt{03} & \texttt{04} & \texttt{05} & \texttt{06} & \texttt{07} & \texttt{08} & \texttt{09} & \texttt{10} & \textbf{Avg.} \\
\midrule
\textit{\#Frames} & 4541 & 1101 & 4661 & 801 & 271 & 2761 & 1101 & 1101 & 4071 & 1591 & 1201 & 2109 \\
\textit{Scale}       & 3.7km & 2.5km & 5.1km & 0.6km & 0.4km & 2.2km & 1.2km & 0.7km & 3.2km & 1.7km & 0.9km & 2.0km \\
\textit{Contains Loop?} 
& $\checkmark$ & $\times$ & $\checkmark$ & $\times$ & $\times$ 
& $\checkmark$ & $\checkmark$ & $\checkmark$ & $\times$  & $\checkmark$ & $\times$ & -- \\ 
\midrule
\rowcolor{gray!12}
\multicolumn{13}{c}{\textit{\textbf{Optimization-based (Calibrated)}}} \\
\gr{DROID-SLAM} & \gr{92.10} & \gr{344.60} & \gr{107.61} & \gr{\textbf{2.38}} & \gr{1.00} & \gr{118.50} & \gr{62.47} & \gr{21.78} & \gr{161.60} & \gr{\textbf{72.32}} & \gr{118.70} & \gr{100.28} \\
\gr{DPV-SLAM} & \gr{112.80} & \gr{\textbf{11.50}} & \gr{123.53} & \gr{2.50} & \gr{0.81} & \gr{57.80} & \gr{54.86} & \gr{18.77} & \gr{\textbf{110.49}} & \gr{76.66} & \gr{\textbf{13.65}} & \gr{53.03} \\
\gr{DPV-SLAM++} & \gr{\textbf{8.30}} & \gr{11.86} & \gr{\textbf{39.64}} & \gr{2.50} & \gr{\textbf{0.78}} & \gr{\textbf{5.74}} & \gr{\textbf{11.60}} & \gr{\textbf{1.52}} & \gr{110.90} & \gr{76.70} & \gr{13.70} & \gr{\textbf{25.75}} \\
\midrule
\rowcolor{gray!12}
\multicolumn{13}{c}{\textit{\textbf{Feedforward (Uncalibrated)}}} \\
VGGT-Long & \nd{8.67} & 121.17 & \nd{32.08} & 6.12 & 4.23 & \nd{8.31} & \rd{5.34} & \fs{4.63} & 53.10 & 41.99 & 18.37 & 27.64 \\
VGGT-SLAM 2.0 & 108.39 & 155.00 & 140.15 & 13.45 & 2.87 & \rd{10.12} & 29.17 & 17.27 & 75.33 & 27.70 & 24.09 & 54.87 \\
FastVGGT & OOM & 639.39 & OOM & 21.53 & 9.51 & OOM & 40.56 & 51.35 & OOM & 201.54 & 196.22 & -- \\
InfiniteVGGT & 186.46 & 623.62 & 289.16 & 166.74 & 68.00 & 143.84 & 117.57 & 85.33 & 221.56 & 215.41 & 156.92 & 206.78 \\
CUT3R & 190.38 & 90.59 & 264.39 & 20.40 & 7.31 & 92.25 & 67.54 & 22.48 & 145.08 & 67.42 & 40.00 & 91.62 \\
TTT3R & 119.94 & 99.59 & 238.07 & 16.83 & 3.98 & 36.38 & 47.20 & 11.62 & 107.33 & 86.96 & 33.58 & 72.86 \\
AMB3R & 177.78 & 244.67 & 233.65 & 22.08 & 6.81 & 157.62 & 61.93 & 35.02 & 255.53 & 134.62 & 50.41 & 125.47 \\ 
$\pi^3$-Chunk & 26.65 & 196.04 & 157.92 & 5.13 & \nd{1.09} & 12.79 & 27.66 & 5.94 & 61.26 & 56.31 & 21.96 & 52.07 \\
LingBot-Map & \rd{23.22} & 88.84 & 71.15 & \fs{2.81} & \fs{0.91} & 17.40 & \nd{4.91} & \rd{5.16} & \fs{14.24} & \rd{21.71} & 14.54 & \rd{24.08} \\
LoGeR & 62.34 & \fs{41.64} & 39.64 & \rd{4.89} & \rd{1.82} & 41.27 & 13.99 & 16.24 & \rd{26.46} & 22.71 & \nd{8.84} & 25.44 \\
LoGeR* & 30.47 & \rd{47.91} & \rd{36.32} & 5.38 & 1.95 & 26.34 & 6.60 & 5.55 & \nd{24.41} & \nd{10.12} & \rd{10.11} & \nd{18.65} \\
\project & \fs{6.58} & \nd{41.80} & \fs{28.38} & \nd{4.88} & \rd{1.82} & \fs{6.58} & \fs{4.77} & \nd{4.87} & 26.51 & \fs{5.64} & \fs{8.83} & \fs{12.79} \\
\bottomrule
\label{tab:kitti}
\end{tabular}
}
\vspace{-8mm}
\end{table}
\begin{table}[t]
\vspace{-3mm}
\centering
\caption{\textbf{Comparison of absolute trajectory error} (ATE RMSE $\downarrow$ [m]) on DROID-W \citep{li2026droidw}. 
Optimization-based methods are shown in \textcolor{gray!75}{gray}, with the best result per sequence in \textbf{bold}. Among feedforward methods, best results are highlighted as \colorbox{colorFst}{\textbf{first}}, \colorbox{colorSnd}{second}, and \colorbox{colorTrd}{third}.
Our approach yields robust and accurate camera pose estimation on dynamic sequences, despite not explicitly handling dynamic objects.
}
\setlength{\tabcolsep}{8pt}  
\resizebox{\textwidth}{!}{
\begin{tabular}{lrrrrrrr!{\smash{\tikz[baseline]{\draw[densely dashed, gray!80, line width=0.8pt] (0pt,-2pt)--(0pt,8pt);}}}r}
\toprule
Method & \texttt{downtown1} & \texttt{downtown2} & \texttt{downtown3} & \texttt{downtown4} & \texttt{downtown5} & \texttt{downtown6} & \texttt{downtown7} & \textbf{Avg.} \\
\midrule
\textit{\#Frames} & 1427 & 2200 & 1438 & 1794 & 2157 & 1900 & 1900 & 1831 \\
\textit{Scale}    & 91m & 122m & 62m & 85m & 130m & 105m & 109m & 101m \\
\textit{Contains Loop?} 
& $\checkmark$ & $\checkmark$ & $\checkmark$ & $\checkmark$ &  $\times$ & $\checkmark$ & $\checkmark$ &  -- \\ 
\midrule
\rowcolor{gray!12}
\multicolumn{9}{c}{\textit{\textbf{Optimization-based (Calibrated)}}} \\
\gr{DROID-SLAM} & \gr{0.26} & \gr{7.84} & \gr{1.05} & \gr{0.33} & \gr{0.64} & \gr{\textbf{0.06}} & \gr{\textbf{0.05}} & \gr{1.46} \\
\gr{DPV-SLAM} & \gr{0.41} & \gr{0.50} & \gr{1.05} & \gr{\textbf{0.15}} & \gr{1.05} & \gr{7.55} & \gr{7.95} & \gr{2.67} \\       
\gr{DPV-SLAM++} & \gr{0.41} & \gr{\textbf{0.12}} & \gr{\textbf{0.15}} & \gr{0.16} & \gr{1.06} & \gr{2.57} & \gr{7.04} & \gr{1.64} \\
\gr{WildGS-SLAM} & \gr{\textbf{0.10}} & \gr{0.95} & \gr{0.43} & \gr{0.36} & \gr{0.87} & \gr{1.22} & \gr{0.53} & \gr{0.63} \\
\gr{DROID-W} & \gr{0.15} & \gr{0.25} & \gr{\textbf{0.15}} & \gr{0.32} & \gr{\textbf{0.24}} & \gr{0.43} & \gr{0.07} & \gr{\textbf{0.23}} \\
\midrule
\rowcolor{gray!12}
\multicolumn{9}{c}{\textit{\textbf{Feedforward (Uncalibrated)}}} \\
VGGT-Long & 4.68 & 4.94 & 2.48 & 5.11 & 4.61 & 3.14 & 1.79 & 3.82\\
VGGT-SLAM 2.0 & \nd{0.47} & 2.58 & 2.03 & 4.36 & 4.82 & \fs{0.42} & 1.30 & 2.28 \\
CUT3R & 10.04 & 11.91 & 6.54 & 9.70 & 21.78 & 7.74 & 5.85 & 10.51 \\ 
TTT3R & 4.64 & 11.25 & 4.30 & 7.35 & 11.28 & 5.09 & 7.26 & 7.31 \\
AMB3R & 1.90 & 5.72 & 5.57 & 4.99 & 7.65 & 7.69 & 6.59 & 5.73 \\
LingBot-Map & \rd{1.00} & \fs{0.70} & \fs{0.46} & \fs{0.74} & \rd{1.37} & \rd{1.63} & \nd{0.45} & \nd{0.91} \\
LoGeR & 2.16 & \rd{1.98} & \rd{1.02} & \rd{1.78} & \nd{0.74} & 1.64 & \rd{0.73} & \rd{1.44} \\
\project & \fs{0.34} & \nd{0.88} & \nd{0.99} & \nd{1.52} & \fs{0.73} & \nd{0.50} & \fs{0.30} & \fs{0.75} \\
\bottomrule
\label{tab:droidw}
\end{tabular}
}
\vspace{-8mm}
\end{table}

\begin{table}[t]
\centering
\caption{\textbf{Comparison of absolute trajectory error} (ATE RMSE $\downarrow$ [m]) on Oxford Spires \citep{tao2024oxford}. 
Our approach achieves the best pose estimation performance on this challenging dataset, demonstrating the superiority and robustness of our method.
}
\resizebox{\textwidth}{!}{
\begin{tabular}{lrrrrrrrrrrrrrr!{\smash{\tikz[baseline]{\draw[densely dashed, gray!80, line width=0.8pt] (0pt,-2pt)--(0pt,8pt);}}}r}
\toprule
Method & \texttt{ke-02} & \texttt{ke-03} & \texttt{ke-04} & \texttt{ke-05} & \texttt{ob-01} & \texttt{ob-02} & \texttt{bl-01} & \texttt{bl-02} & \texttt{bl-05} & \texttt{ch-01} & \texttt{ch-02} & \texttt{ch-03} & \texttt{ch-05} & \texttt{bo-02} & \textbf{Avg.} \\
\midrule
\textit{\#Frames}       & 5969 & 5682 & 13518 & 11544 & 5746 & 5469 & 8041 & 7276 & 6755 & 15803 & 11775 & 6204 & 3961 & 4966 & 8051 \\
\textit{Scale}          & 294m & 283m & 783m  & 706m  & 397m & 393m & 494m & 393m & 386m & 925m  & 642m  & 339m & 823m & 691m & 539m \\
\textit{Contains Loop?} & $\checkmark$ & $\checkmark$ & $\checkmark$ & $\checkmark$ & $\checkmark$ & $\checkmark$ & $\checkmark$ & $\checkmark$ & $\checkmark$ & $\checkmark$ & $\checkmark$ & $\checkmark$ & $\checkmark$ & $\times$ & -- \\
\midrule
\rowcolor{gray!12}
\multicolumn{16}{c}{\textit{\textbf{Optimization-based (Calibrated)}}} \\
\gr{DROID-SLAM} & \gr{6.66} & \gr{29.71} & \gr{\textbf{14.10}} & \gr{36.60} & \gr{\textbf{0.17}} & \gr{\textbf{0.36}} & \gr{47.79} & \gr{6.05} & \gr{15.88} & \gr{13.85} & \gr{10.17} & \gr{0.33} & \gr{\textbf{8.92}} & \gr{5.64} & \gr{14.02} \\
\gr{DPV-SLAM}   & \gr{5.84} & \gr{\textbf{4.53}} & \gr{19.74} & \gr{25.25} & \gr{0.27} & \gr{6.24} & \gr{\textbf{3.67}} & \gr{\textbf{0.89}} & \gr{\textbf{11.62}} & \gr{\textbf{1.05}} & \gr{\textbf{1.15}} & \gr{0.94} & \gr{42.62} & \gr{3.51} & \gr{\textbf{9.10}} \\
\gr{DPV-SLAM++} & \gr{\textbf{1.33}} & \gr{29.35} & \gr{18.39} & \gr{\textbf{3.66}} & \gr{0.20} & \gr{2.94} & \gr{3.88} & \gr{1.00} & \gr{12.06} & \gr{3.93} & \gr{7.82} & \gr{\textbf{0.28}} & \gr{40.19} & \gr{\textbf{3.32}} & \gr{9.17} \\
\midrule
\rowcolor{gray!12}
\multicolumn{16}{c}{\textit{\textbf{Feedforward (Uncalibrated)}}} \\
VGGT-Long      & \rd{3.63} & 14.00 & 32.66 & \fs{4.64} & 3.80 & 5.64 & \rd{11.92} & 14.04 & 11.41 & 17.59 & 9.03 & 4.02 & \rd{14.51} & 16.76 & 11.69 \\
VGGT-SLAM 2.0  & 7.28 & 20.57 & 6.90 & 17.06 & \fs{1.76} & \fs{1.85} & 18.74 & \fs{1.23} & \fs{1.26} & \fs{1.50} & \fs{1.64} & \fs{0.96} & 16.41 & \fs{8.51} & \rd{7.55} \\
CUT3R          & 25.44 & 17.75 & 42.51 & 36.12 & 26.83 & 27.05 & 50.64 & 42.53 & 49.45 & 33.46 & 41.67 & 11.83 & 32.47 & 56.71 & 35.32 \\
TTT3R          & 29.05 & 17.45 & 32.54 & 41.37 & 26.37 & 26.65 & 51.27 & 45.89 & 35.25 & 31.17 & 41.46 & 14.18 & 27.26 & 69.42 & 34.95 \\
AMB3R          & 15.95 & 29.38 & 24.68 & 40.05 & \rd{2.99} & \rd{3.36} & 47.88 & 41.59 & 42.97 & 28.91 & 43.34 & 14.56 & 26.64 & 18.79 & 27.22 \\
LingBot-Map    & \nd{3.42} & \rd{3.27} & \fs{3.00} & 9.39 & \nd{2.81} & \nd{2.50} & 18.95 & \nd{3.82} & \nd{2.78} & \rd{3.32} & 13.39 & 2.94 & 28.79 & \nd{9.27} & 7.69 \\
LoGeR          & 4.01 & \nd{3.11} & \rd{6.51} & \rd{5.76} & 7.01 & 6.98 & \fs{6.58} & 7.33 & 6.98 & 9.18 & \rd{6.72} & \rd{1.84} & \nd{10.22} & 14.09 & \nd{6.88} \\
\project       & \fs{2.75} & \fs{2.40} & \nd{5.79} & \nd{4.65} & 4.33 & 4.03 & \nd{7.78} & \rd{5.18} & \rd{5.06} & \nd{2.57} & \nd{5.70} & \nd{1.19} & \fs{8.03} & \rd{14.08} & \fs{5.26} \\
\bottomrule
\label{tab:oxford}
\end{tabular}
}
\vspace{-8mm}
\end{table}
\begin{figure}[t]
\vspace{-4mm}
\centering
\setlength{\tabcolsep}{2pt}
\renewcommand{\arraystretch}{1.0}
\begin{tabular}{@{}c@{\hspace{2pt}}*{5}{>{\centering\arraybackslash}m{0.19\linewidth}}@{}}
 & \texttt{VGGT-Long} & \texttt{VGGT-SLAM 2.0} & \texttt{LingBot-Map} & \texttt{LoGeR} & \textbf{\texttt{\project}} \\
\raisebox{-10pt}{\rotatebox{90}{\texttt{KITTI 09}}} & \includegraphics[width=\linewidth]{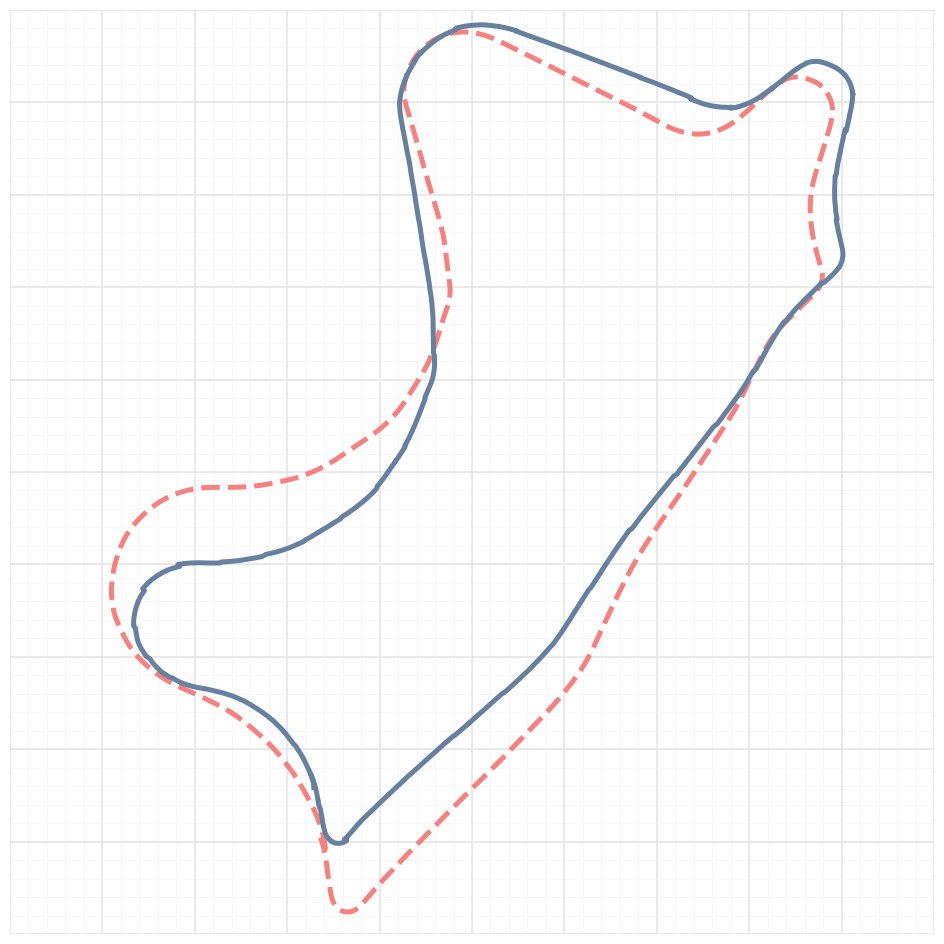} & \includegraphics[width=\linewidth]{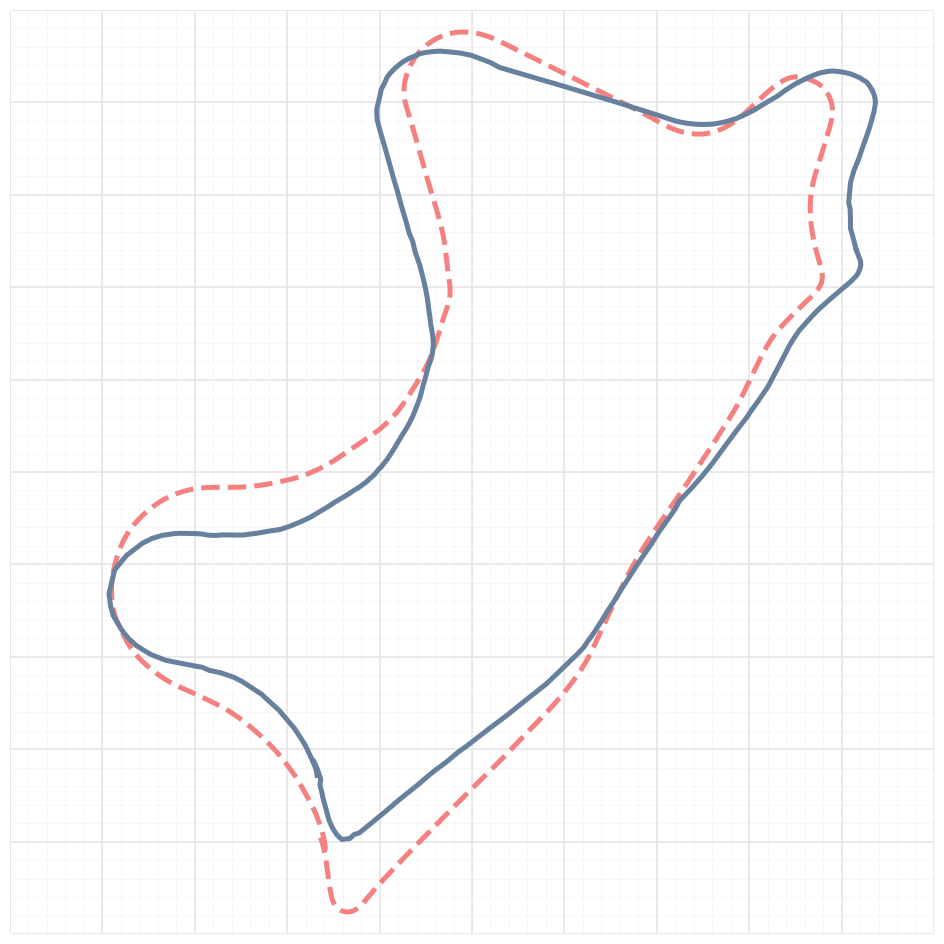} & \includegraphics[width=\linewidth]{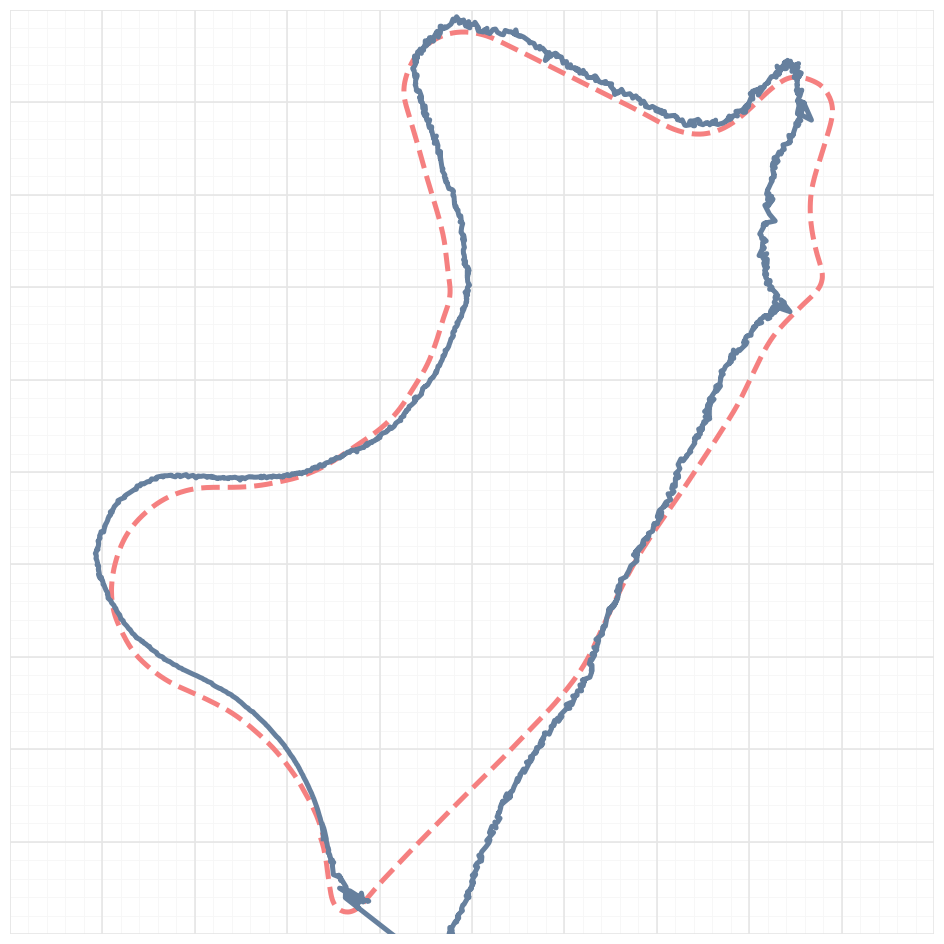} & \includegraphics[width=\linewidth]{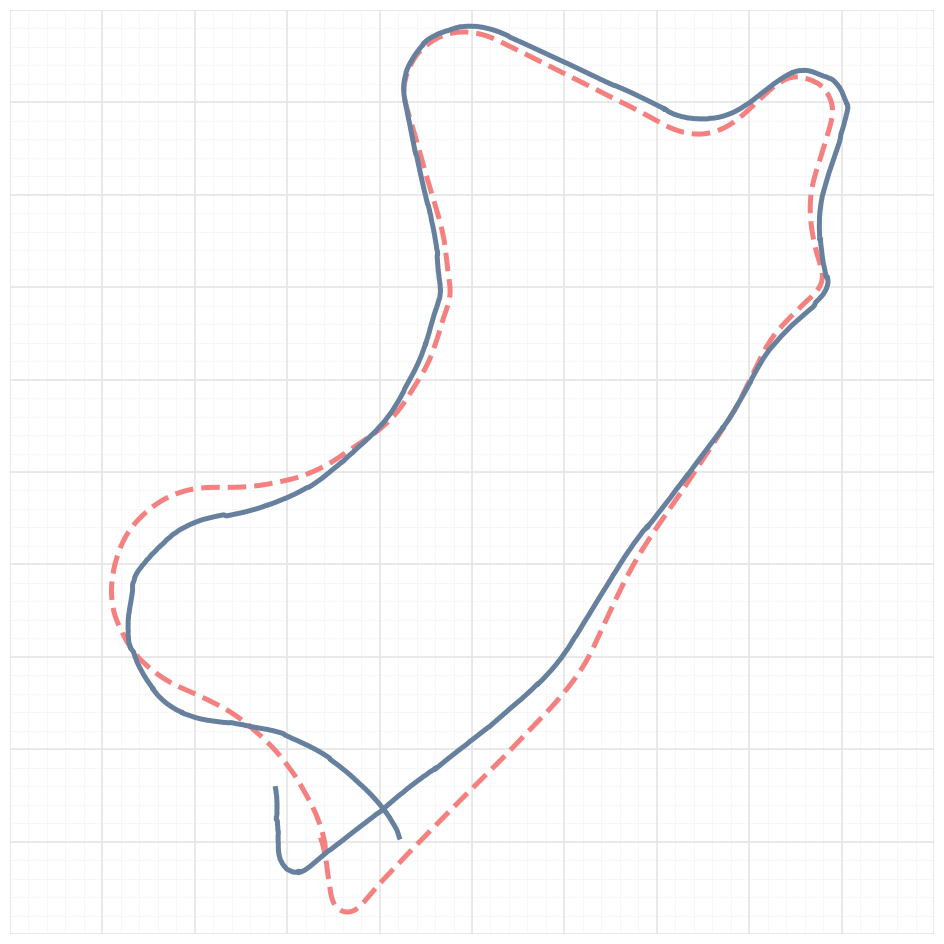} & \includegraphics[width=\linewidth]{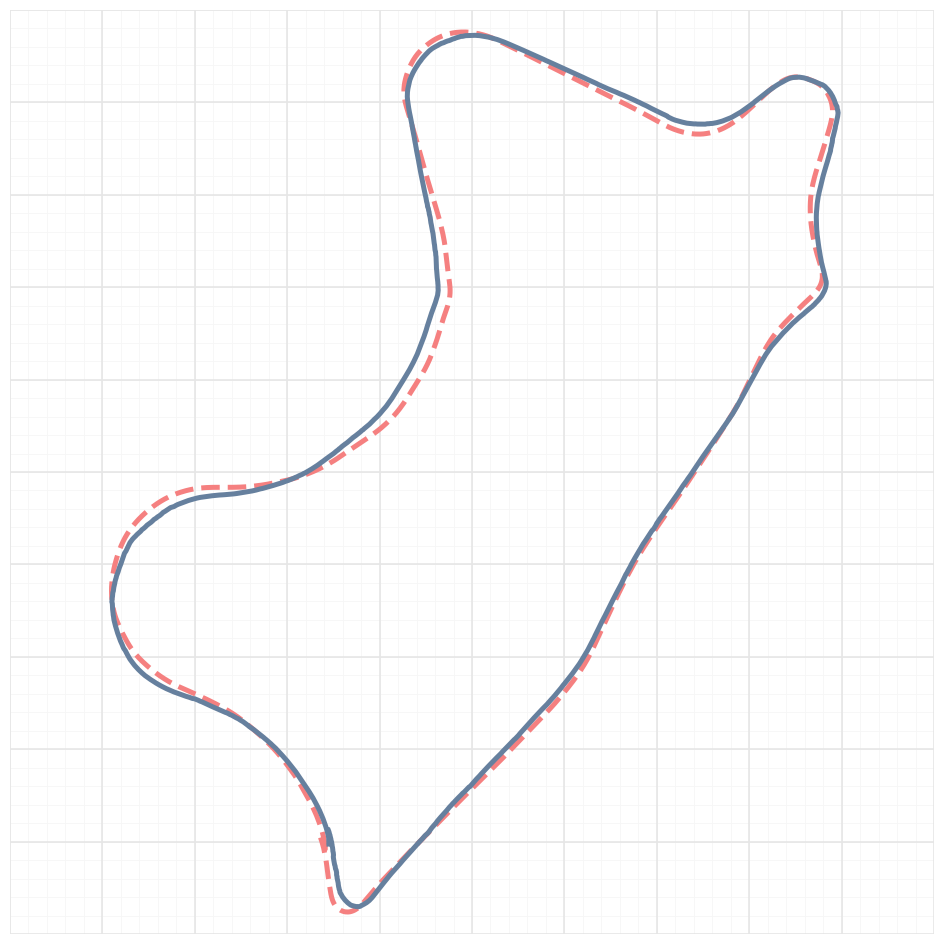} \\


\raisebox{-22pt}{\rotatebox{90}{\texttt{VBR campus\_1}}} & \includegraphics[width=\linewidth]{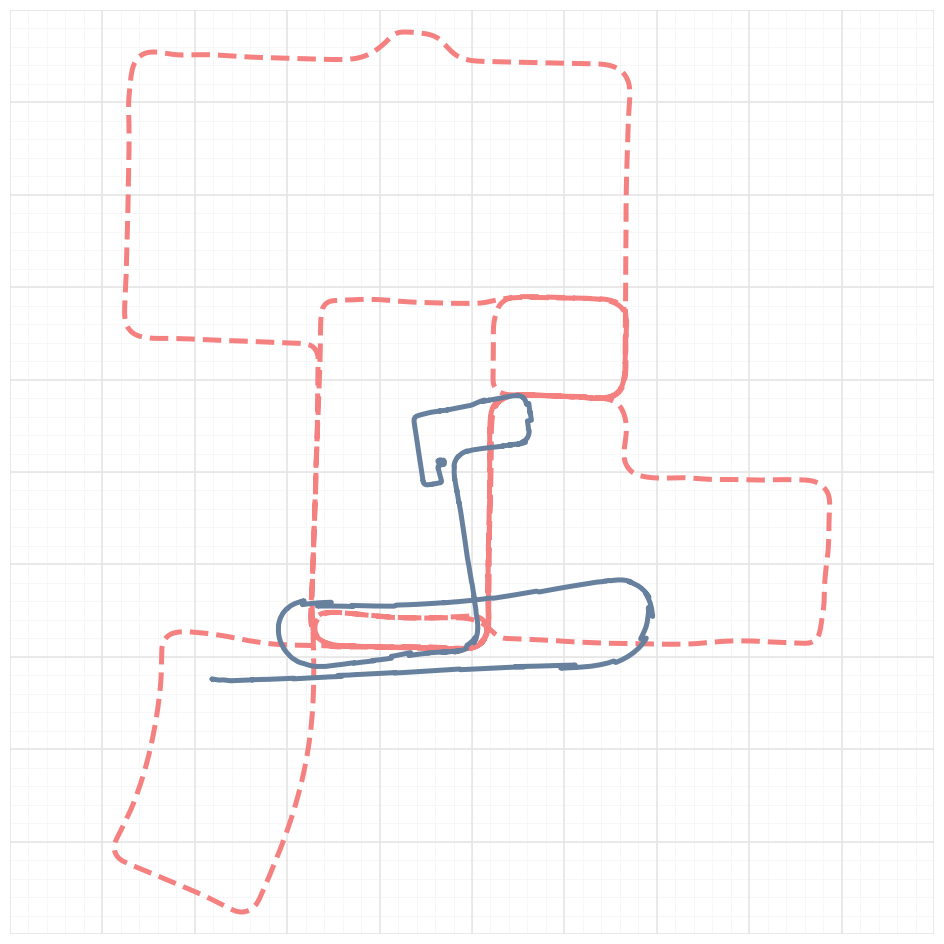} & \includegraphics[width=\linewidth]{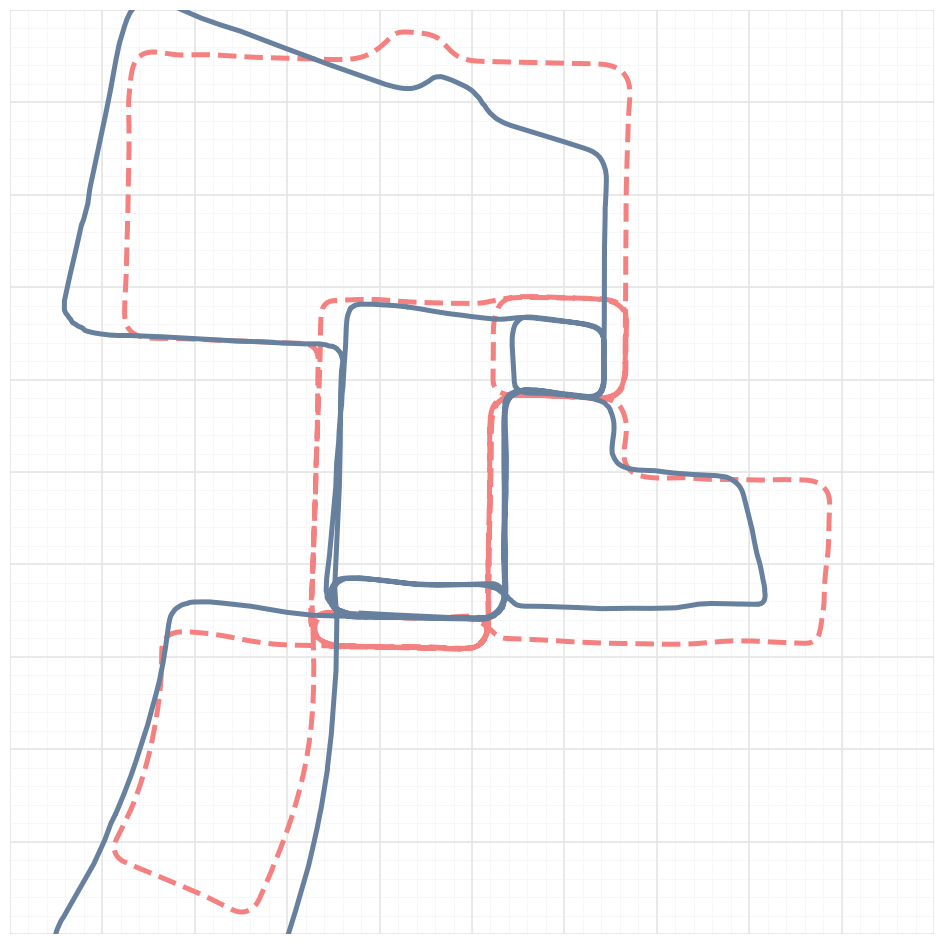} & \includegraphics[width=\linewidth]{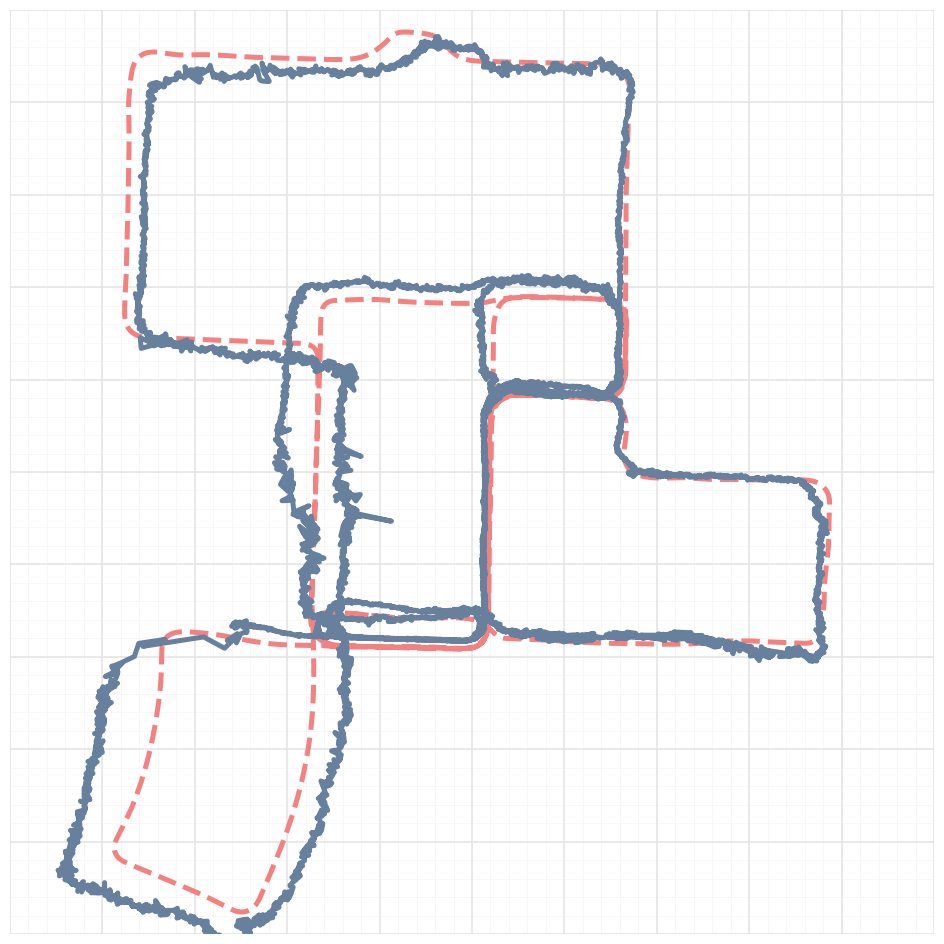} & \includegraphics[width=\linewidth]{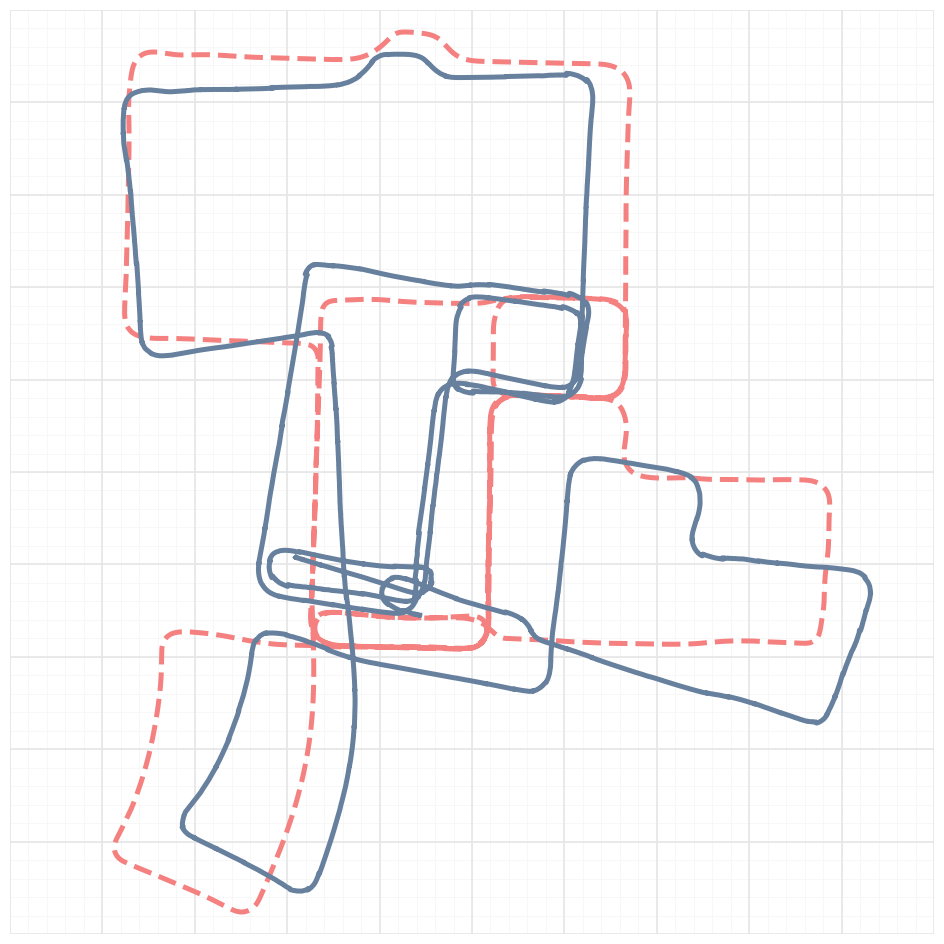} & \includegraphics[width=\linewidth]{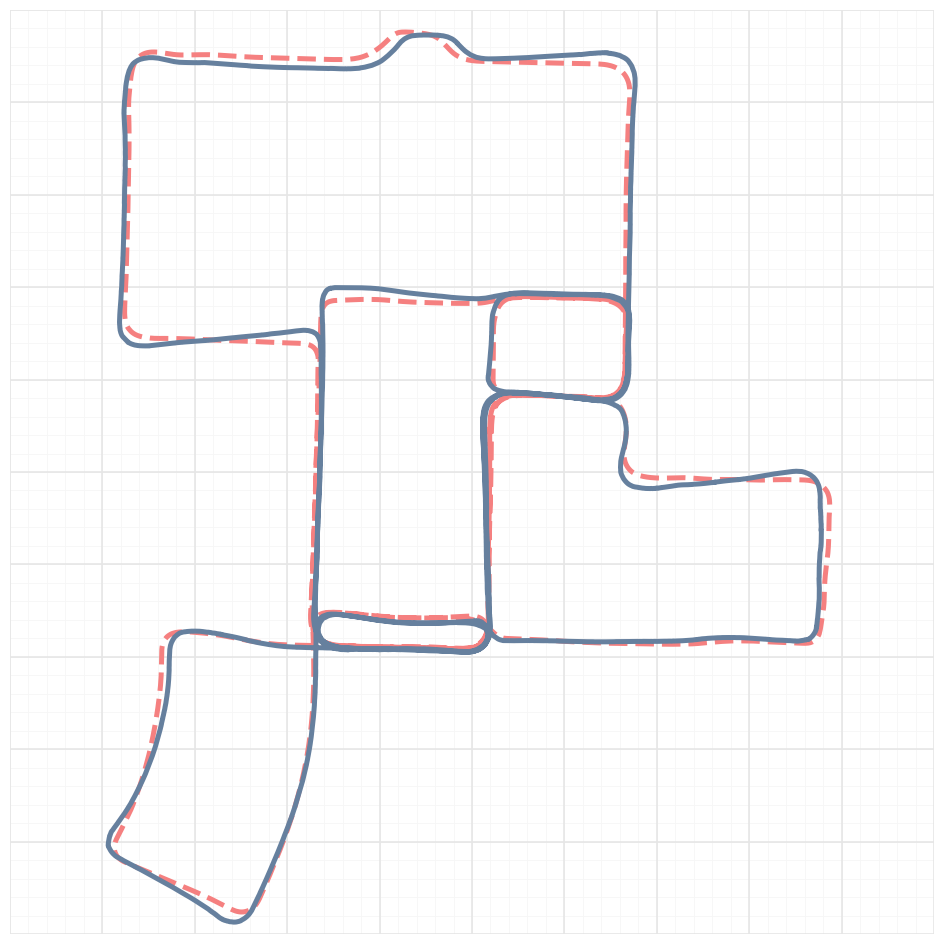} \\

\raisebox{-27pt}{\rotatebox{90}{\texttt{DROID-W down\_7}}} & \includegraphics[width=\linewidth]{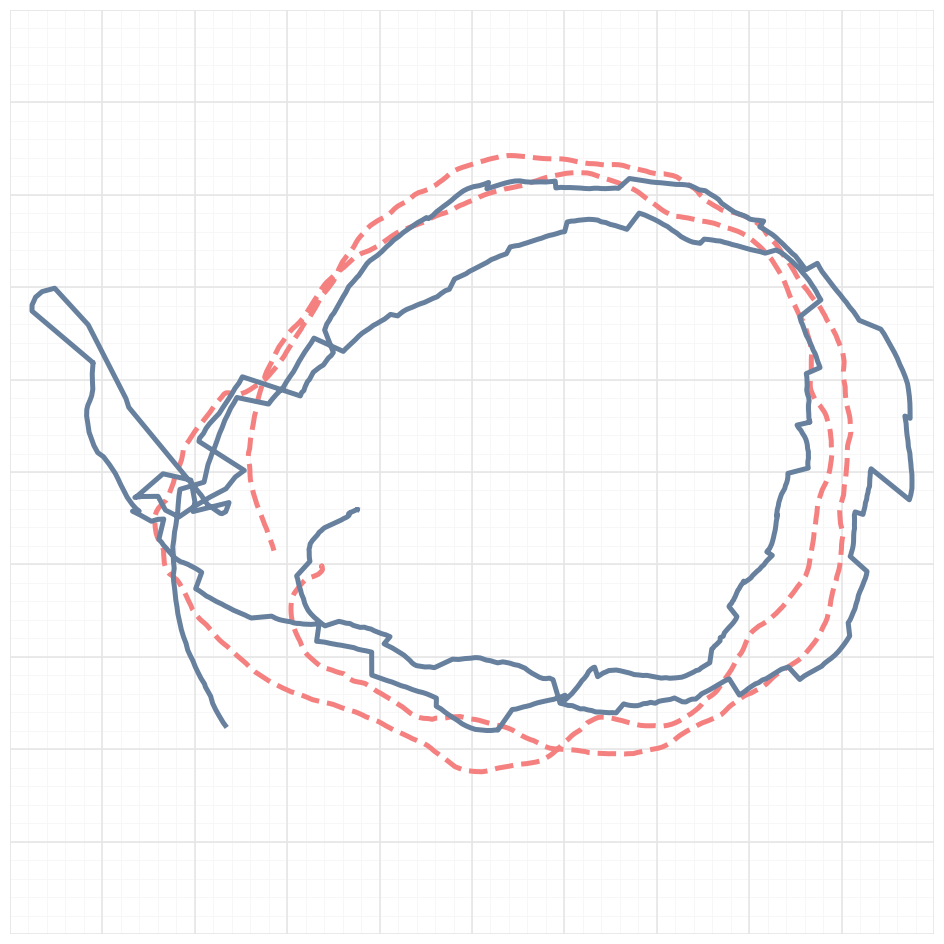} & \includegraphics[width=\linewidth]{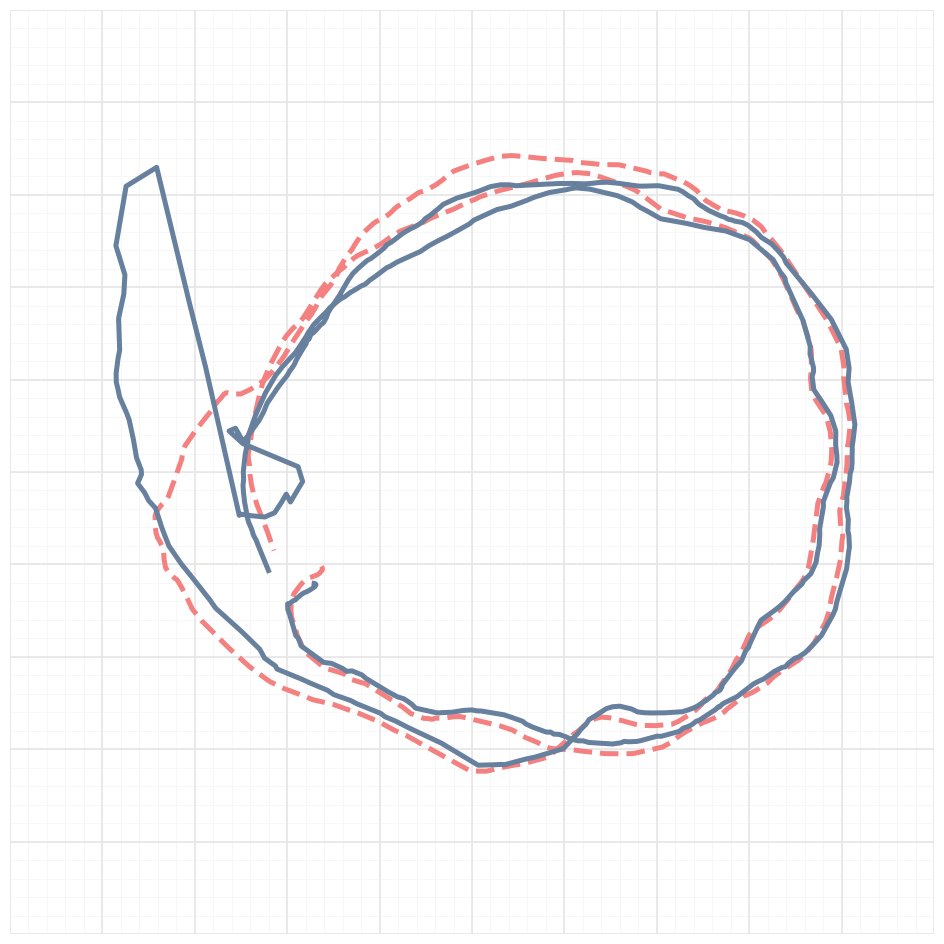} & \includegraphics[width=\linewidth]{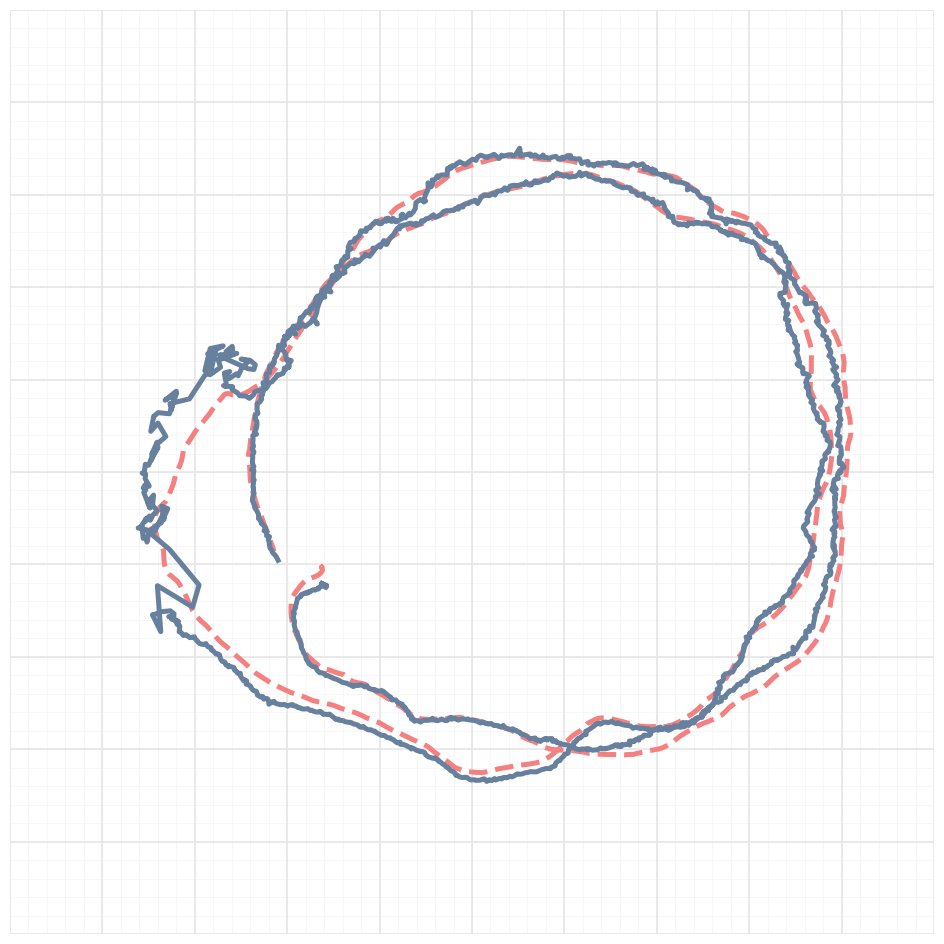} & \includegraphics[width=\linewidth]{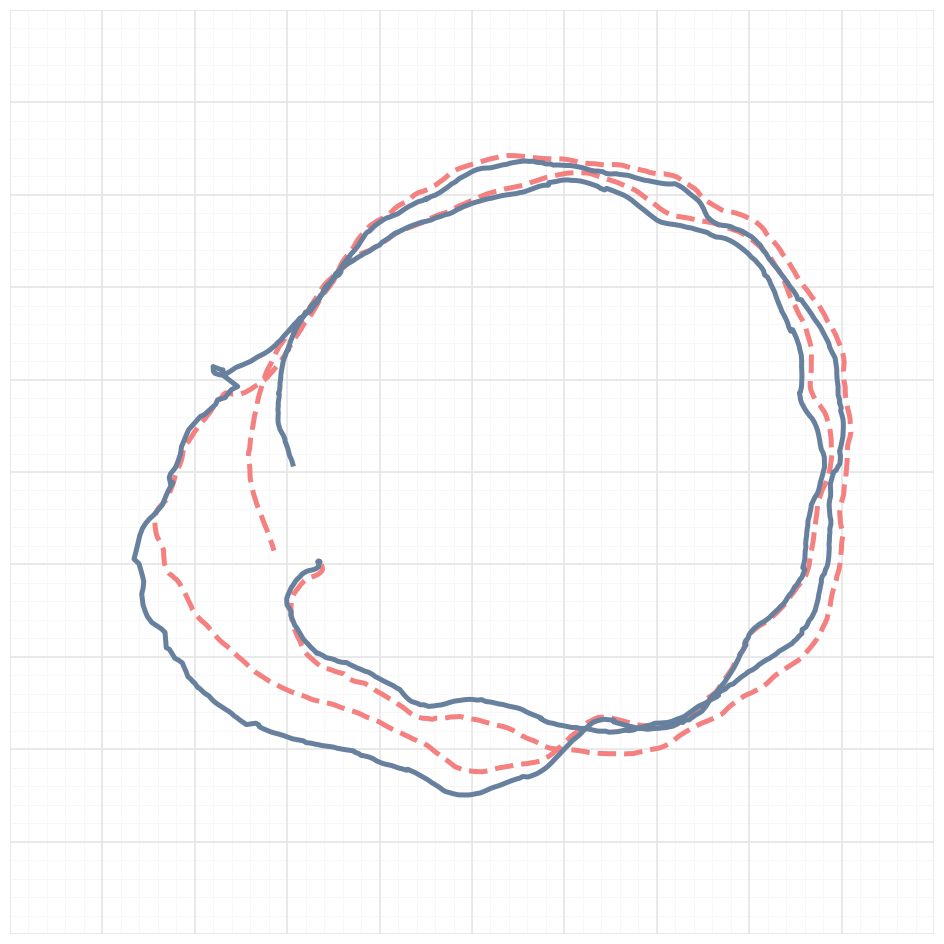} & \includegraphics[width=\linewidth]{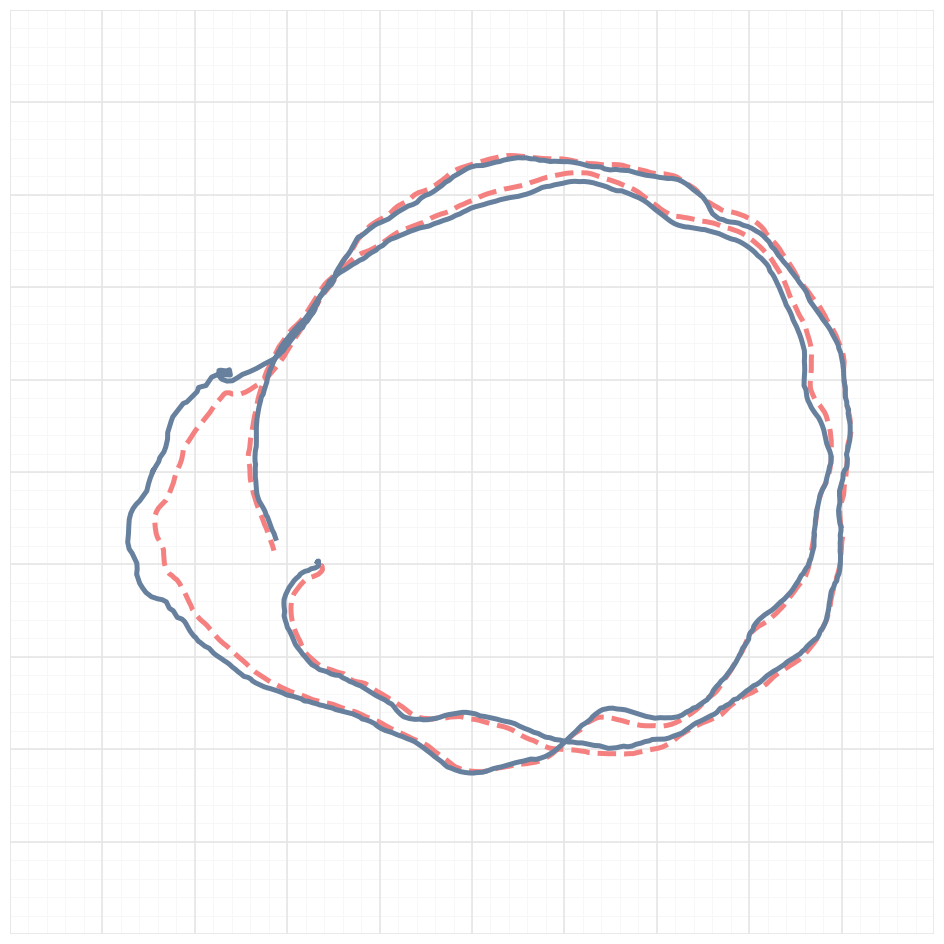} \\

\raisebox{-30pt}{\rotatebox{90}{\texttt{Oxford ke-03}}} & \includegraphics[width=\linewidth]{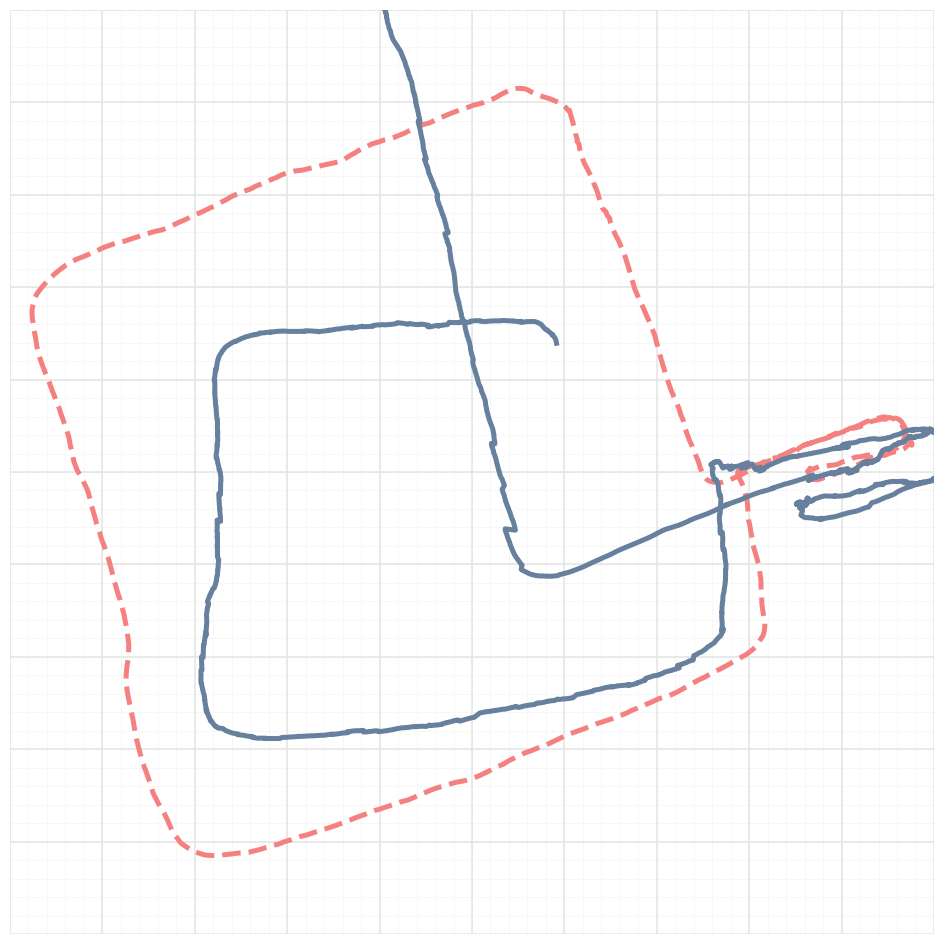} & \includegraphics[width=\linewidth]{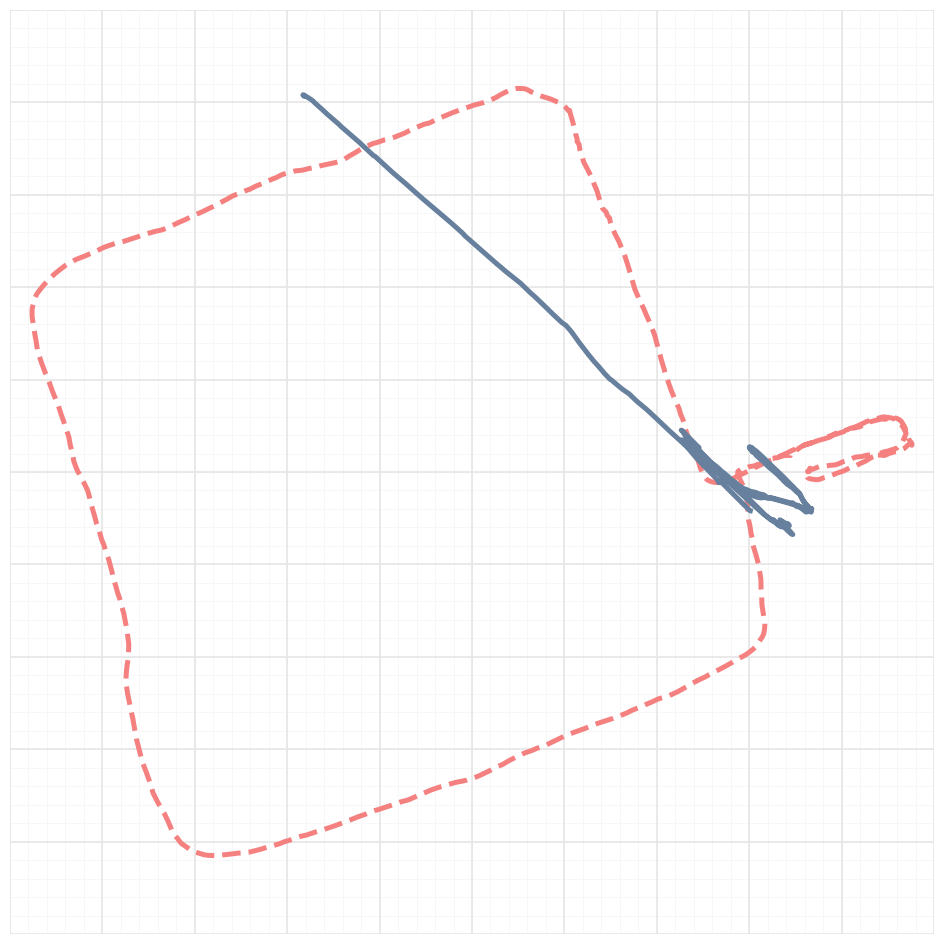} & \includegraphics[width=\linewidth]{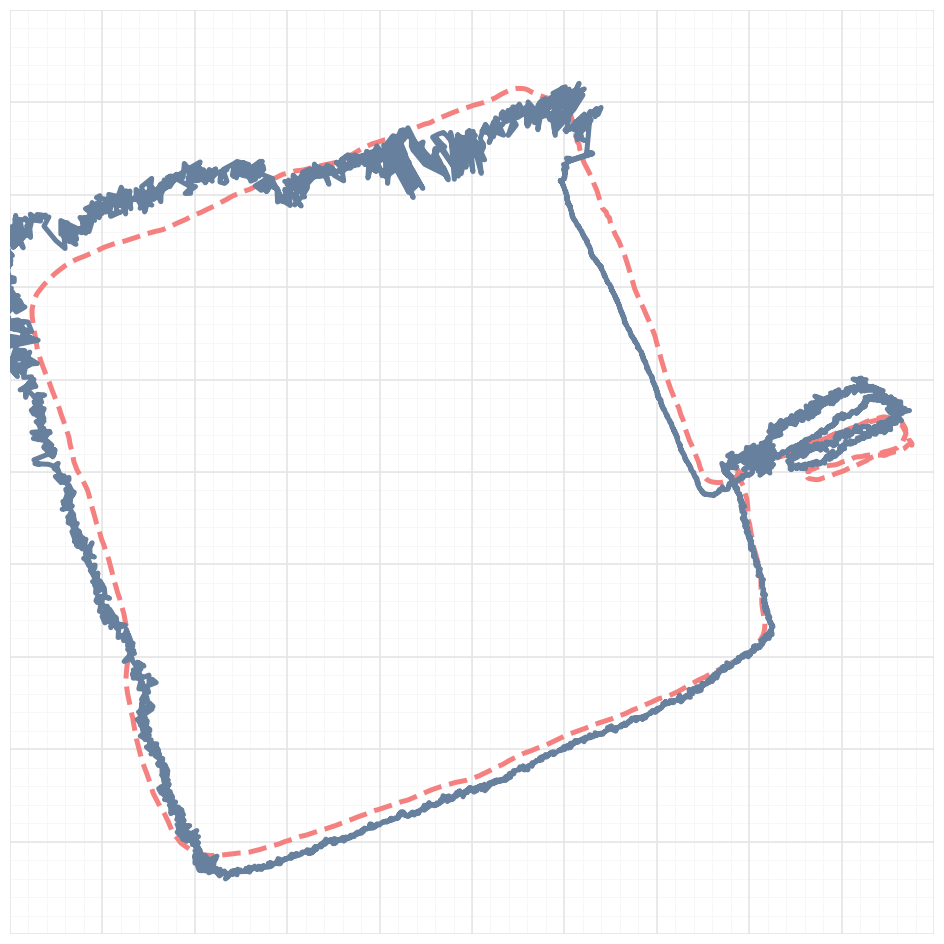} & \includegraphics[width=\linewidth]{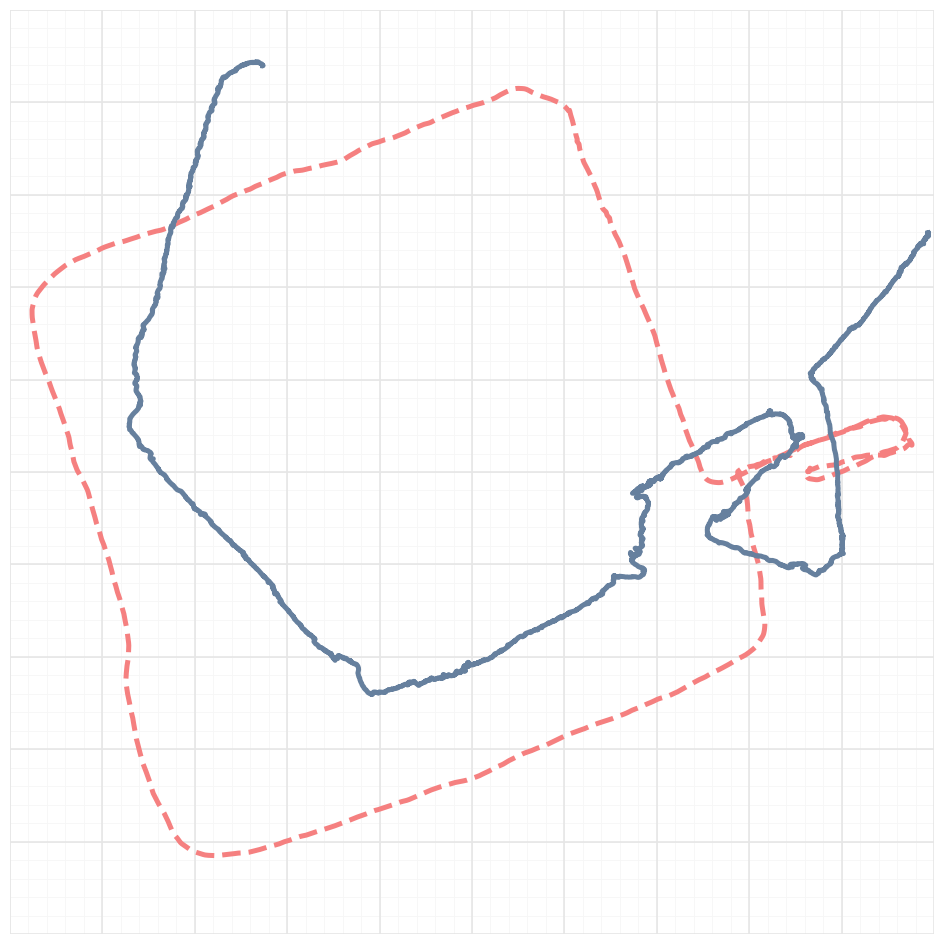} & \includegraphics[width=\linewidth]{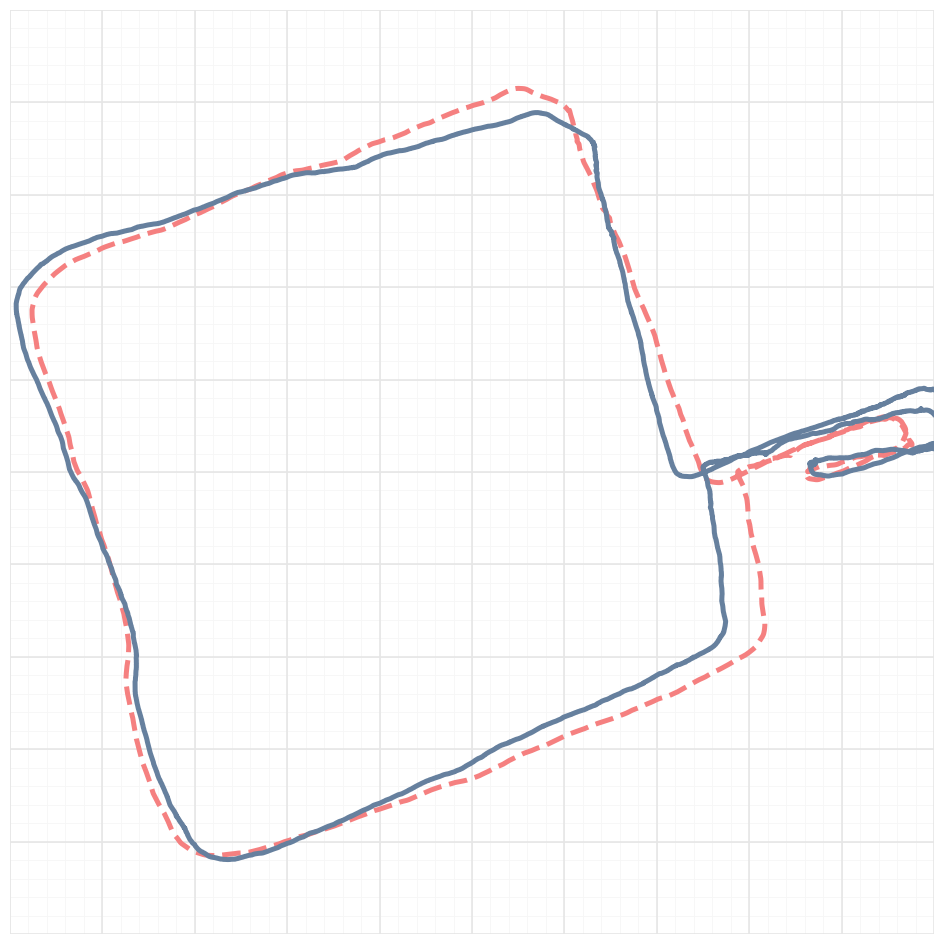} \\

\end{tabular}
\vspace{-2mm}
\caption{\textbf{Trajectory comparison} on KITTI Odometry~\citep{geiger2013kitti}, VBR~\citep{brizi2024vbr}, DROID-W~\citep{li2026droidw}, and Oxford Spires~\citep{tao2024oxford} datasets. Groundtruth trajectories (\textcolor{coral}{coral}, dashed) are overlaid with the estimates from each method (\textcolor{grayblue}{gray-blue}, solid) after Sim(3) Umeyama alignment~\citep{umeyama1991least}.}
\label{fig:trajs}
\vspace{-6mm}
\end{figure}

\paragraph{Baselines}
We conduct extensive comparisons with both \textbf{\textit{optimization-based}} and \textbf{\textit{feedforward}} methods.
For optimization-based methods, we compare against DROID-SLAM~\citep{teed2021droid}, DPV-SLAM / DPV-SLAM++~\citep{lipson2024dpv}, additionally including the dynamic SLAM methods WildGS-SLAM~\citep{zheng2025wildgs} and DROID-W~\citep{li2026droidw} on the DROID-W dataset.
For feedforward approaches, we compare against (a) \textit{streaming reconstruction foundation models}: CUT3R~\citep{wang2025cut3r}, TTT3R~\citep{chen2025ttt3r}, FastVGGT~\citep{shen2025fastvggt}, InfiniteVGGT~\citep{yuan2026infinitevggt}, AMB3R-VO~\citep{wang2025amb3r}, LingBot-Map~\citep{chen2026lingbot-map}, and LoGeR / LoGeR*~\citep{zhang2026loger}, where LoGeR* incorporates alignment-aware training and adopts a larger window size; (b) \textit{VGGT-based SLAM}: VGGT-SLAM~\citep{maggio2025vggt-slam, maggio2026vggt-slam2} and VGGT-Long~\citep{deng2025vggt-long}, for which we adopt their default chunk sizes of 16 and 60, respectively.
Following prior work~\citep{zhang2026loger}, we additionally evaluate $\pi^3$~\citep{wang2025pi3} in a chunk-wise manner.

\paragraph{Quantitative Results} Camera tracking comparisons on four benchmarks are reported in Tables~\ref{tab:vbr}, \ref{tab:kitti}, \ref{tab:droidw}, and \ref{tab:oxford}. 
As shown in \tabref{tab:vbr} and \tabref{tab:kitti}, both VBR and KITTI consist of kilometer-scale sequences, on which our approach substantially outperforms LoGeR / LoGeR*~\citep{zhang2026loger}, as our loop closure module effectively handles revisited regions. LoGeR* achieves the second-best performance due to its alignment-aware training and large window size.
Prior VGGT-based alignment methods exhibit larger tracking errors due to inaccurate point cloud alignment, and recent feedforward streaming models suffer from noticeable drift. 
We further compare against classical deep visual SLAM systems. 
DROID-SLAM~\citep{teed2021droid} and DPV-SLAM~\citep{lipson2024dpv} often fall behind in outdoor kilometer-scale scenes under monocular input, where scale drift and forward-dominant motion (e.g., KITTI Odometry) jointly degrade tracking. 
In contrast, our approach remains robust, benefiting from strong 3D reconstruction priors and effective loop closure. 
DROID-W dataset contains strong dynamic motion (e.g., moving crowds), where dynamics-aware methods such as DROID-W~\citep{li2026droidw} unsurprisingly perform best. However, our method substantially outperforms all other baselines and, notably, remains competitive with in-the-wild methods despite not incorporating any dynamics-specific handling. \tabref{tab:oxford} further highlights the superiority of our approach on the challenging Oxford Spires sequences.

\paragraph{Qualitative Comparisons}
We provide qualitative point cloud comparisons in \figref{fig:point_clouds}. On \texttt{colosseo\_0}, LoGeR reconstructs the Colosseum as two misaligned duplicates (red boxes), revealing its failure to close the loop upon revisit. In contrast, our method recovers a single, globally consistent structure. The same failure pattern appears on \texttt{Downtown1}, where LoGeR yields fragmented geometry around the loop region whereas ours produces a coherent reconstruction.
\figref{fig:trajs} compares estimated trajectories on four benchmarks. VGGT-SLAM and VGGT-Long both construct submaps with VGGT and align submaps via point cloud registration on overlapping frames. However, independent per-submap predictions and noisy point clouds lead to inaccurate alignment and unreliable scale, causing substantial tracking errors. LingBot-Map~\citep{chen2026lingbot-map} produces noisy trajectories with substantial jitter, while LoGeR produces erroneous poses, especially in revisited regions due to large tracking drift. In contrast, our method delivers the most accurate and robust trajectories estimation across these sequences, benefiting from its effective loop closure mechanism.

\begin{wraptable}[13]{r}{0.5\textwidth}
\vspace{-4mm}
\centering
\caption{\textbf{Ablation study.} Average tracking error on the VBR~\citep{brizi2024vbr} dataset. We ablate five components of our method and compare against the full model.}
\label{tab:ablation}
\resizebox{0.5\textwidth}{!}{
\begin{tabular}{lr}
\toprule
Method & ATE RMSE [m] $\downarrow$ \\
\midrule
(a) LoGeR (w/o Loop Closure) & 36.16 \\
(b) Neighbor-only constr.\ & 14.25 \\
(c) w/o Huber loss & 14.15 \\
(d) Sim(3) optimization & 13.33 \\
(e) SL(4) optimization & 20.36 \\
\midrule
\textbf{Full method} & \textbf{11.15} \\
\bottomrule
\end{tabular}}
\vspace{-4mm}
\end{wraptable}
\paragraph{Ablation Study}
We report ablations of several design choices in \tabref{tab:ablation}. 
In variant~(b), we set $N_\text{seq}=1$ to enforce neighbor-only sequential constraints; the resulting performance drop shows that denser sequential constraints provide stronger short-range regularization of the trajectory.
In variant~(c), removing the Huber loss degrades performance, confirming its effectiveness in down-weighting outlier loop constraints. 
In variant~(d), we replace our SE(3) pose graph with a Sim(3) formulation that additionally optimizes per-frame scale. 
Since our loop closure window inherits scale consistency from the streaming backbone, the extra scale degrees of freedom introduce optimization ambiguity and lead to inferior accuracy. 
In variant~(e), we further replace it with an SL(4) formulation proposed in VGGT-SLAM, whose 15-DOF projective transformation introduces more degrees of freedom than necessary in our setting, resulting in additional performance degradation. Overall, our full method achieves the best performance, validating the effectiveness of each design choice.


\section{Limitations}
\label{sec:limitation}
We present a feedforward streaming reconstruction method that incorporates loop closure to substantially reduce drift over kilometer-scale trajectories. However, when the underlying odometry suffers not from accumulated drift but from complete failure, our method cannot recover the correct trajectory. 
Such failures arise when the feedforward frontend (LoGeR in our case) is pushed beyond its operating conditions, e.g., in highly dynamic environments or under aggressive camera motion.



\section{Conclusion}
\label{sec:conclusion}
We presented \textbf{CLoSeR}, a streaming reconstruction framework that revisits loop closure within a feedforward backbone, enabling accurate kilometer-scale reconstruction from uncalibrated monocular videos without any retraining. By exploiting the globally consistent scale of the streaming backbone LoGeR, we introduce a loop-conditioned window mechanism that recovers relative poses between looped frames directly in the global coordinate system. Our experiments demonstrate that pose graph optimization on the SE(3) manifold is sufficient within our framework, avoiding the Sim(3) or SL(4) optimization adopted by prior submap-based methods. 
In future work, we aim to improve the robustness of our framework under challenging conditions such as motion blur, low-texture regions, and exposure variations. We also plan to explore multi-agent collaborative reconstruction and online deployment, broadening its applicability to real-world large-scale environments.



{
\small
\bibliographystyle{plainnat}  
\bibliography{main}
}


\newpage
\appendix
\section{Technical appendices and supplementary material}
\label{sec:appendix}

\subsection{Experimental Setting}
\label{sec:exp_setting}
We adopt \textit{LoGeR} as our base model for all evaluations, as \textit{LoGeR$^{*}$} uses different weights and a larger window size that incur substantially higher GPU memory consumption.
In addition, \textit{LoGeR$^{*}$} performs an SE(3) alignment across all window predictions to enforce a more consistent global reconstruction.
We follow the evaluation protocol of LoGeR~\citep{zhang2026loger} and downsample all input images at inference to a uniform pixel budget of approximately $2.55 \times 10^{5}$ pixels. This ensures fair comparison across datasets with different native resolutions. The resulting resolutions are $924 \times 266$ for KITTI Odometry~\citep{geiger2013kitti}, $672 \times 378$ for VBR~\citep{brizi2024vbr}, $672 \times 378$ for DROID-W~\citep{li2026droidw}, and $588 \times 434$ for Oxford Spires~\citep{tao2024oxford}.

We employ a window size of 32 and an overlap size of 3 across all evaluations, except for VBR, where we adopt a window size of 48 following the LoGeR protocol~\citep{zhang2026loger}. For loop detection, we set the temporal separation threshold to $\tau_\text{time} = 4 \times \text{window size}$ and the similarity threshold to $\tau_\text{sim} = 0.7$, and retain only the top-5 most similar pairs. For pose graph optimization, we set $N_\text{seq} = 4$ to enforce stronger sequential constraints.

\subsection{Additional Results}
We provide additional qualitative comparisons of point clouds and camera trajectories in \figref{fig:point_clouds_sup} and \figref{fig:trajs_supp}, respectively. As shown in \figref{fig:point_clouds_sup}, our approach produces more geometrically consistent reconstructions, since effective loop closure enforces constraints on revisited regions and enables drift-free reconstruction. \figref{fig:trajs_supp} compares the estimated trajectories against baseline methods. VGGT-Long and VGGT-SLAM exhibit substantial tracking errors on the challenging VBR and Oxford Spires datasets, while LoGeR suffers from noticeable drift across all sequences due to the absence of explicit constraints on revisited regions. In contrast, our method robustly recovers accurate trajectories across all sequences.

\begin{table}[h]
\centering
\caption{\textbf{Runtime and GPU memory analysis.} We report the average runtime breakdown (as a percentage of total runtime), the average number of windows, FPS, peak GPU memory usage, and trajectory accuracy on the KITTI~\citep{geiger2013kitti} dataset. PGO denotes pose graph optimization.}
\label{tab:runtime}
\setlength{\tabcolsep}{3pt} 
\resizebox{\textwidth}{!}{
\begin{tabular}{lcccc!{\smash{\tikz[baseline]{\draw[densely dashed, gray!80, line width=0.8pt] (0pt,-2pt)--(0pt,8pt);}}}cc!{\smash{\tikz[baseline]{\draw[densely dashed, gray!80, line width=0.8pt] (0pt,-2pt)--(0pt,8pt);}}}c}
\toprule
Method & Loop Detection & PGO & Model Inference & \# Windows & FPS $\uparrow$ & Peak GPU Mem. $\downarrow$ & ATE RMSE [m] $\downarrow$ \\
\midrule
LoGeR~\citep{zhang2026loger} & -- & -- & 100.00\% & 73.1 & \textbf{7.60} & {16.65} GiB & 25.44\\
\project & 5.16\% & 1.12\% & \textcolor{white}{0}93.72\% & 79.5 & 7.12 & \textbf{15.69} GiB & \textbf{12.79} \\
\bottomrule
\end{tabular}}
\end{table}
\subsection{Runtime and GPU Memory Analysis}
\label{sec:runtime}
We evaluate the runtime and GPU memory usage of our method compared with LoGeR~\cite{zhang2026loger}. All experiments are conducted on a single NVIDIA RTX 4090 GPU paired with 4 cores of an AMD EPYC 9554 CPU on a Slurm cluster. As shown in \tabref{tab:runtime}, our method incurs only marginal runtime overhead, achieving 7.12 FPS compared with 7.60 FPS for LoGeR (a 6.3\% reduction in throughput), while reducing peak GPU memory usage from 16.65 GiB to 15.69 GiB (a 5.8\% reduction) due to our efficient optimization. The additional cost mainly stems from our loop closure detection and the additional loop closure windows for model inference. In exchange, our method nearly halves trajectory error, reducing ATE RMSE from 25.44 m to 12.66 m -- a substantial accuracy gain at marginal runtime cost.

\begin{table}
\centering
\caption{\textbf{Analysis of cross-window scale consistency.} We report average relative scale error on the KITTI~\citep{geiger2013kitti} dataset.}
\label{tab:scale_analysis}
\resizebox{\textwidth}{!}{
\begin{tabular}{lrr}
\toprule
Method & Relative Scale Error Regular Window $\downarrow$ & Relative Scale Error LC Window $\downarrow$ \\
\midrule
VGGT-long~\citep{deng2025vggt-long} & 18.8\% & 49.3\% \\
VGGT-SLAM 2.0~\citep{maggio2026vggt-slam2} & 32.4\% & 147.2\% \\
\midrule
\project & \textbf{5.8\%} & \textbf{6.5\%} \\
\bottomrule
\end{tabular}}
\end{table}
\subsection{Analysis of Cross-Window Scale Consistency}
We further analyze cross-window scale consistency for VGGT-Long~\citep{deng2025vggt-long}, VGGT-SLAM 2.0~\citep{maggio2026vggt-slam2}, and our CLoSeR. For each window (or submap) $i$, whether a regular window or a loop-closure one, we fit a $\mathrm{Sim}(3)$ transformation between its decoded pose estimates and the corresponding ground truth via Umeyama alignment~\cite{umeyama1991least}, and extract the scalar scale $s_i$. For VGGT-SLAM 2.0, since its loop-closure submaps consist of only two frames, we instead use the segment-length ratio. We then define a sequence-level scale reference $s_{\mathrm{global}} = \frac{1}{|\mathcal{N}|} \sum_{i \in \mathcal{N}} s_i$, computed over the set of  regular (non-loop-closure) windows $\mathcal{N}$, and compute the per-window relative scale error $e_i = |s_i / s_{\mathrm{global}} - 1|$. We report the scale error for both regular and loop-closure windows, averaged over all windows on KITTI~\cite{geiger2013kitti}. 
As shown in \tabref{tab:scale_analysis}, our method achieves substantially lower relative scale error for both regular and loop-closure windows. The improvement on regular windows benefits from LoGeR's long-context memory. The improvement on loop-closure windows comes from our loop-closure window construction: instead of relying on a separate relative pose estimator or feedforward module, it is naturally incorporated into the LoGeR streaming framework.

\begin{figure*}[htbp]
\centering
\setlength{\tabcolsep}{2pt}
\renewcommand{\arraystretch}{1.2}
\begin{tabular}{c@{\hspace{4pt}}cc}
    \addlinespace[2pt]
    \raisebox{20pt}{\rotatebox{90}{\texttt{VBR campus\_0}}} & \includegraphics[width=0.49\textwidth]{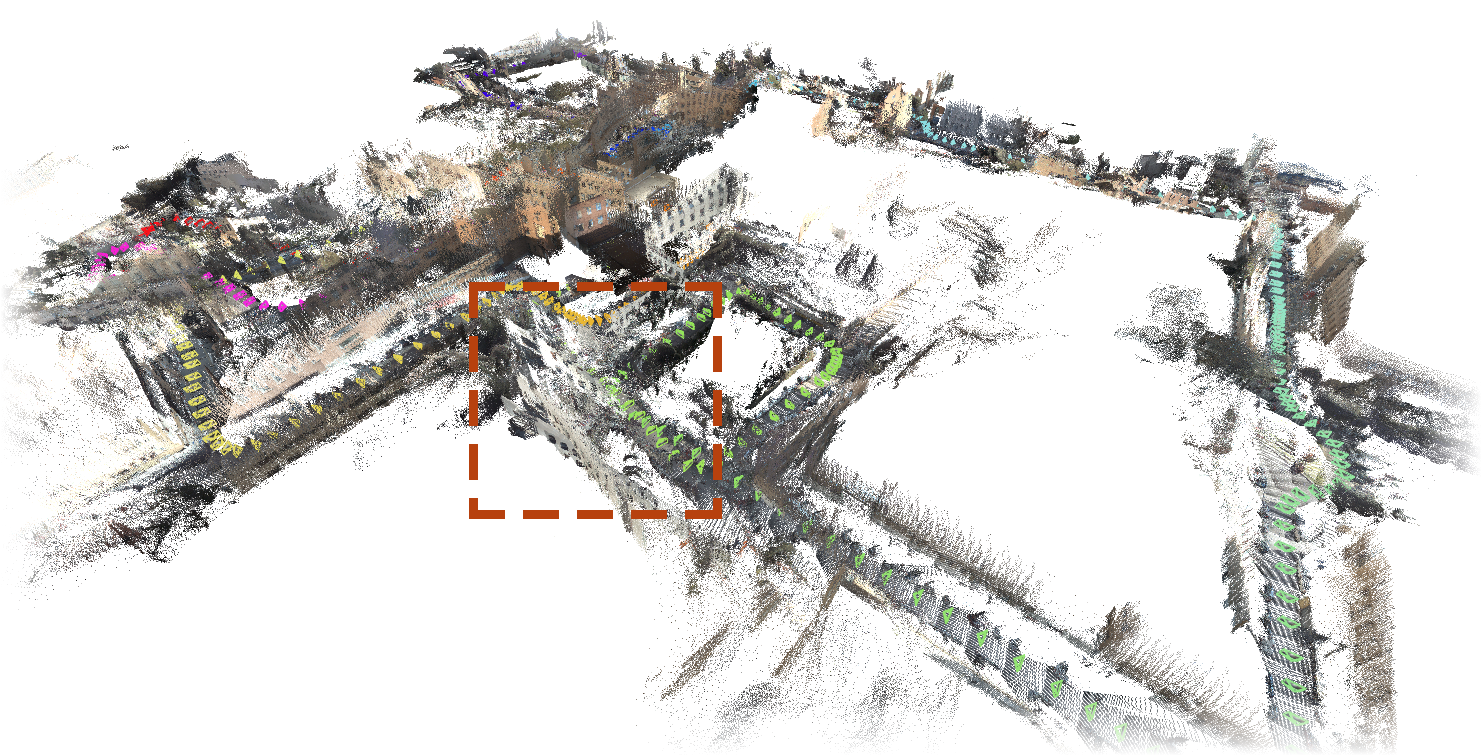} & \includegraphics[width=0.49\textwidth]{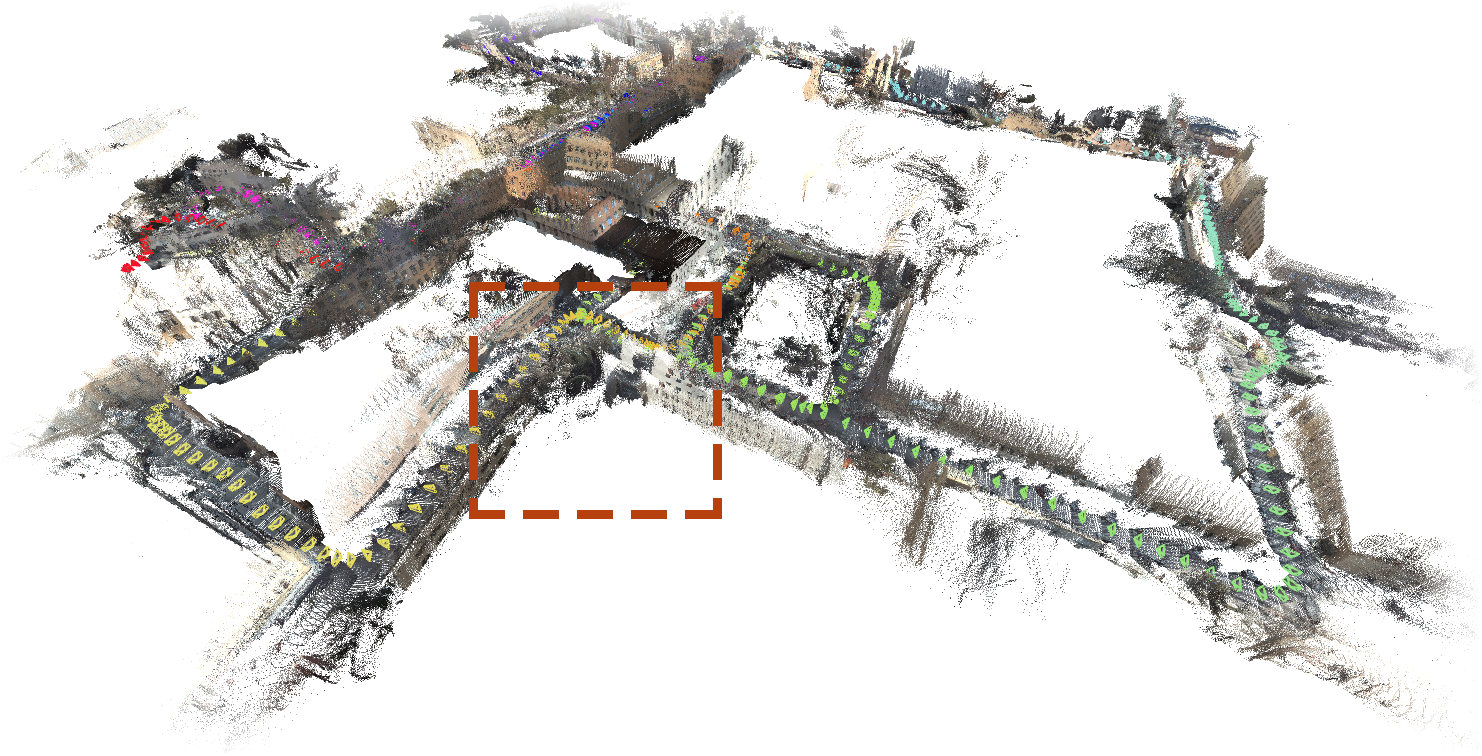} \\
    
    \raisebox{30pt}{\rotatebox{90}{{\texttt{VBR campus\_1}}}} & \includegraphics[width=0.49\textwidth]{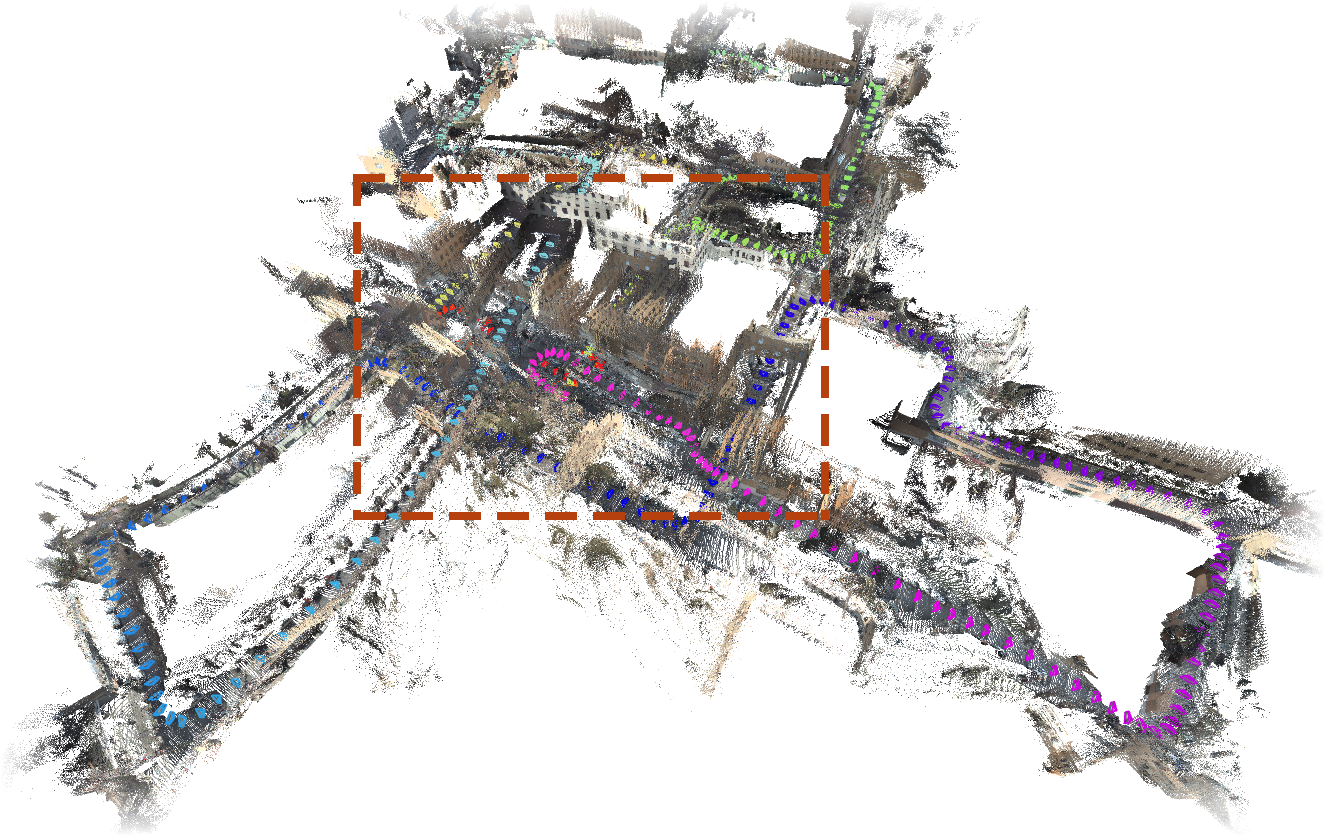} & \includegraphics[width=0.49\textwidth]{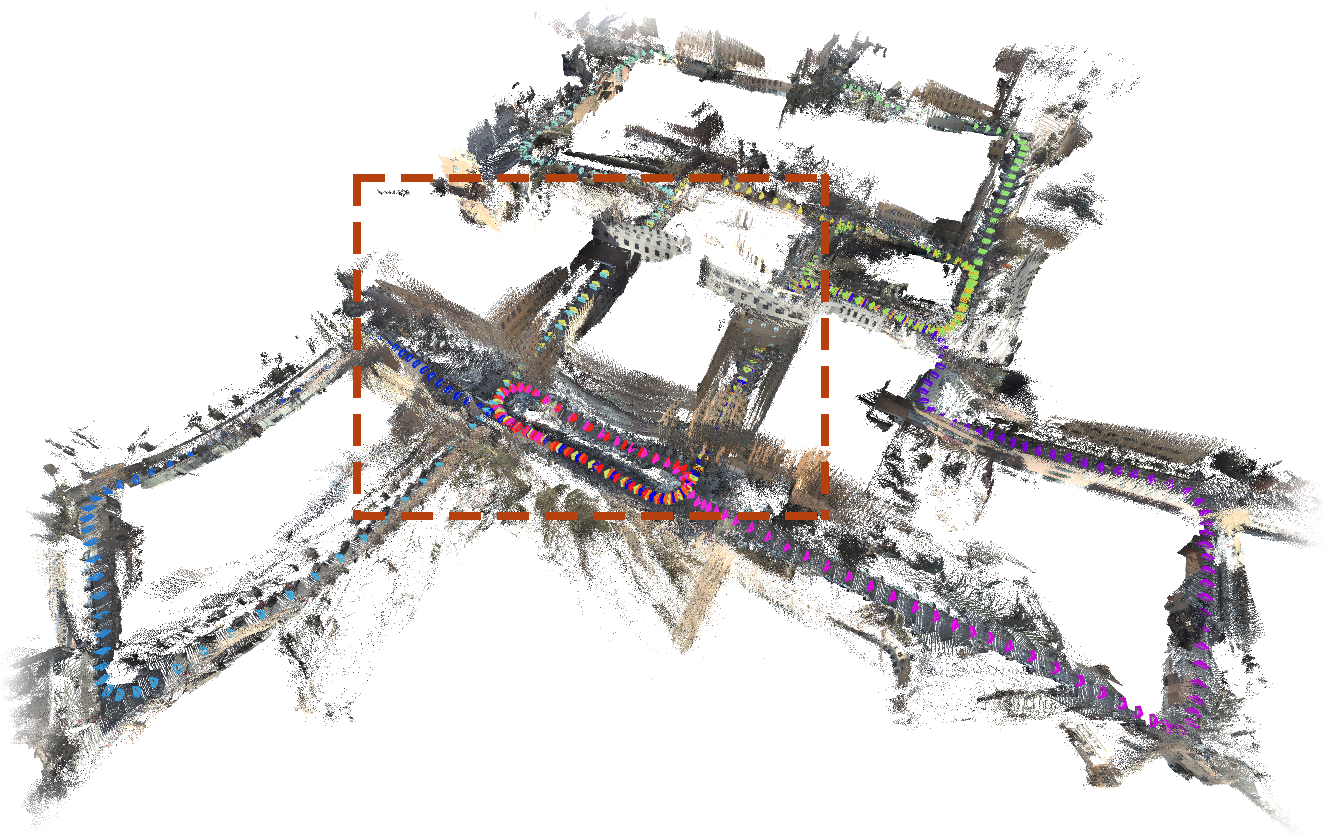} \\

    \raisebox{8pt}{\rotatebox{90}{{\texttt{VBR campus\_1 (Zoom in)}}}} & \includegraphics[width=0.49\textwidth]{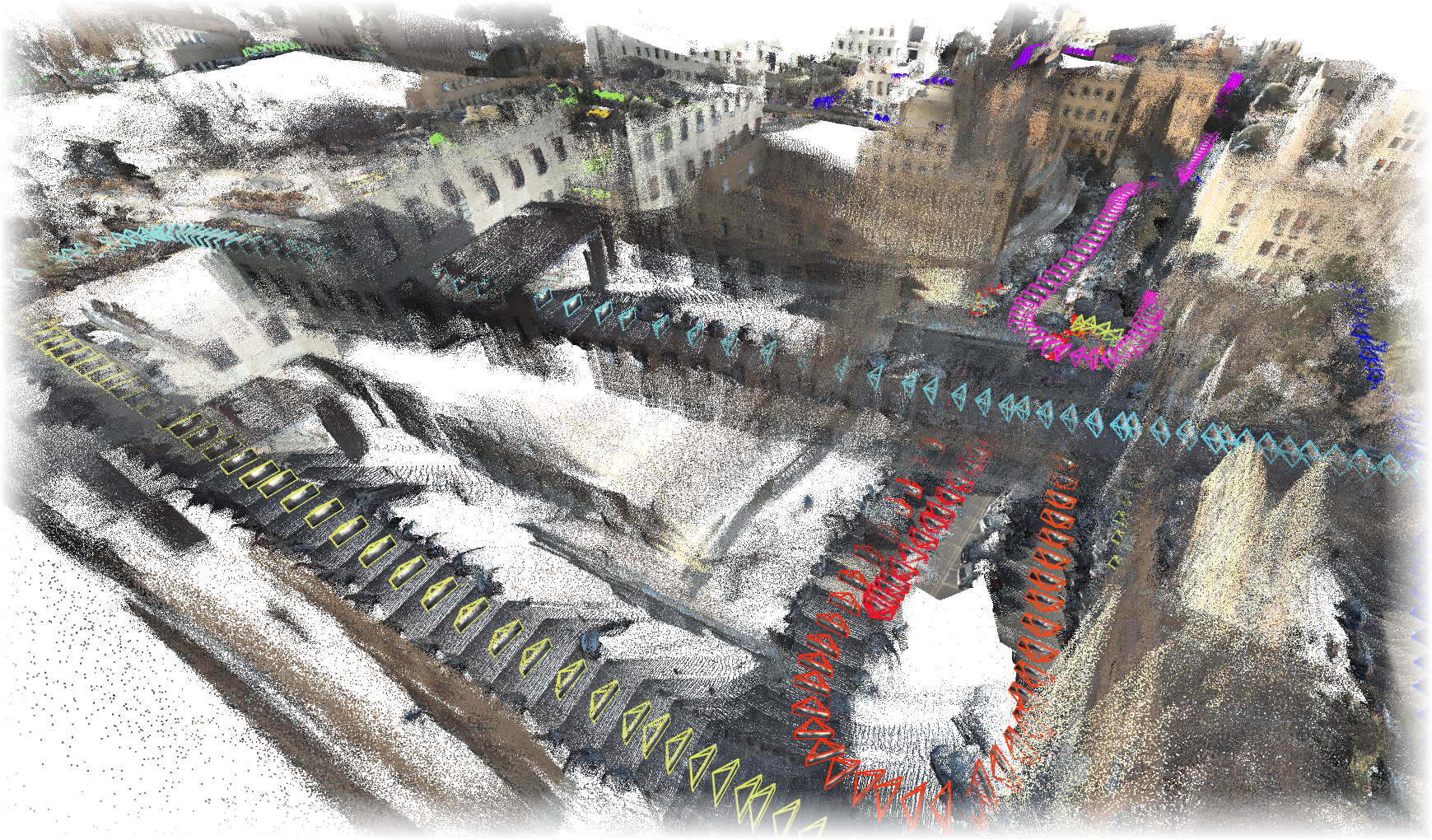} & \includegraphics[width=0.49\textwidth]{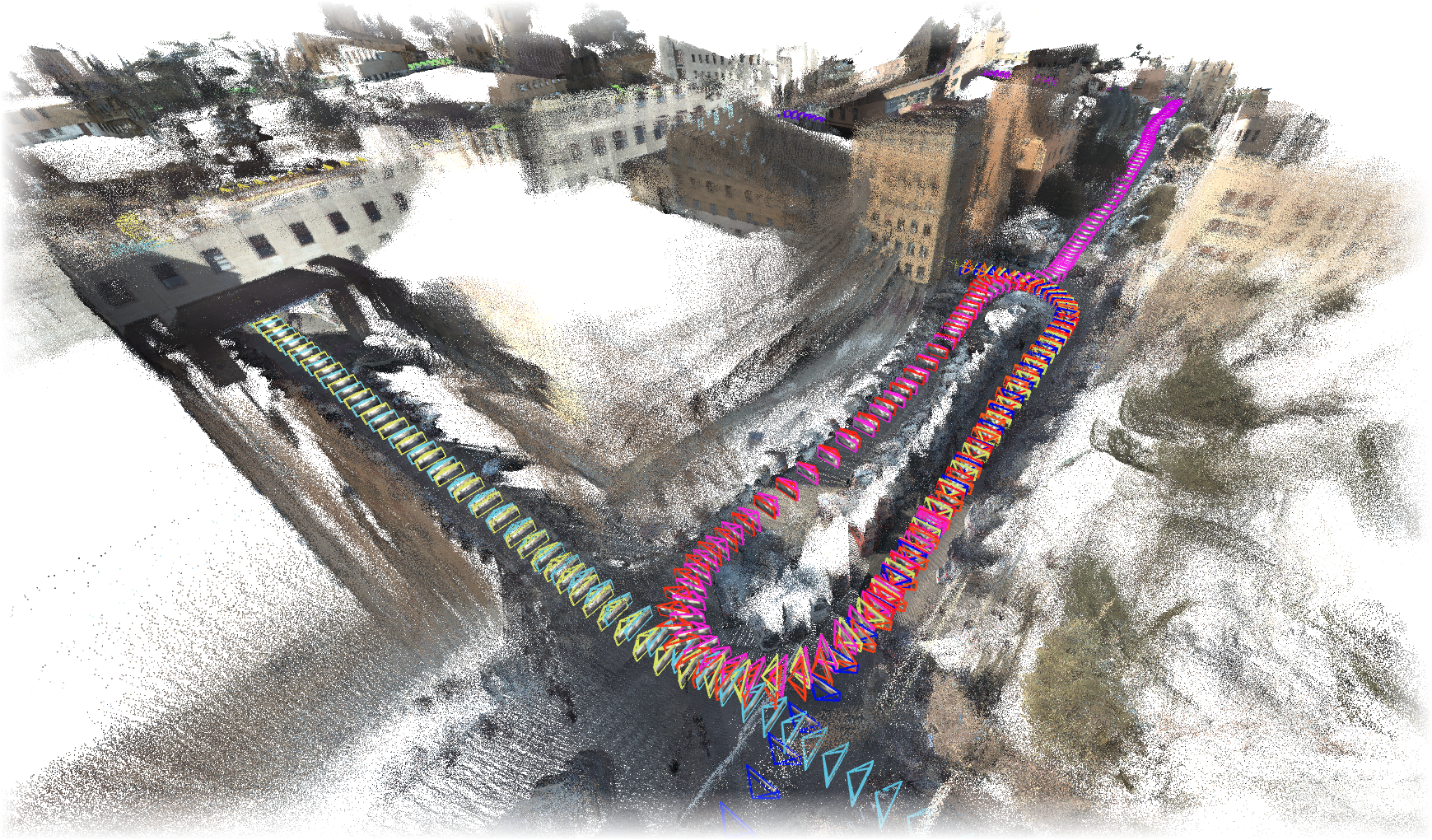} \\

    \raisebox{40pt}{\rotatebox{90}{{\texttt{KITTI 05}}}} & \includegraphics[width=0.49\textwidth]{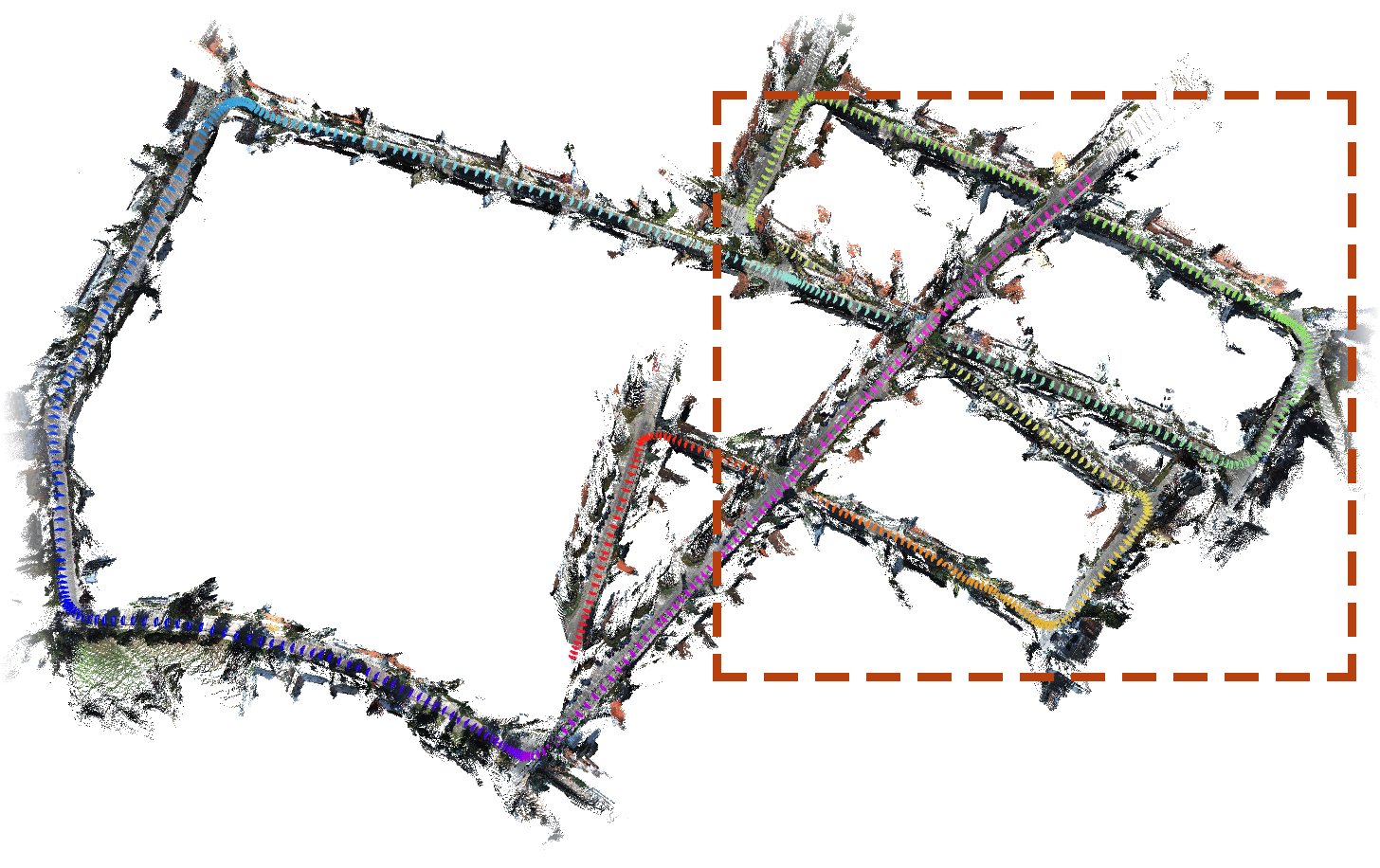} & \includegraphics[width=0.49\textwidth]{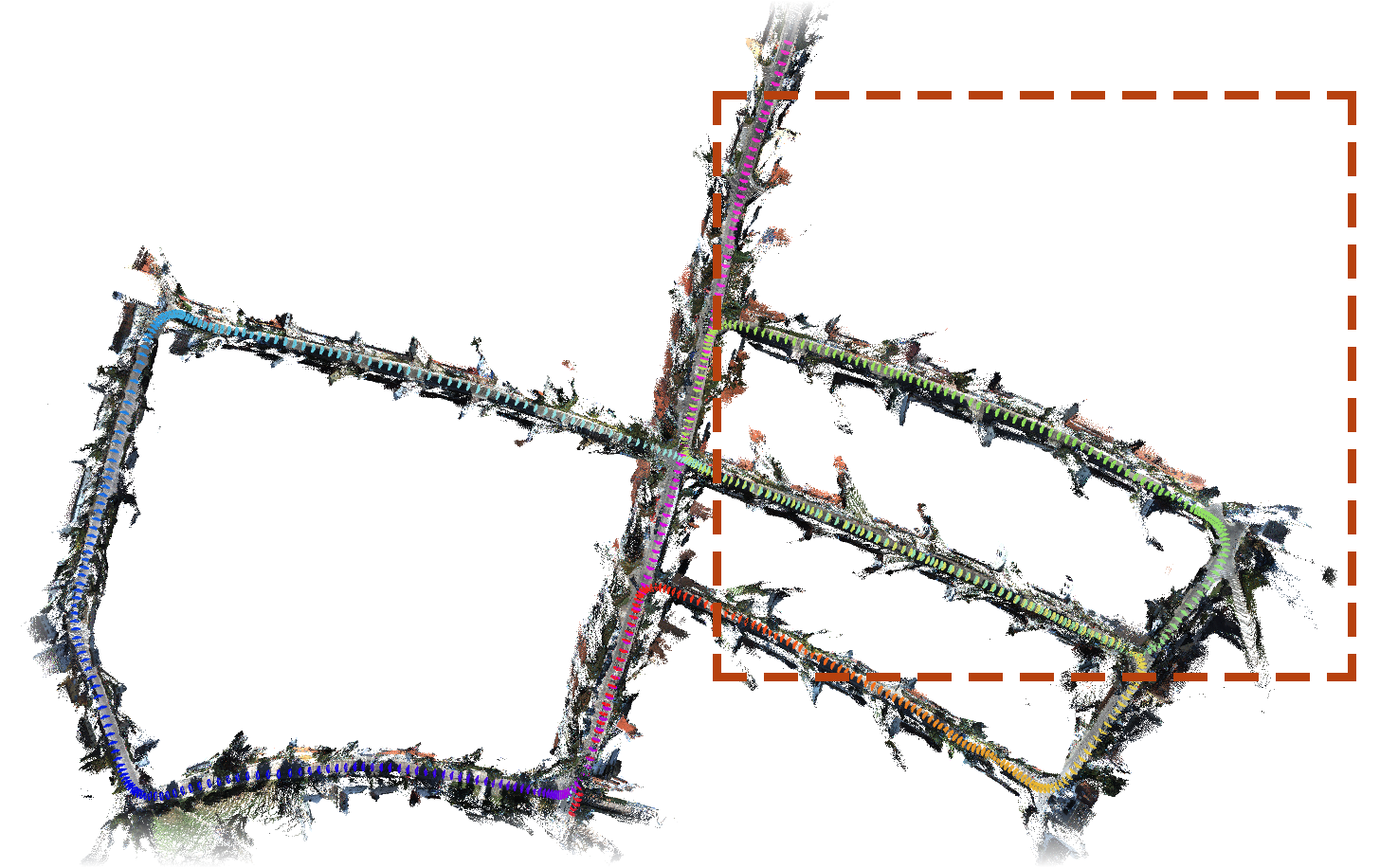} \\

    \raisebox{8pt}{\rotatebox{90}{{\texttt{KITTI 05 (Zoom in)}}}} & \includegraphics[width=0.49\textwidth]{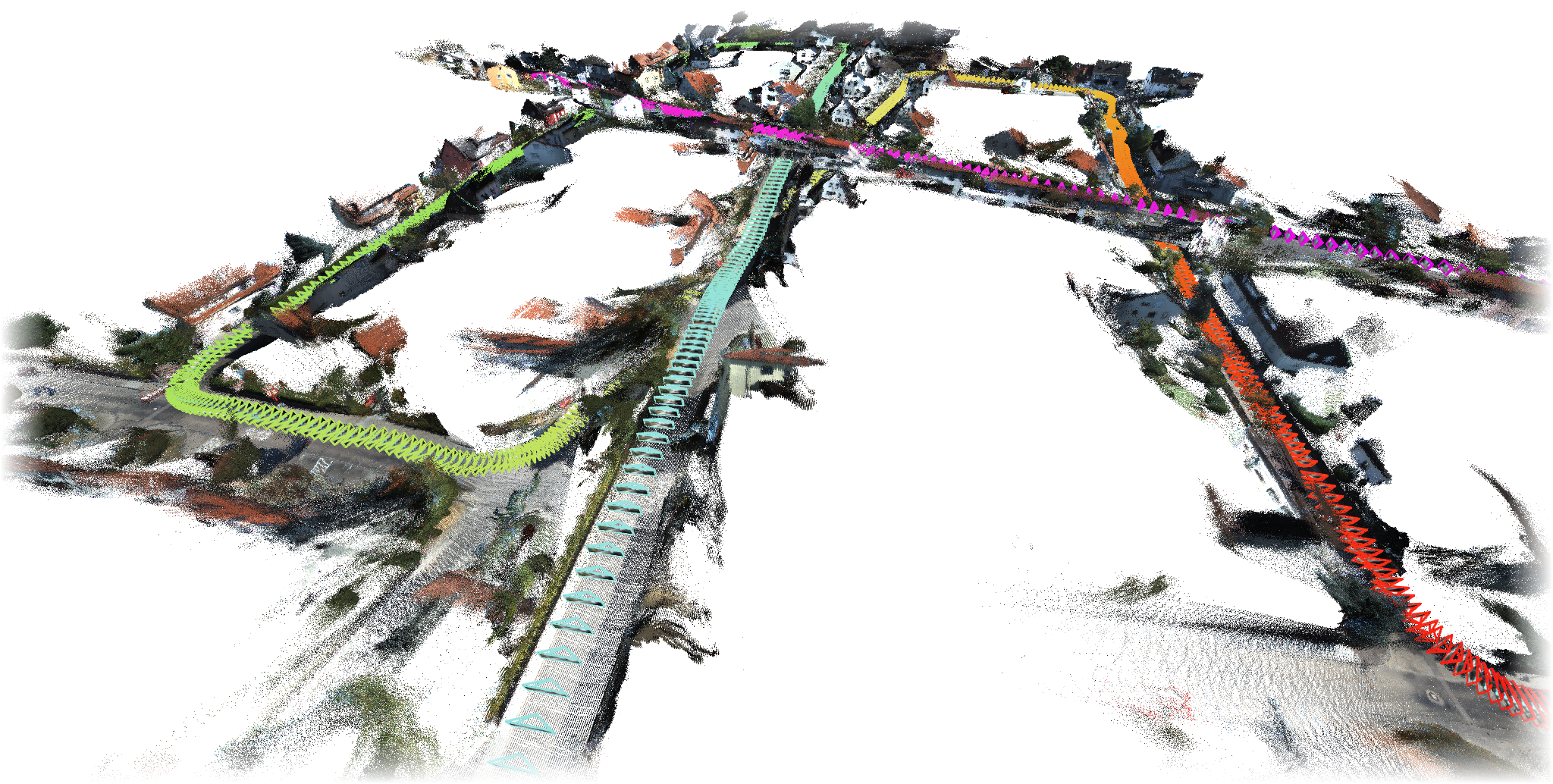} & \includegraphics[width=0.49\textwidth]{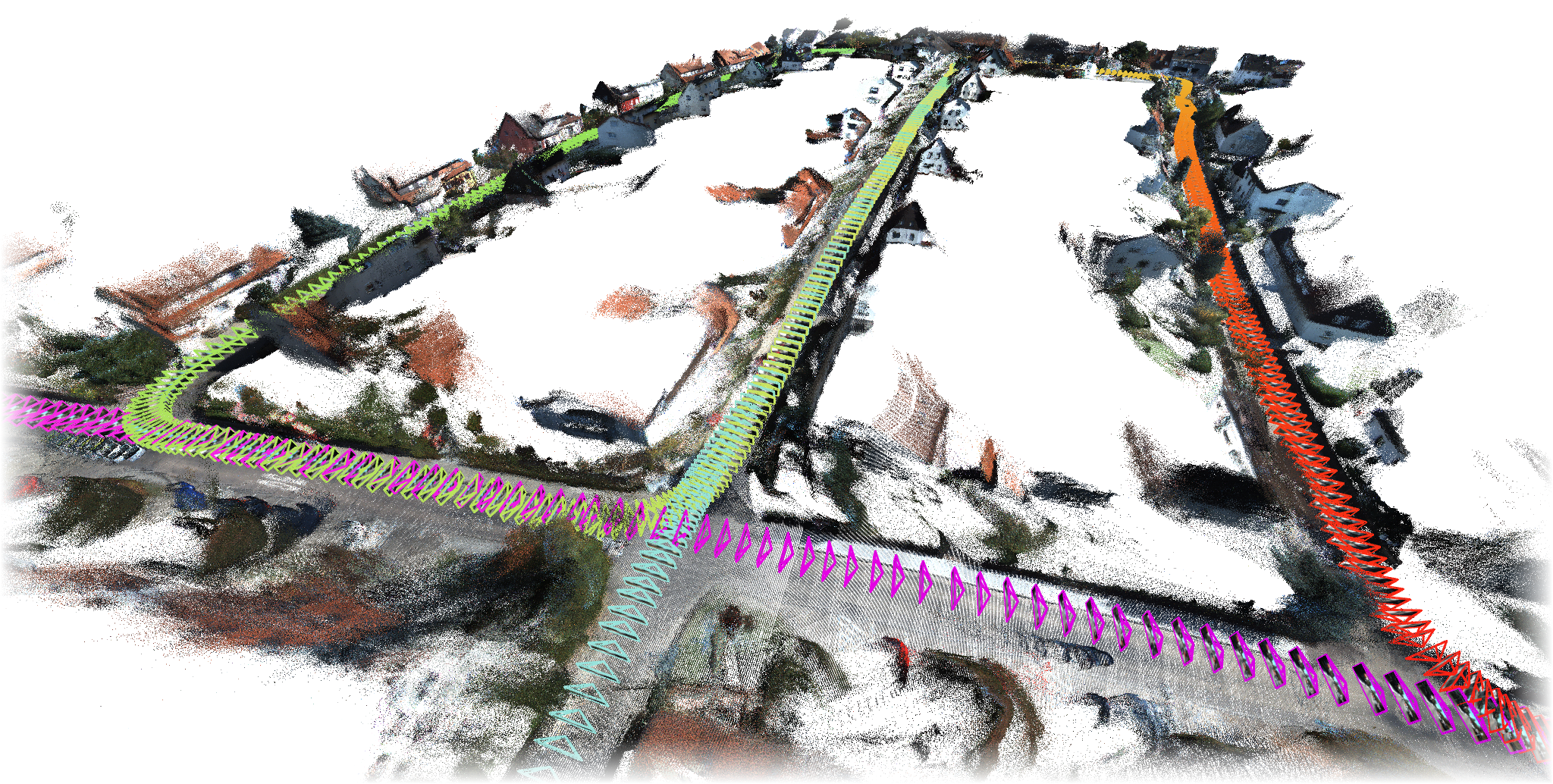} \\
    & LoGeR \citep{zhang2026loger} & \project \\
\end{tabular}
\caption{\textbf{Point clouds comparison.} Our method delivers more accurate geometry, while LoGeR exhibits noticeable ghosting artifacts caused by inaccurate pose estimation.}
\label{fig:point_clouds_sup}
\end{figure*}
\begin{figure}[htbp]
\centering
\vspace{-10pt}
\setlength{\tabcolsep}{2pt}
\renewcommand{\arraystretch}{1.0}
\begin{tabular}{@{}c@{\hspace{2pt}}*{5}{>{\centering\arraybackslash}m{0.19\linewidth}}@{}}
 & \texttt{VGGT-Long} & \texttt{VGGT-SLAM 2.0} & \texttt{LingBot-Map} & \texttt{LoGeR} & \textbf{\texttt{\project}} \\
\raisebox{-10pt}{\rotatebox{90}{\texttt{KITTI 05}}} & \includegraphics[width=\linewidth]{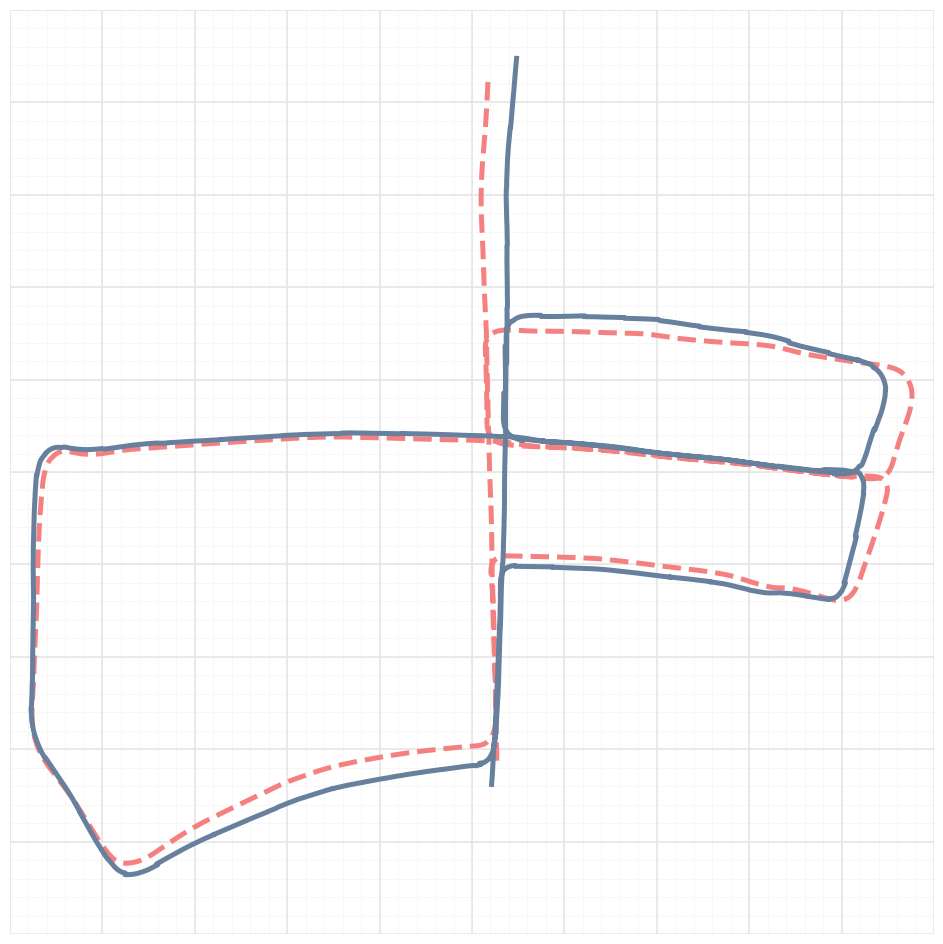} & \includegraphics[width=\linewidth]{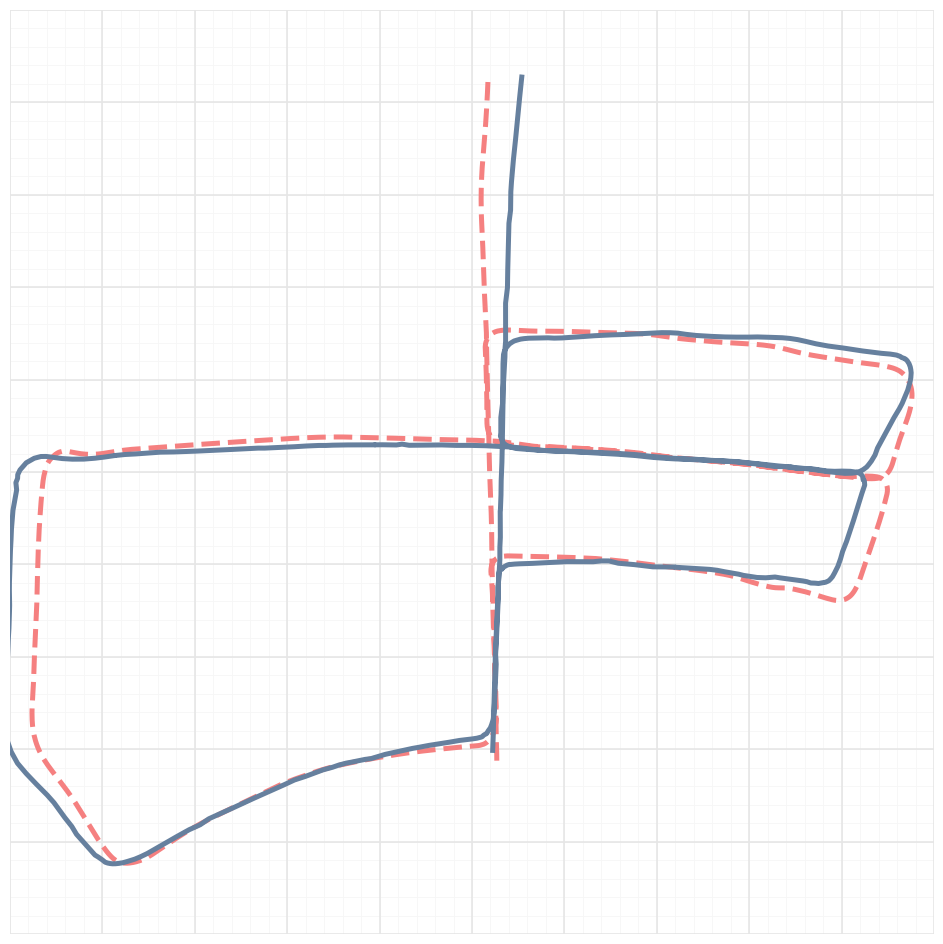} & \includegraphics[width=\linewidth]{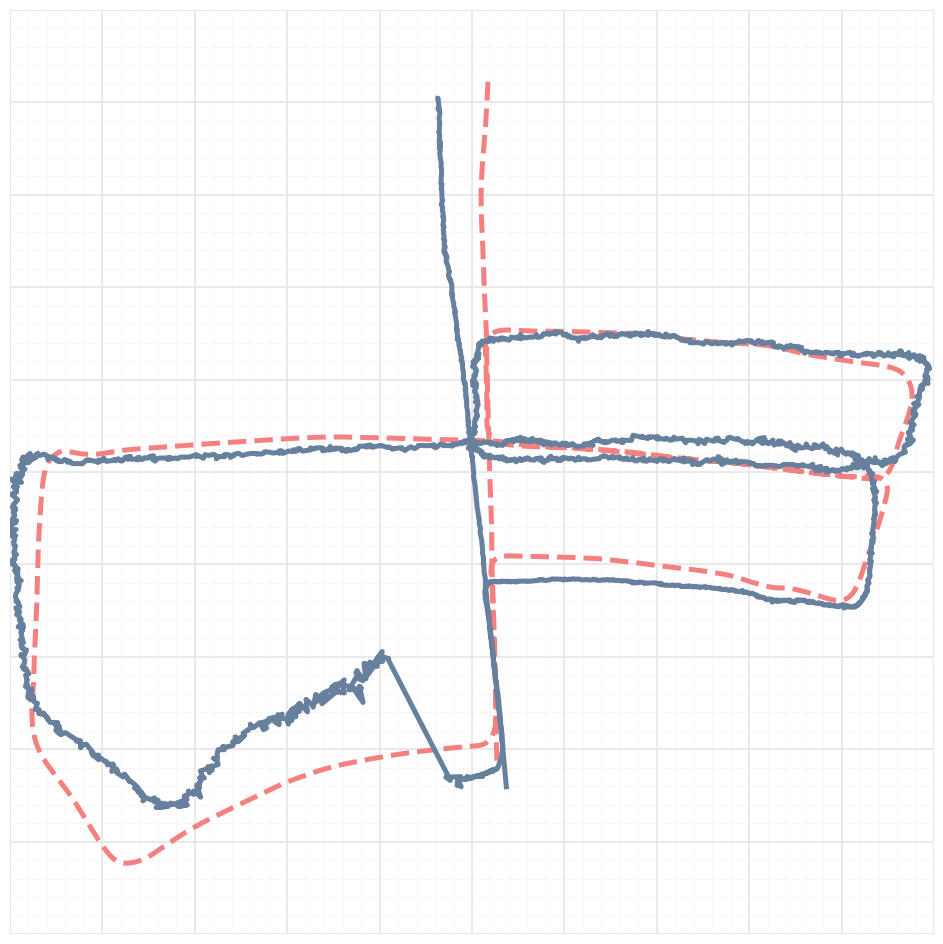} & \includegraphics[width=\linewidth]{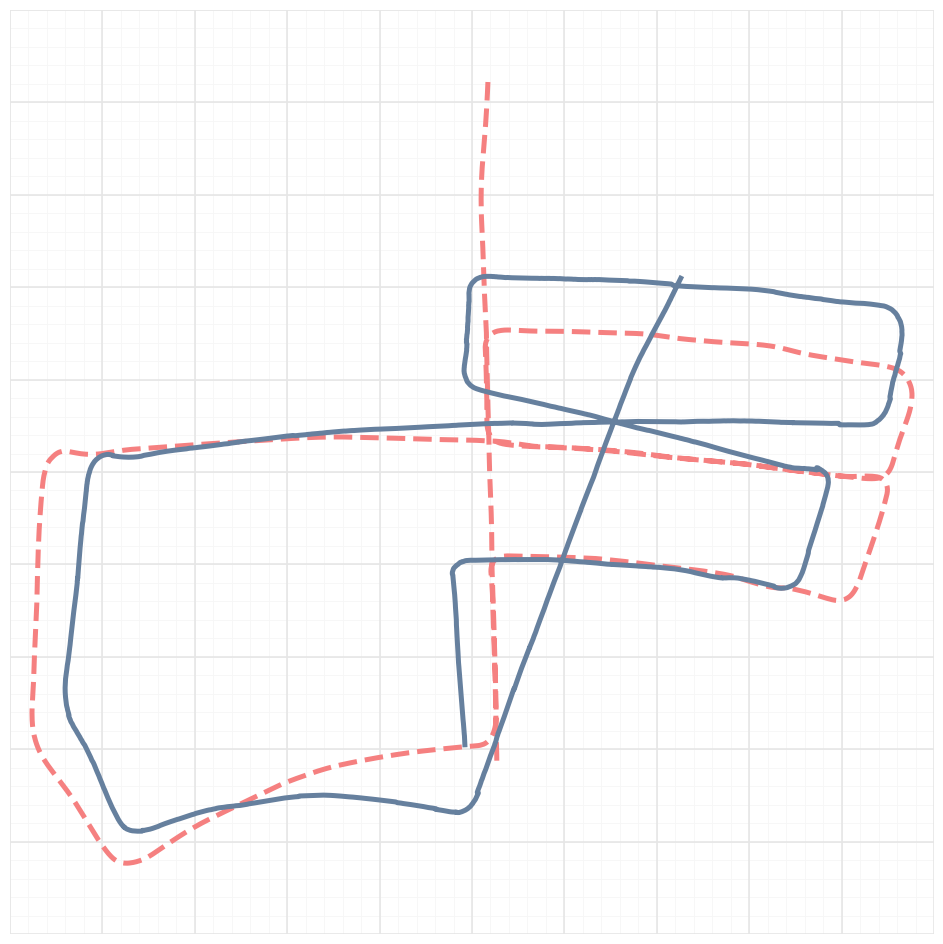} & \includegraphics[width=\linewidth]{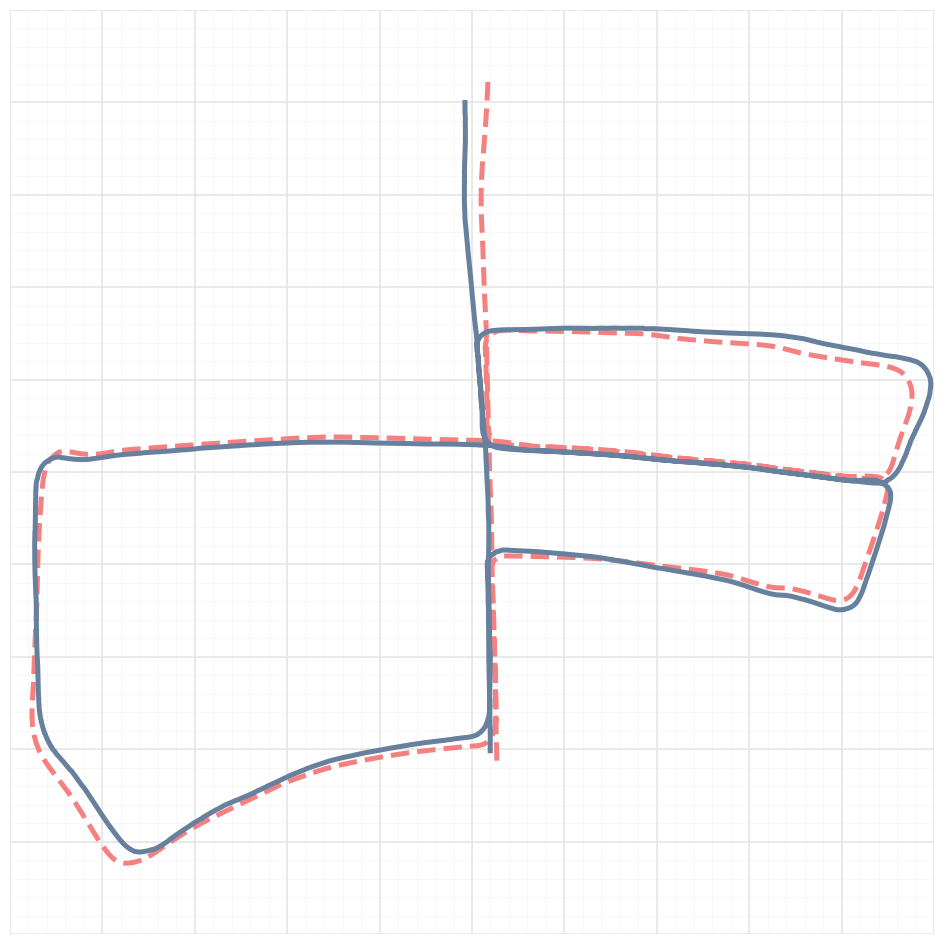} \\

\raisebox{-25pt}{\rotatebox{90}{\texttt{VBR campus\_0}}} & \includegraphics[width=\linewidth]{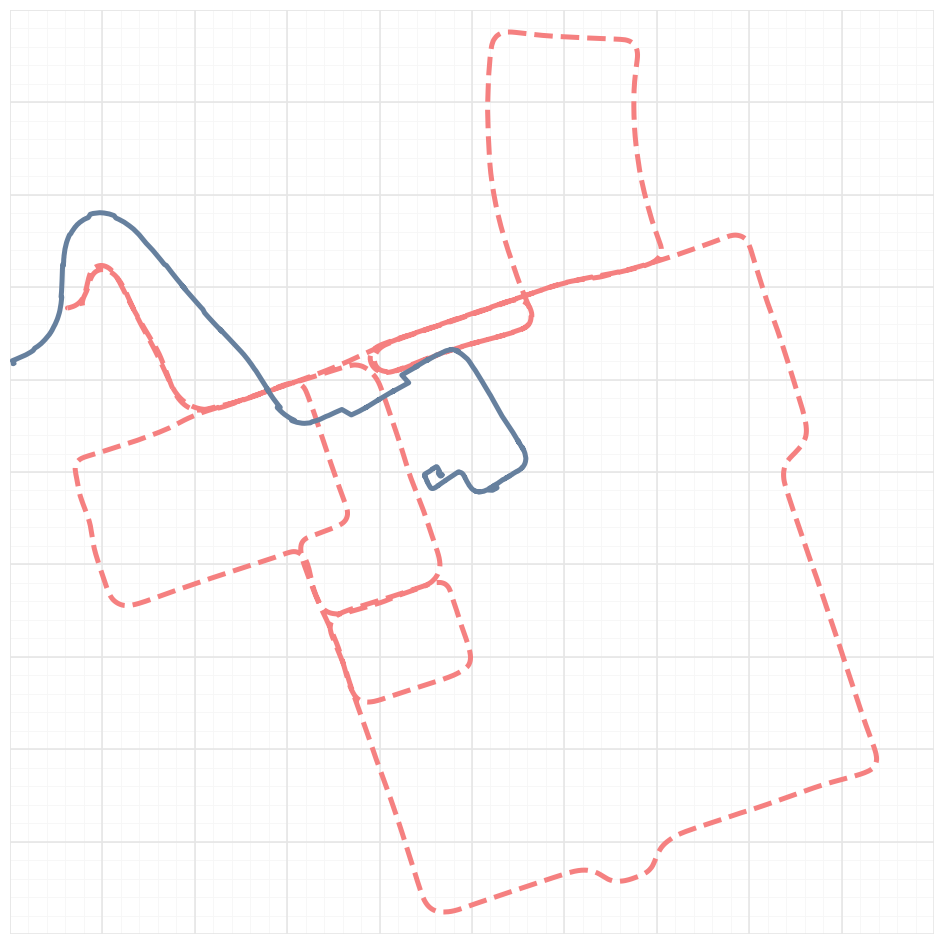} & \includegraphics[width=\linewidth]{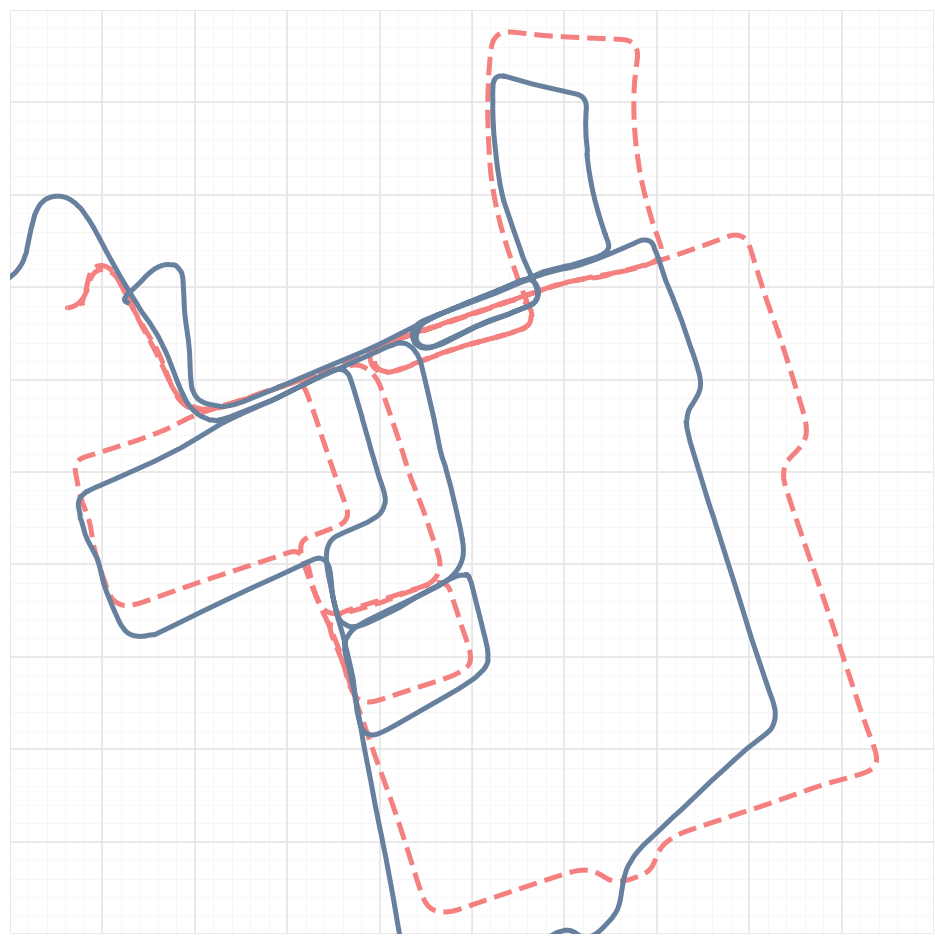} & \includegraphics[width=\linewidth]{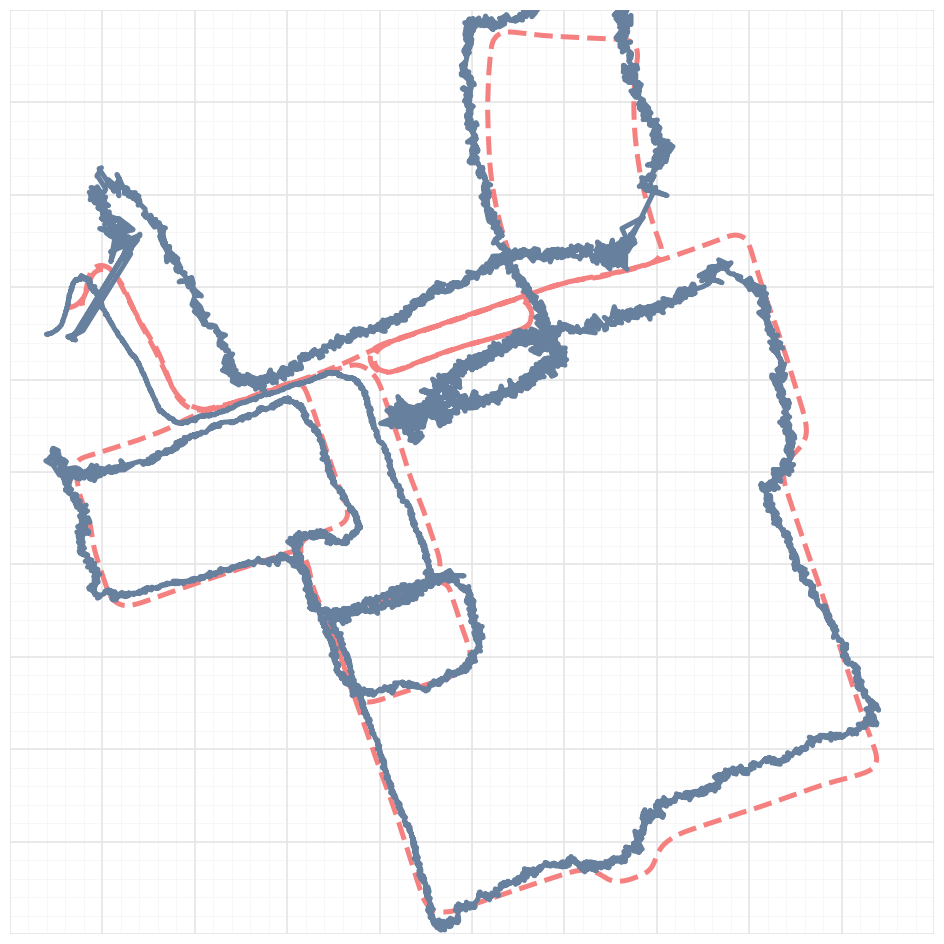} & \includegraphics[width=\linewidth]{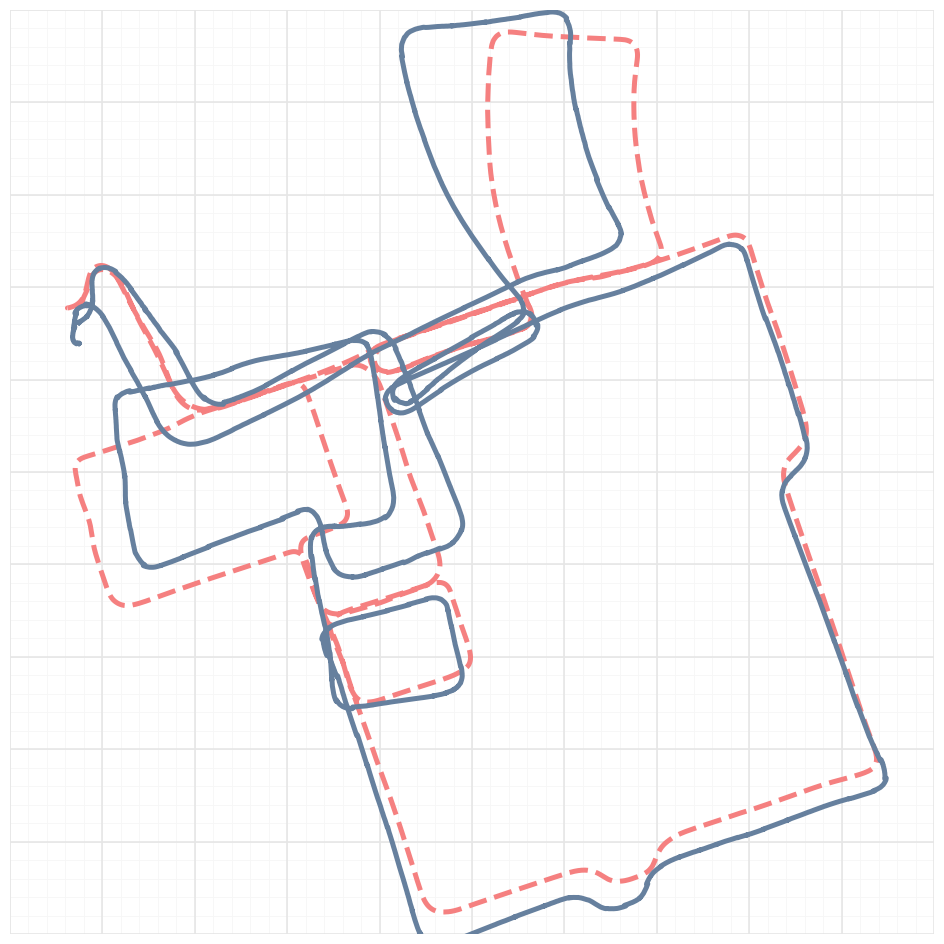} & \includegraphics[width=\linewidth]{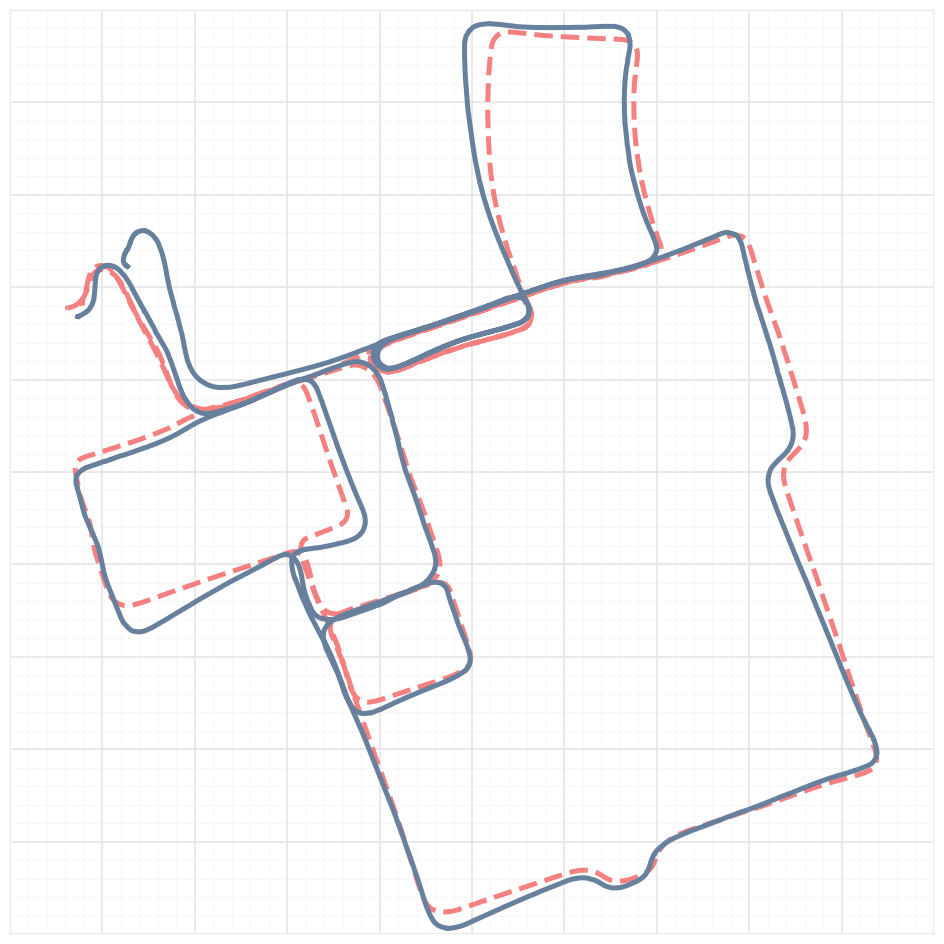} \\

\raisebox{-25pt}{\raisebox{0pt}{\rotatebox{90}{\texttt{VBR spagna\_0}}}} & \includegraphics[width=\linewidth]{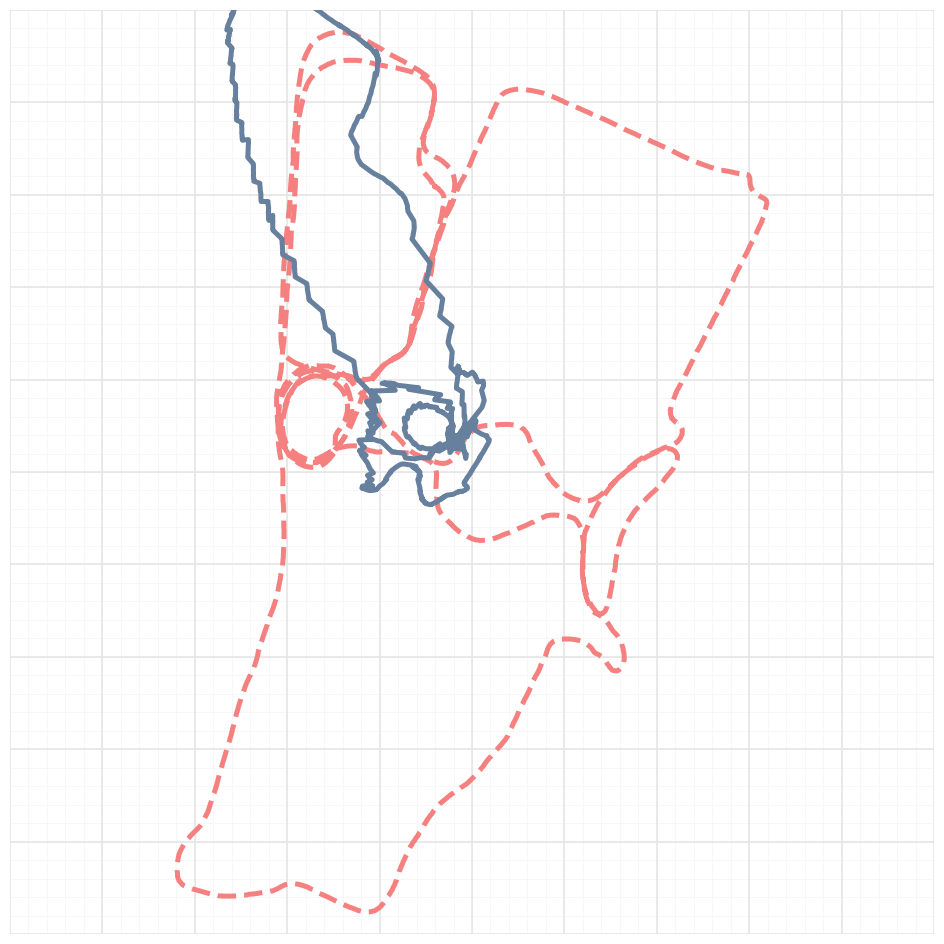} & \includegraphics[width=\linewidth]{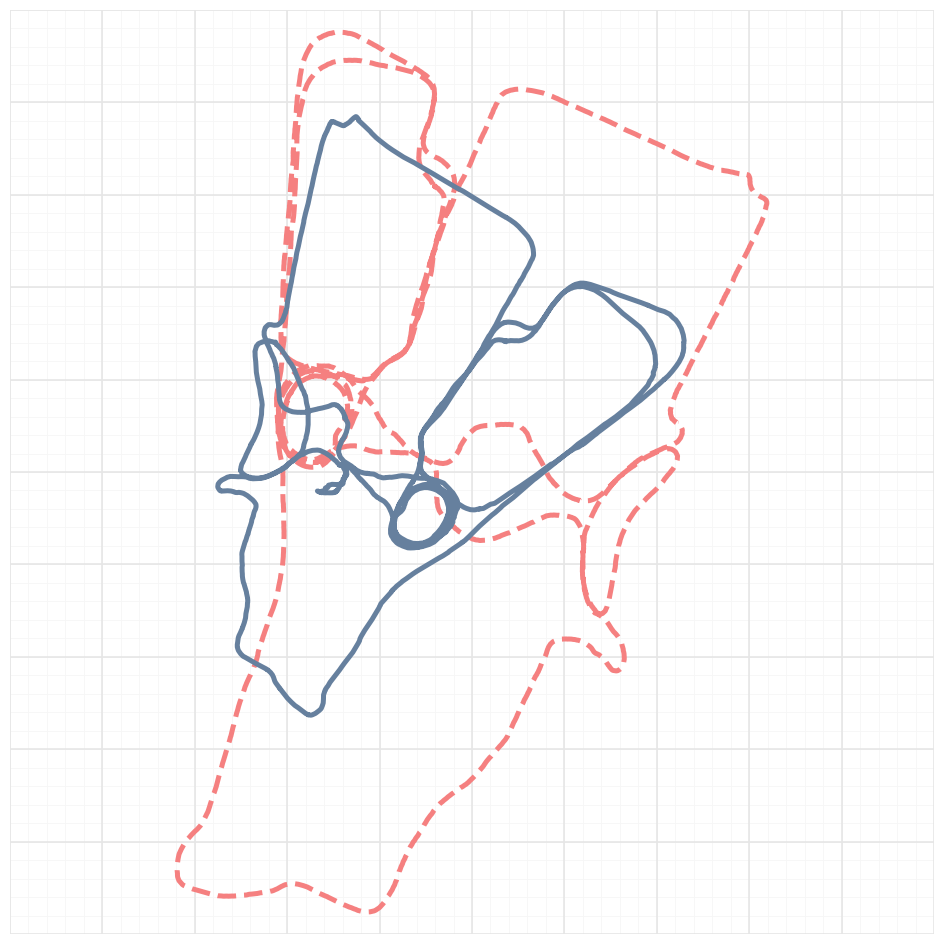} & \includegraphics[width=\linewidth]{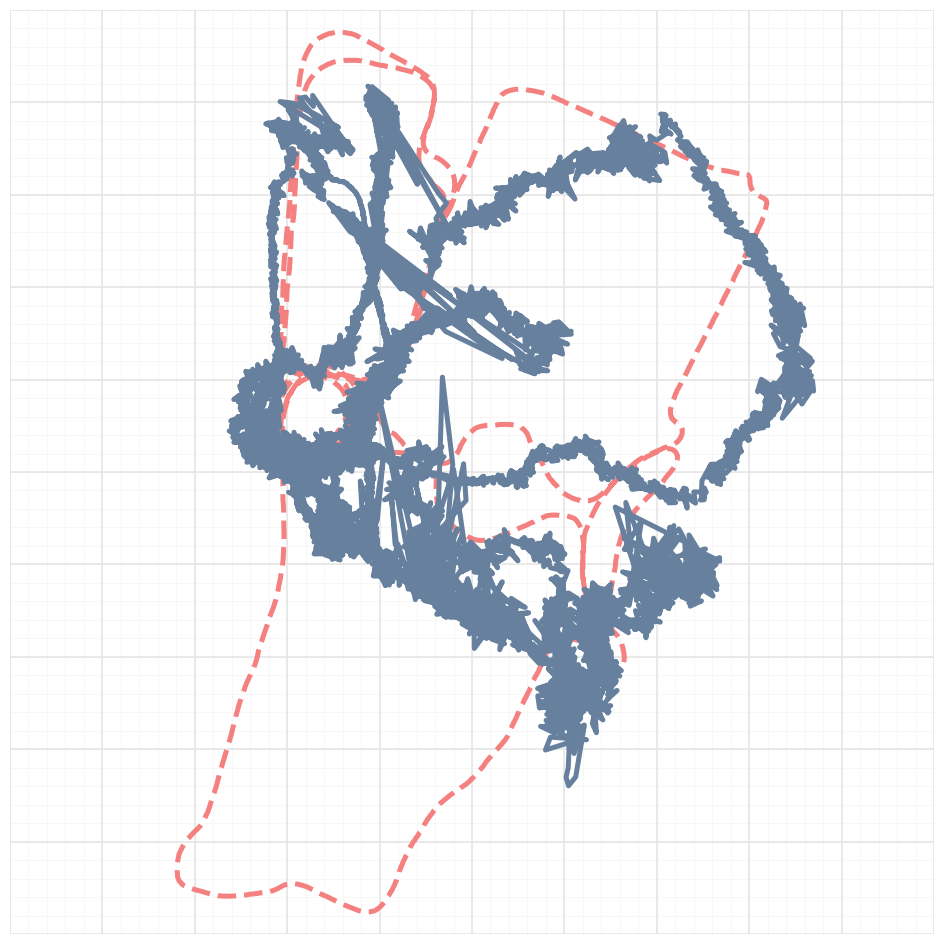} & \includegraphics[width=\linewidth]{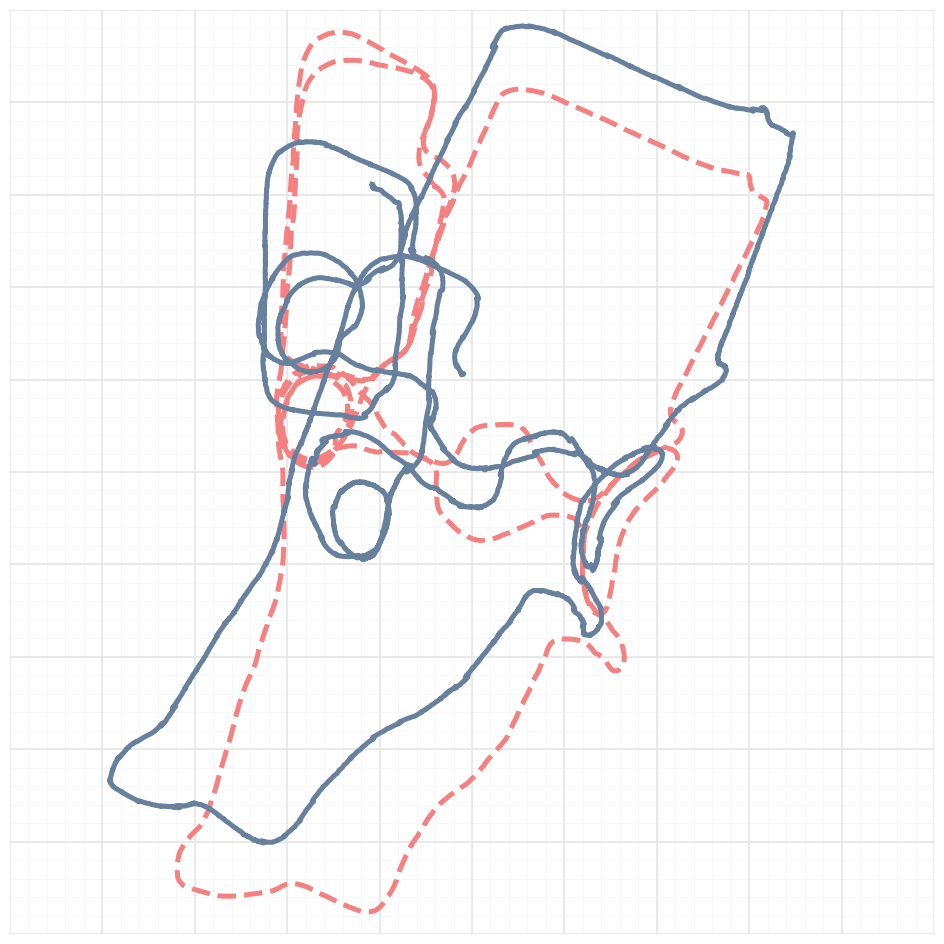} & \includegraphics[width=\linewidth]{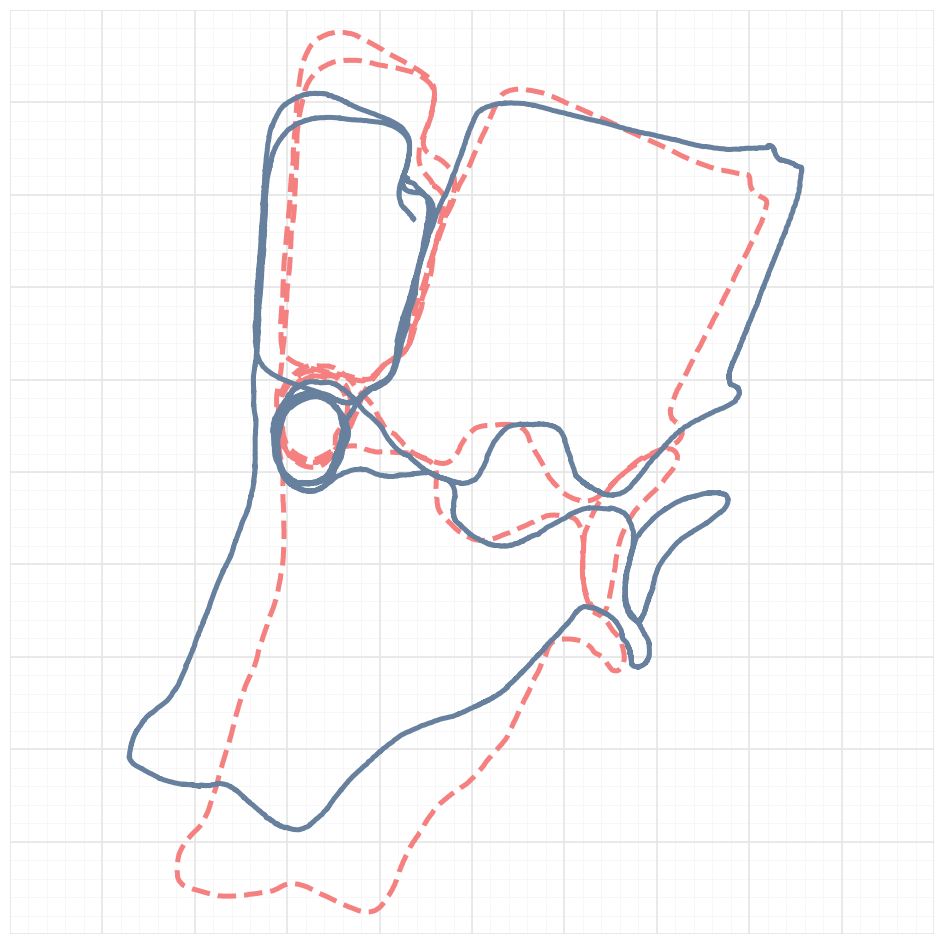} \\

\raisebox{-25pt}{\rotatebox{90}{\texttt{VBR colosseo\_0}}} & \includegraphics[width=\linewidth]{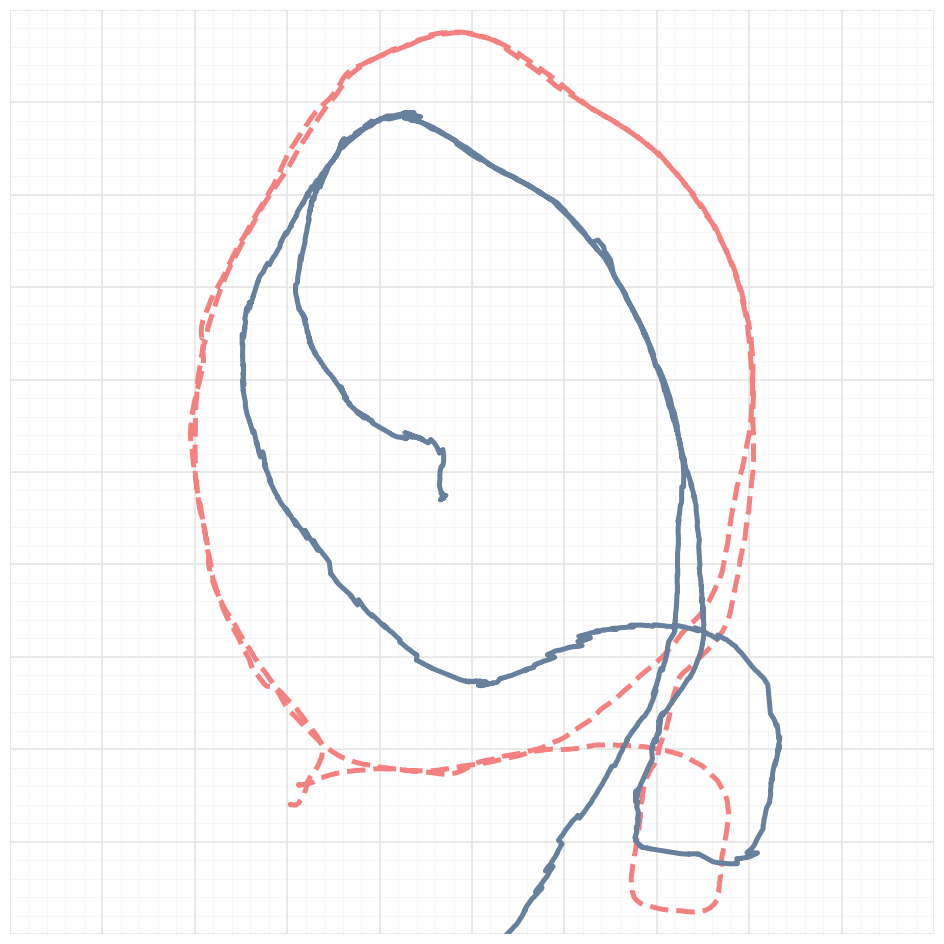} & \includegraphics[width=\linewidth]{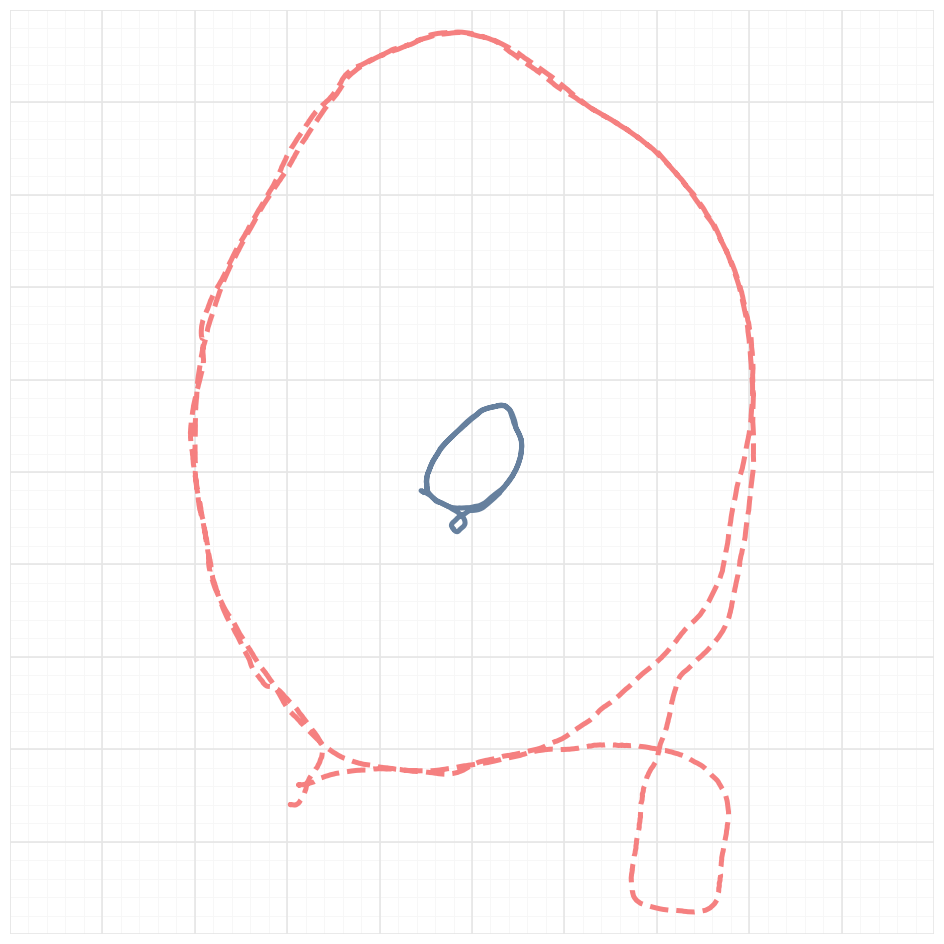} & \includegraphics[width=\linewidth]{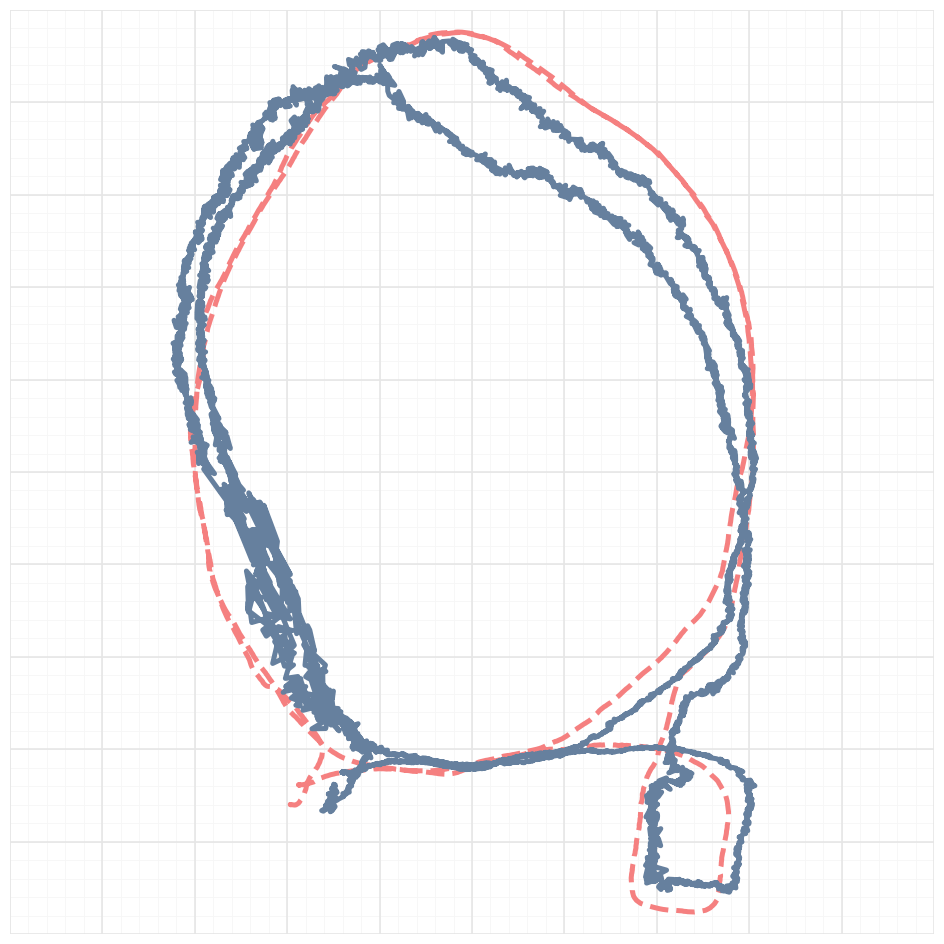} & \includegraphics[width=\linewidth]{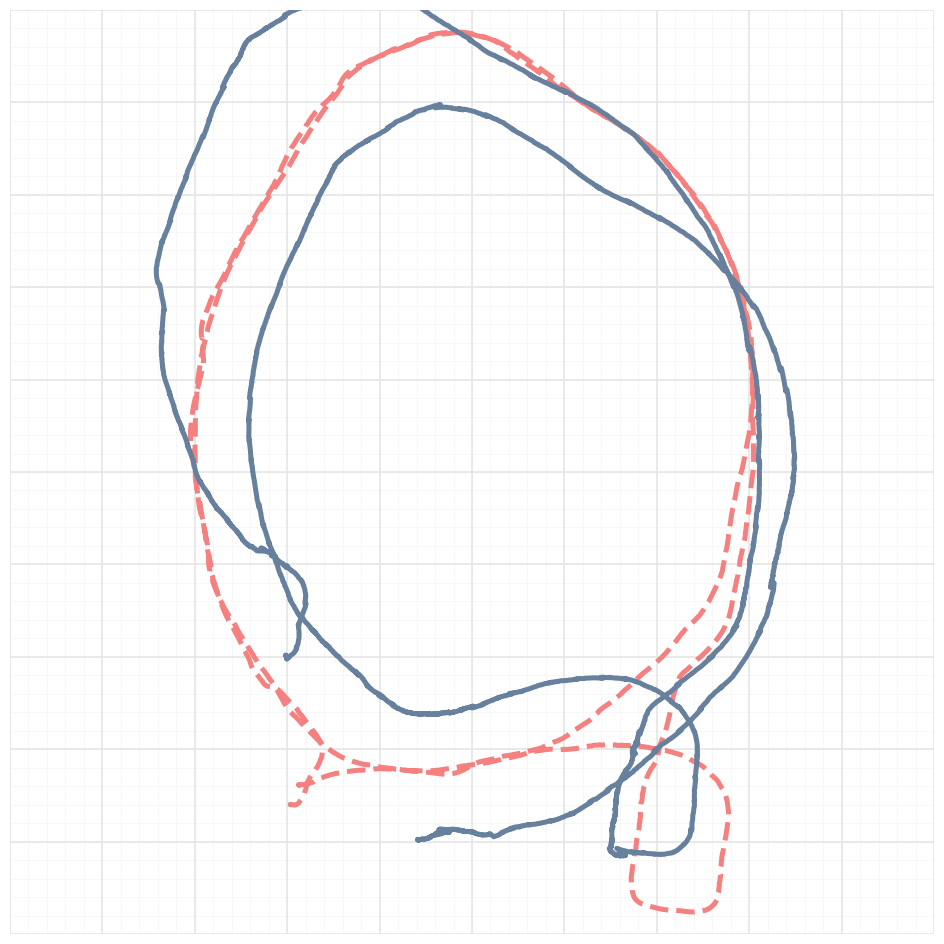} & \includegraphics[width=\linewidth]{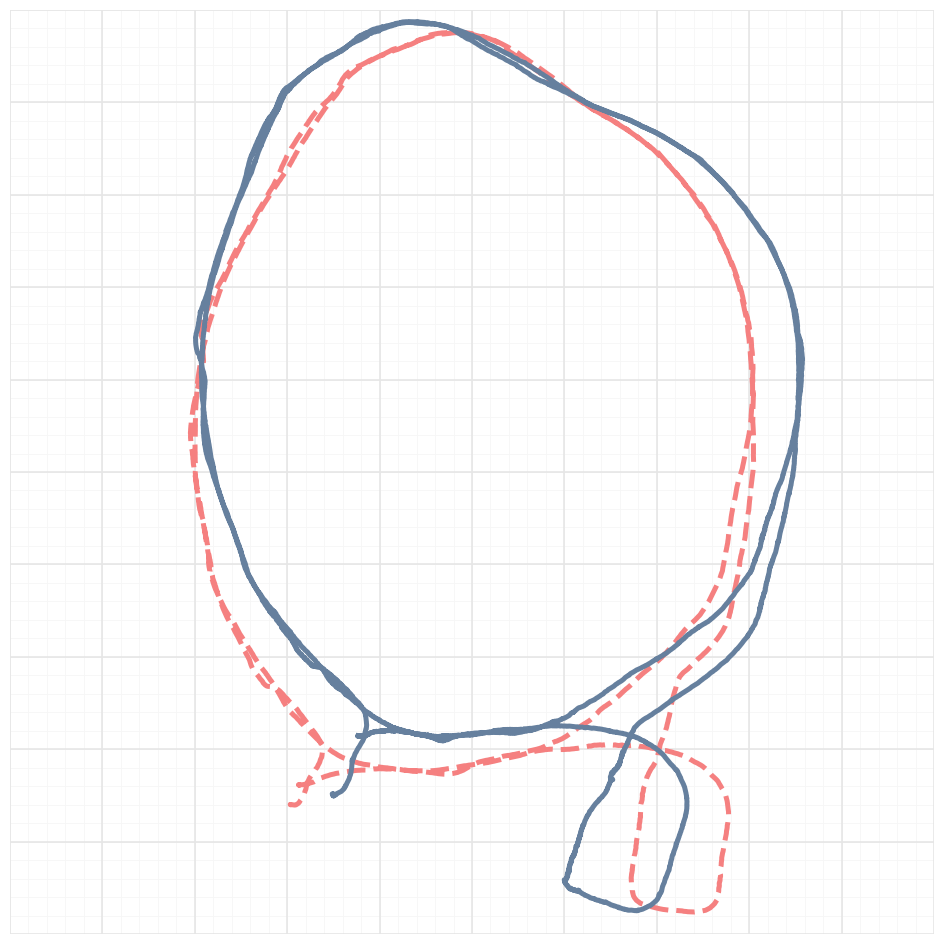} \\

\raisebox{-25pt}{\rotatebox{90}{\texttt{VBR pincio\_0}}} & \includegraphics[width=\linewidth]{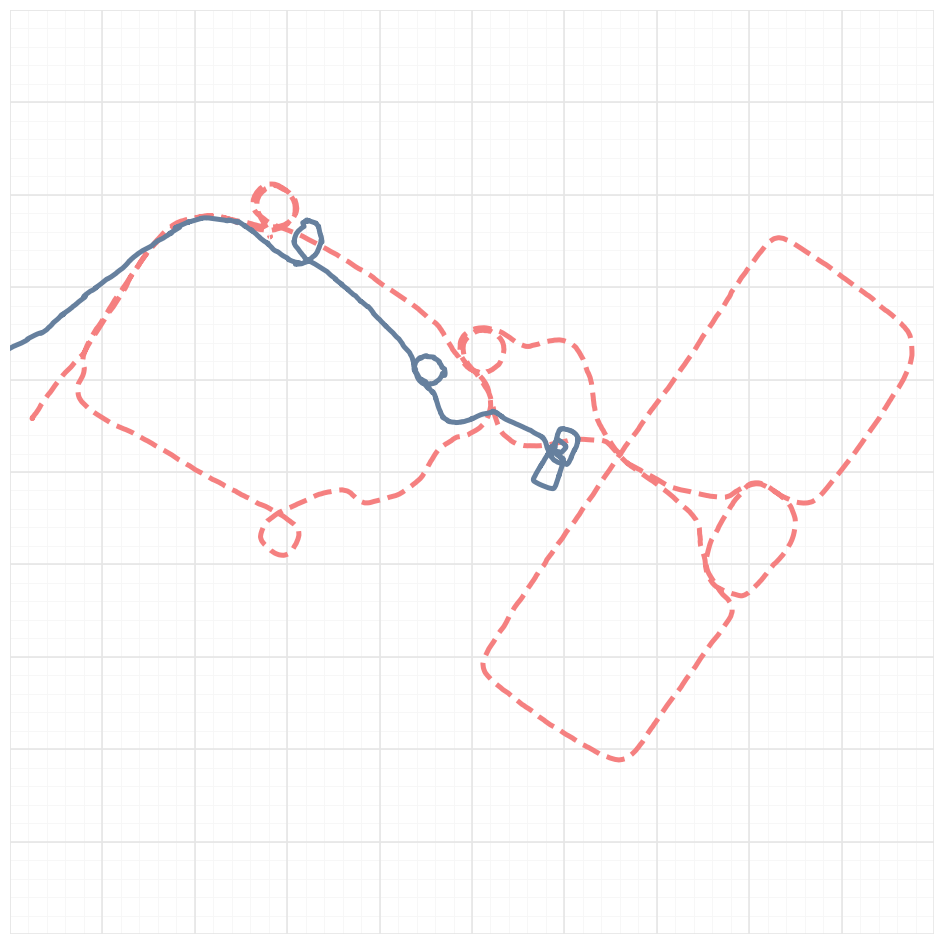} & \includegraphics[width=\linewidth]{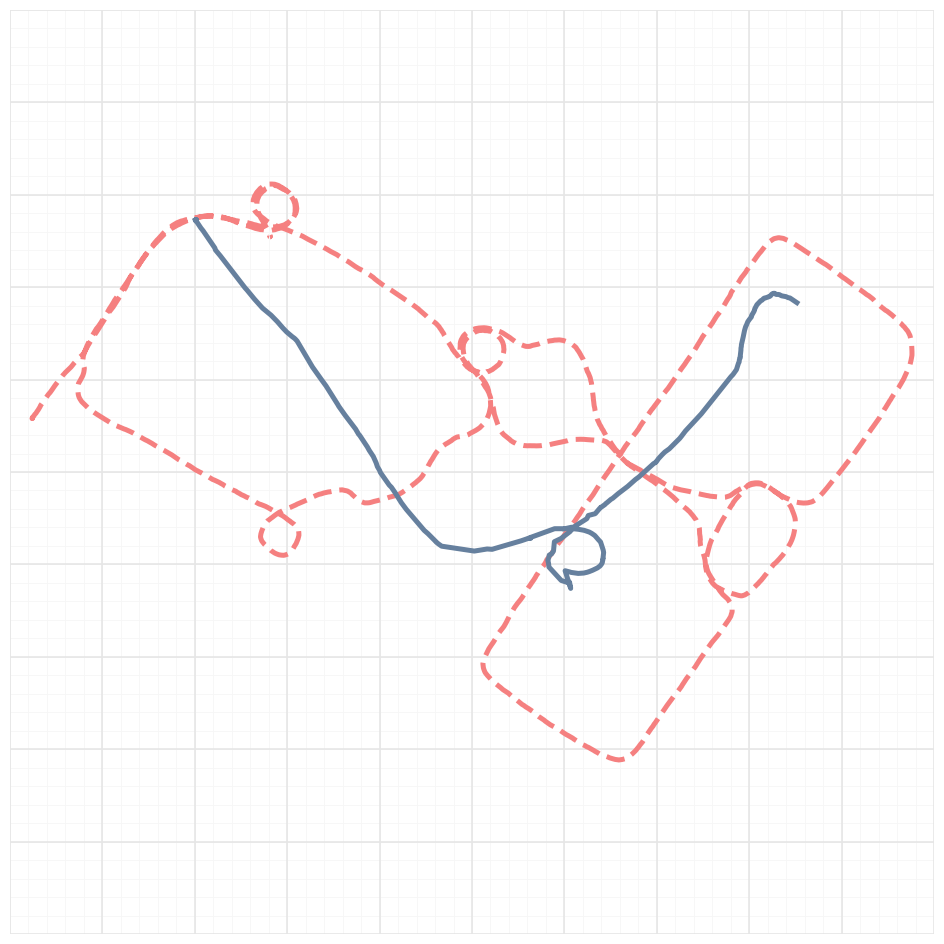} & \includegraphics[width=\linewidth]{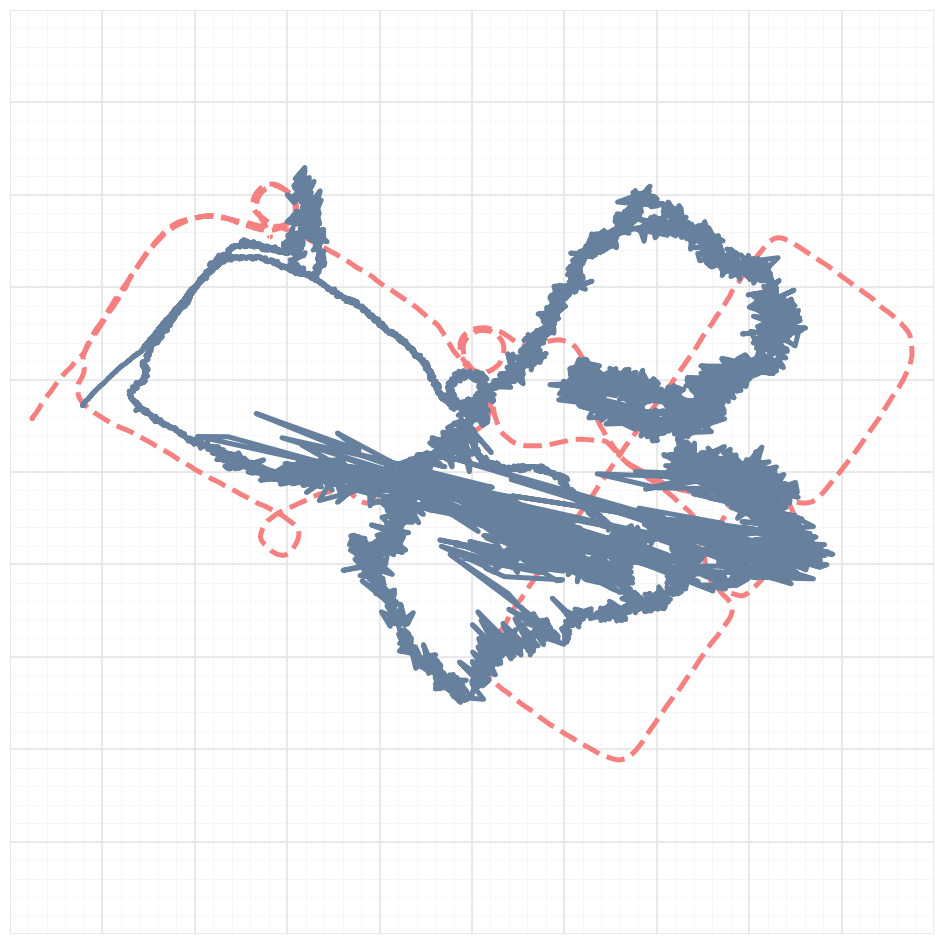} & \includegraphics[width=\linewidth]{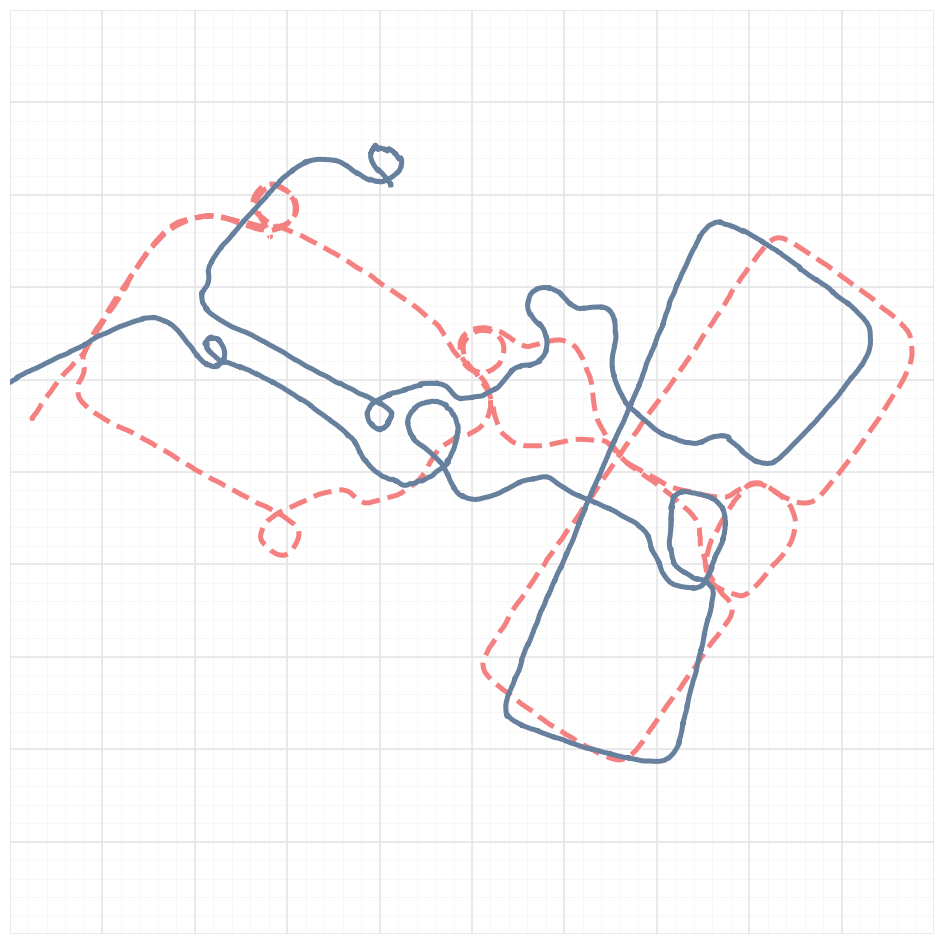} & \includegraphics[width=\linewidth]{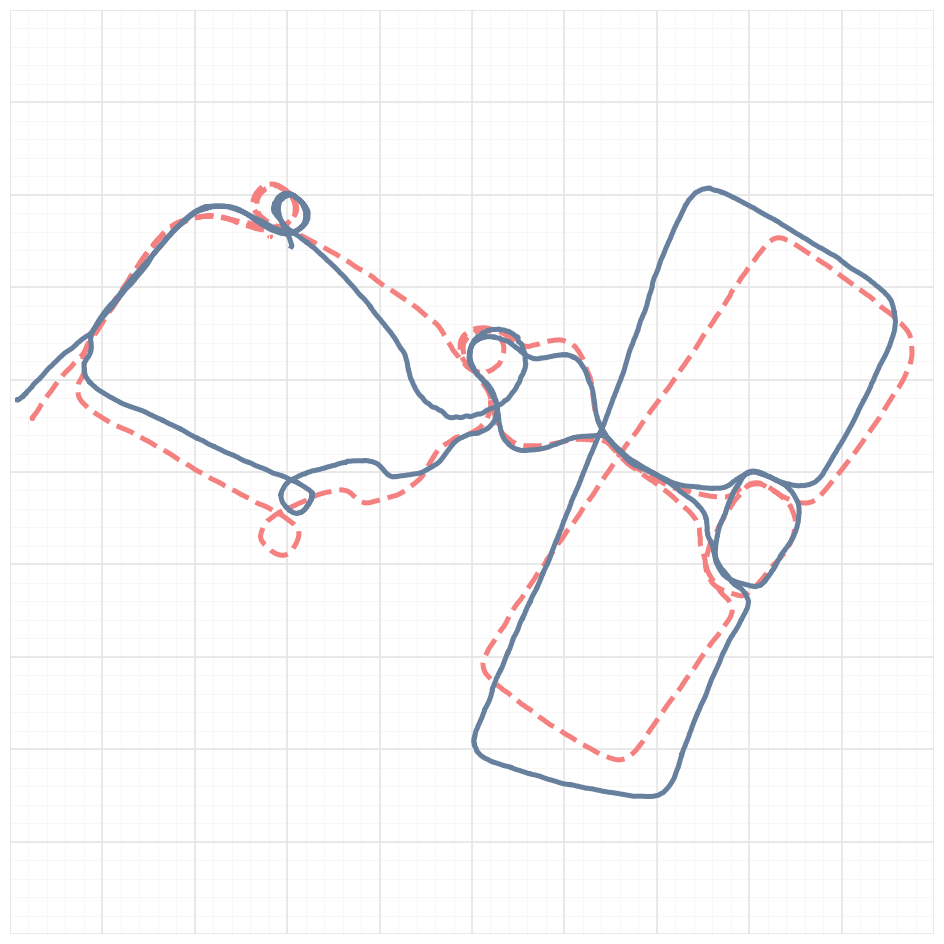} \\

\raisebox{-27pt}{\rotatebox{90}{\texttt{DROID-W down\_2}}} & \includegraphics[width=\linewidth]{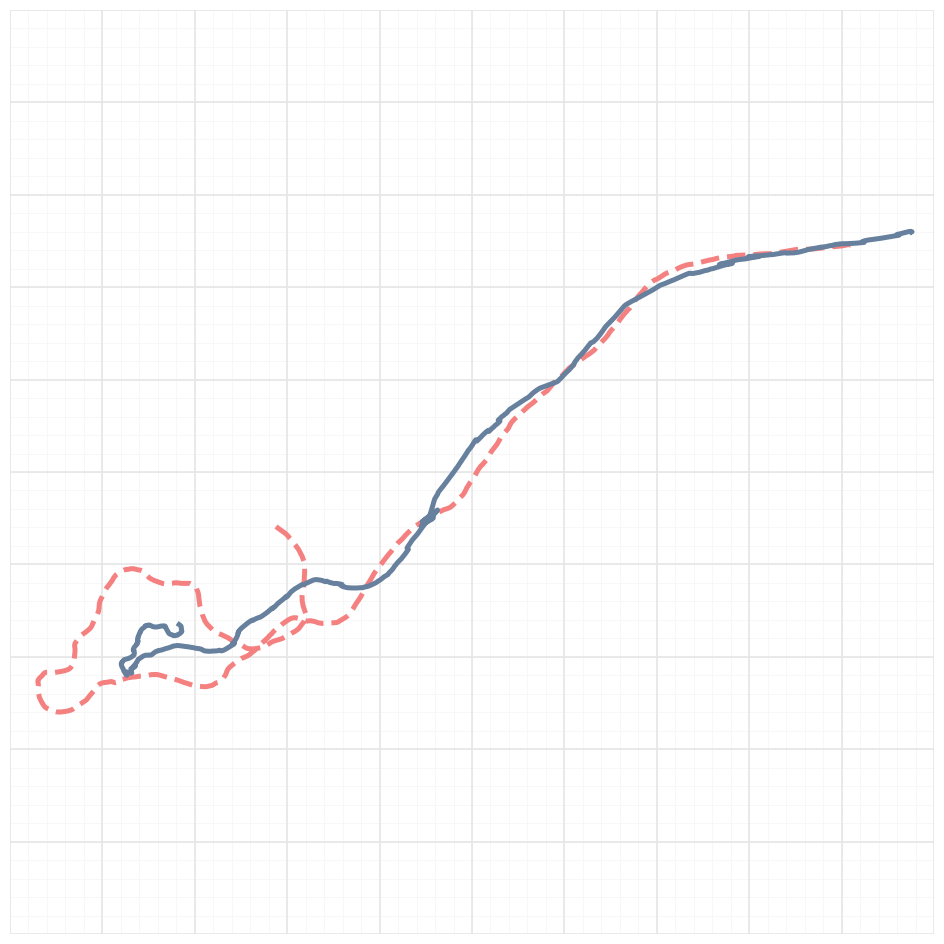} & \includegraphics[width=\linewidth]{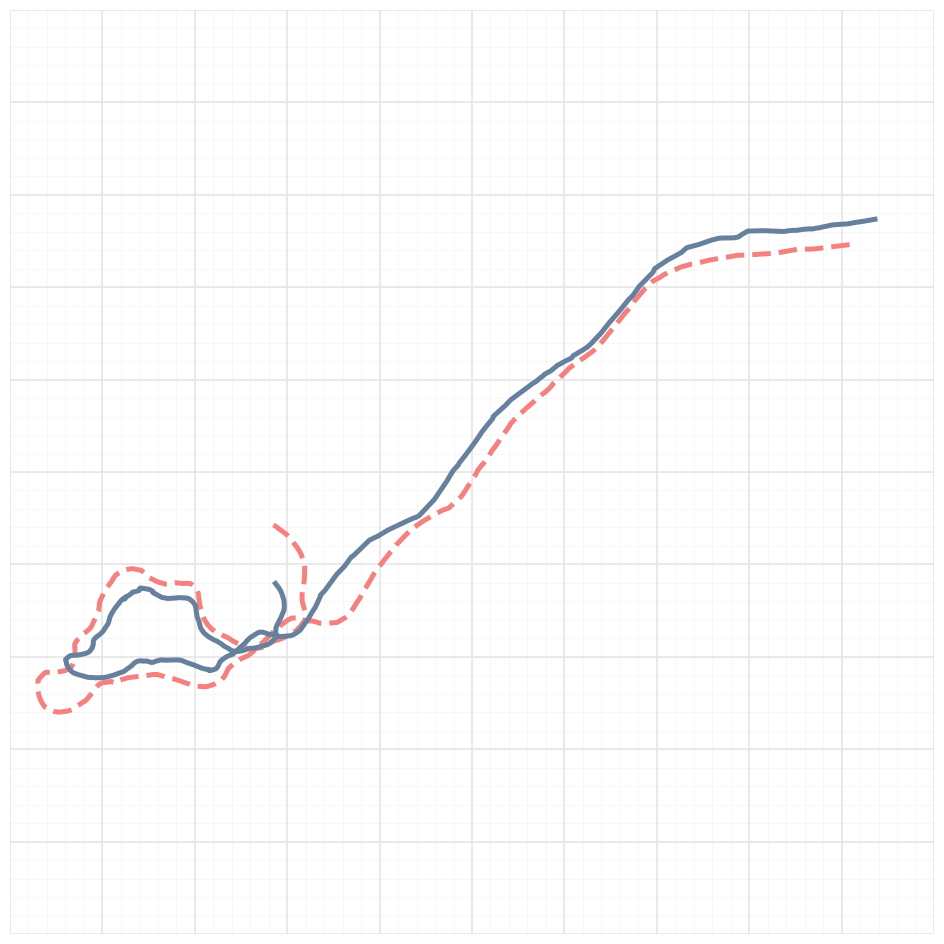} & \includegraphics[width=\linewidth]{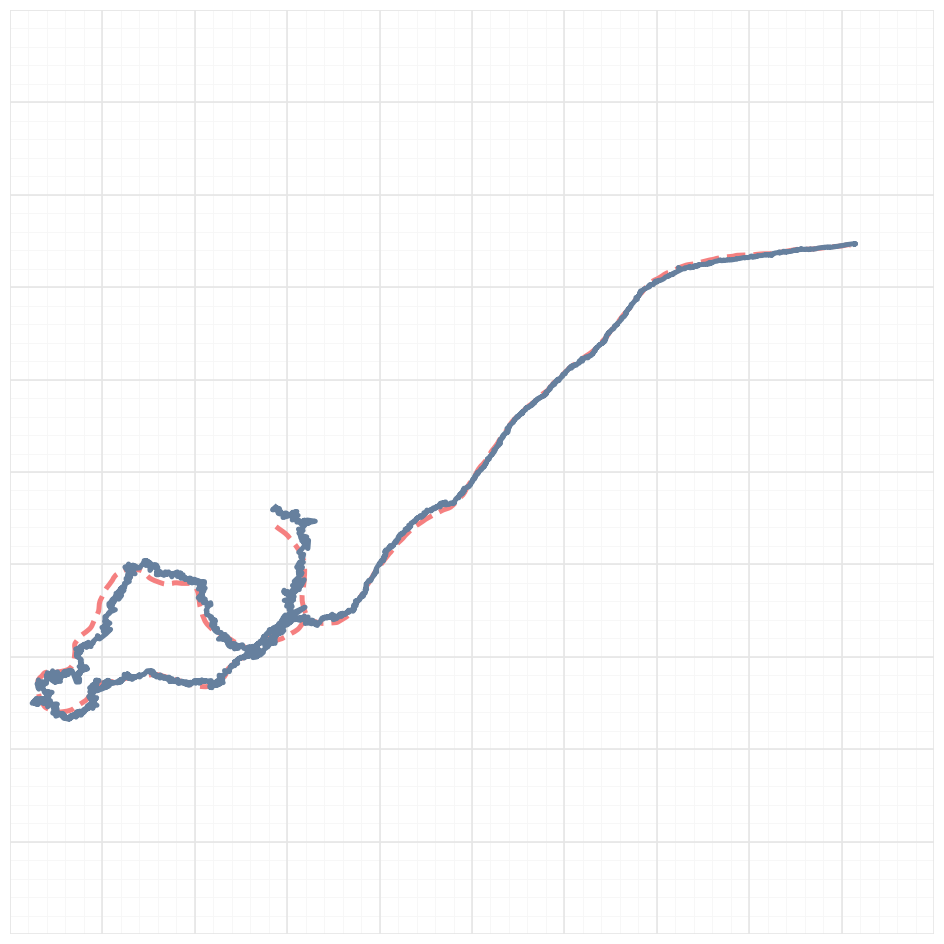} & \includegraphics[width=\linewidth]{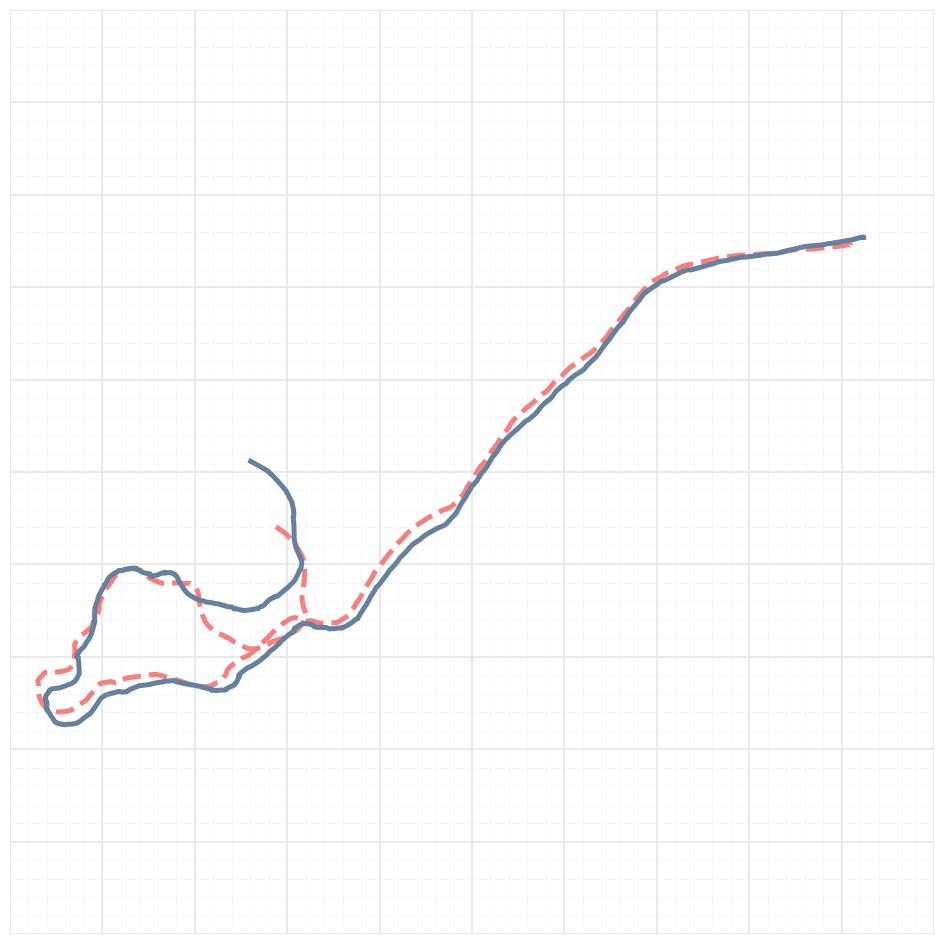} & \includegraphics[width=\linewidth]{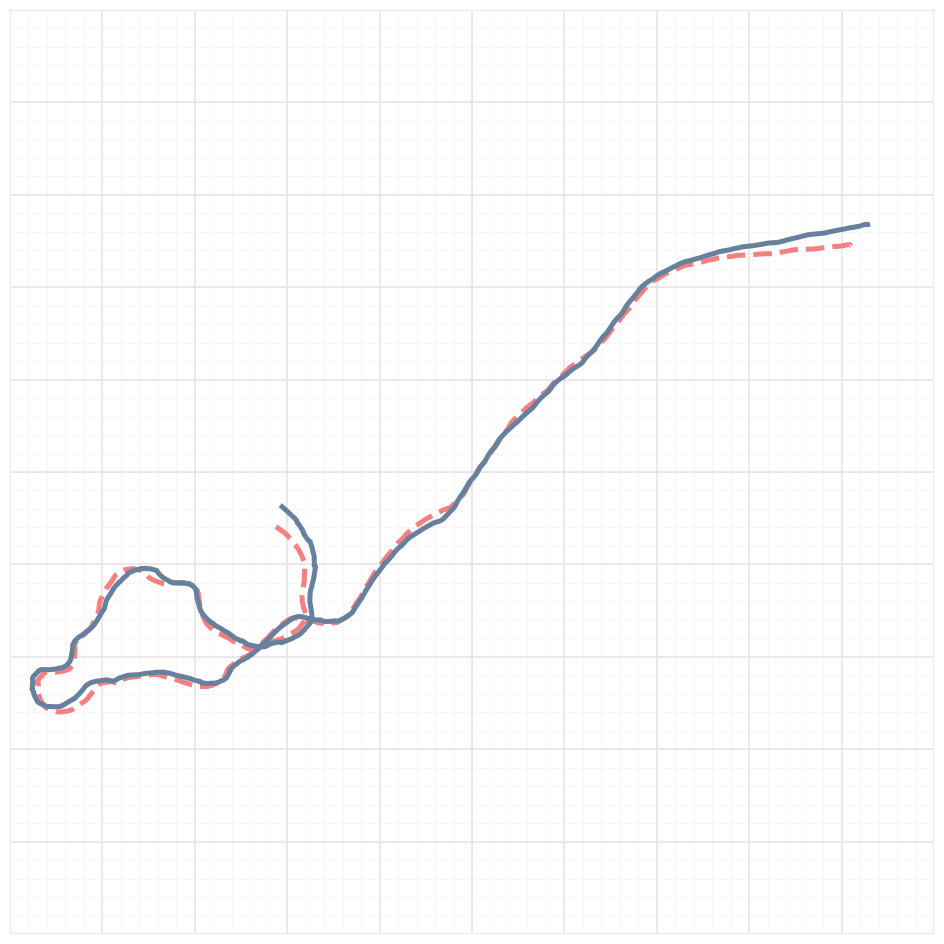} \\

\raisebox{-25pt}{\rotatebox{90}{\texttt{DROID-W down\_6}}} & \includegraphics[width=\linewidth]{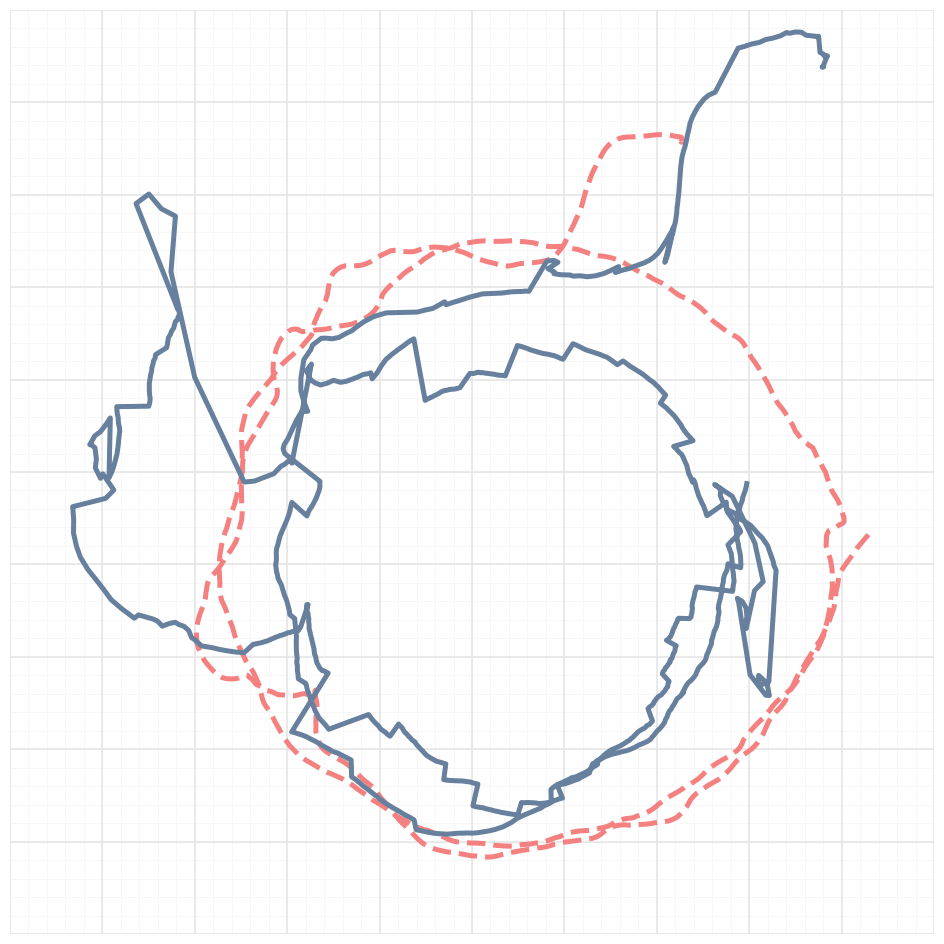} & \includegraphics[width=\linewidth]{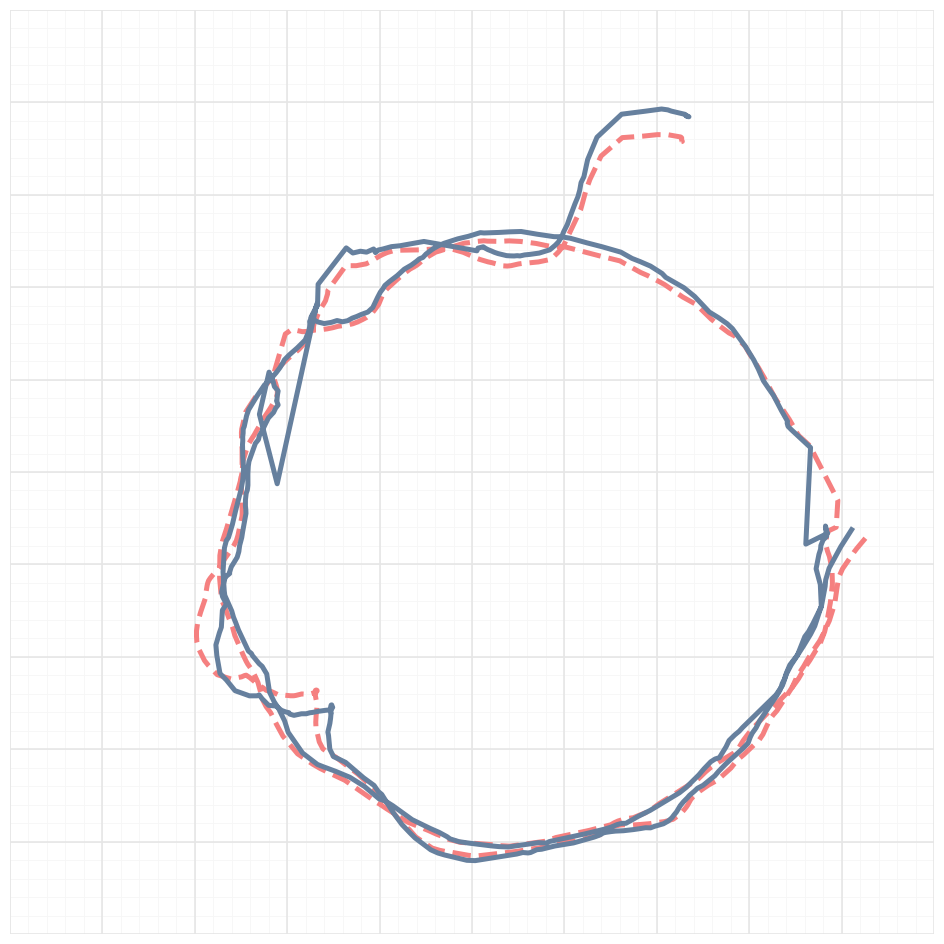} & \includegraphics[width=\linewidth]{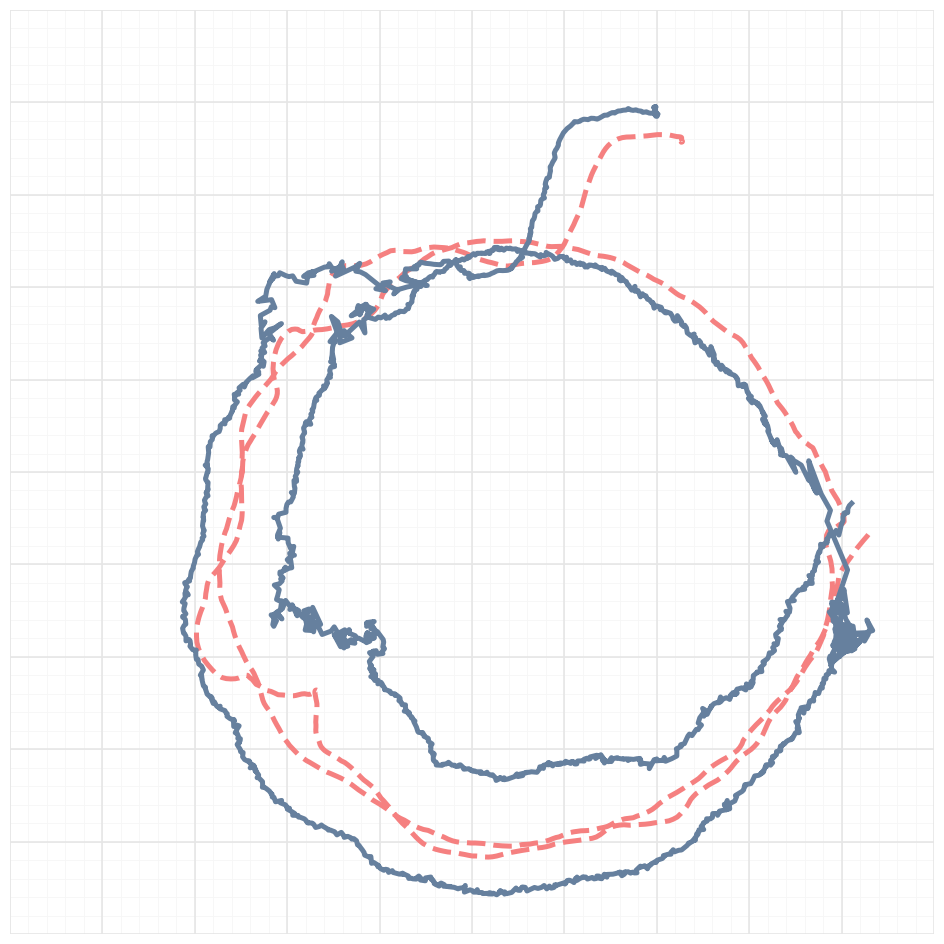} & \includegraphics[width=\linewidth]{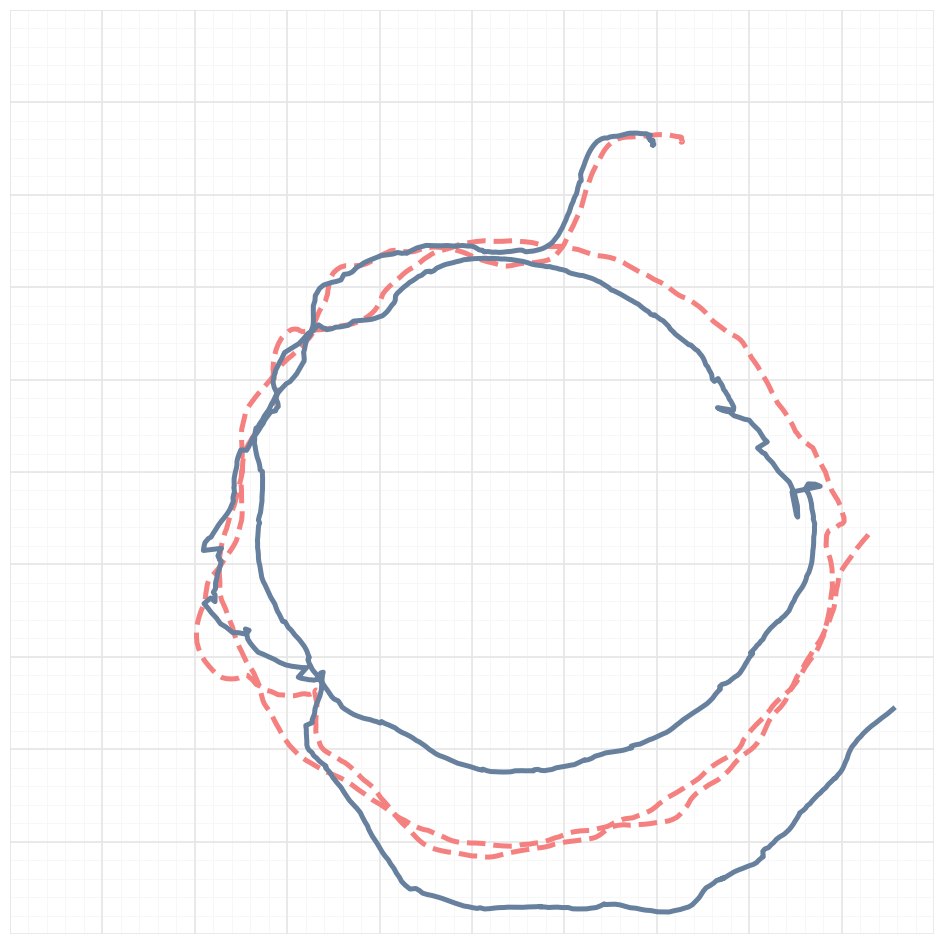} & \includegraphics[width=\linewidth]{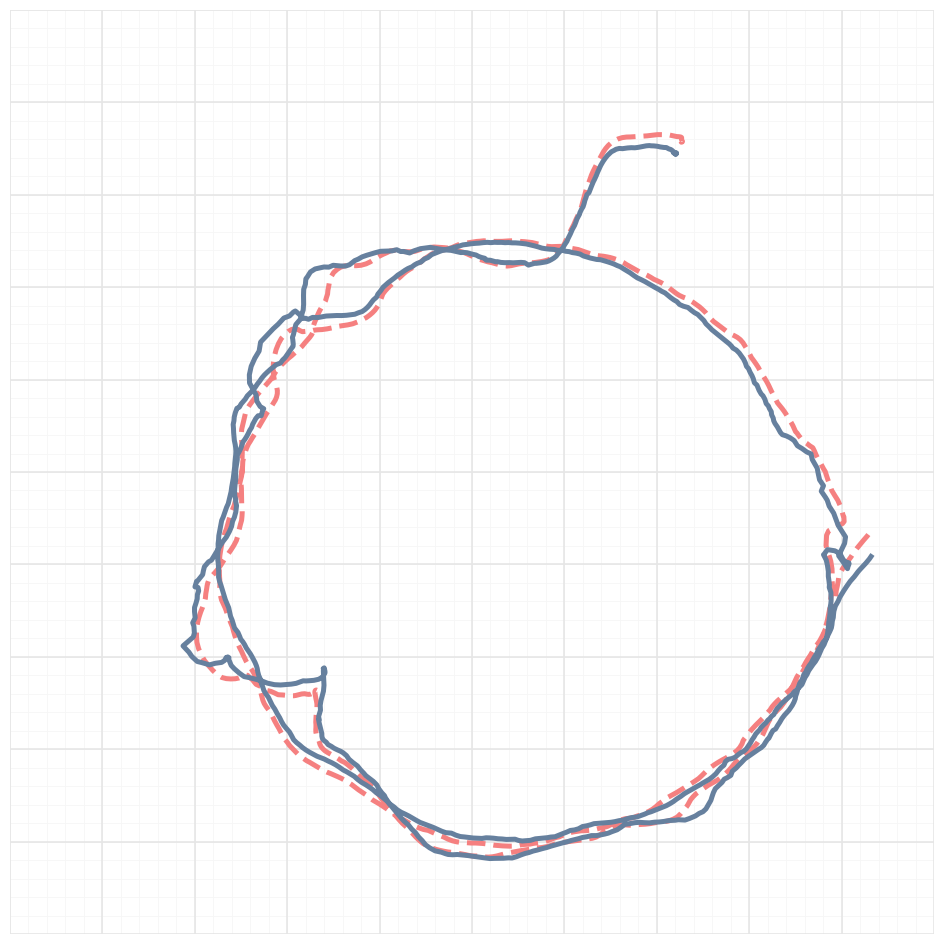} \\

\raisebox{-22pt}{\rotatebox{90}{\texttt{Oxford bl-01}}} & \includegraphics[width=\linewidth]{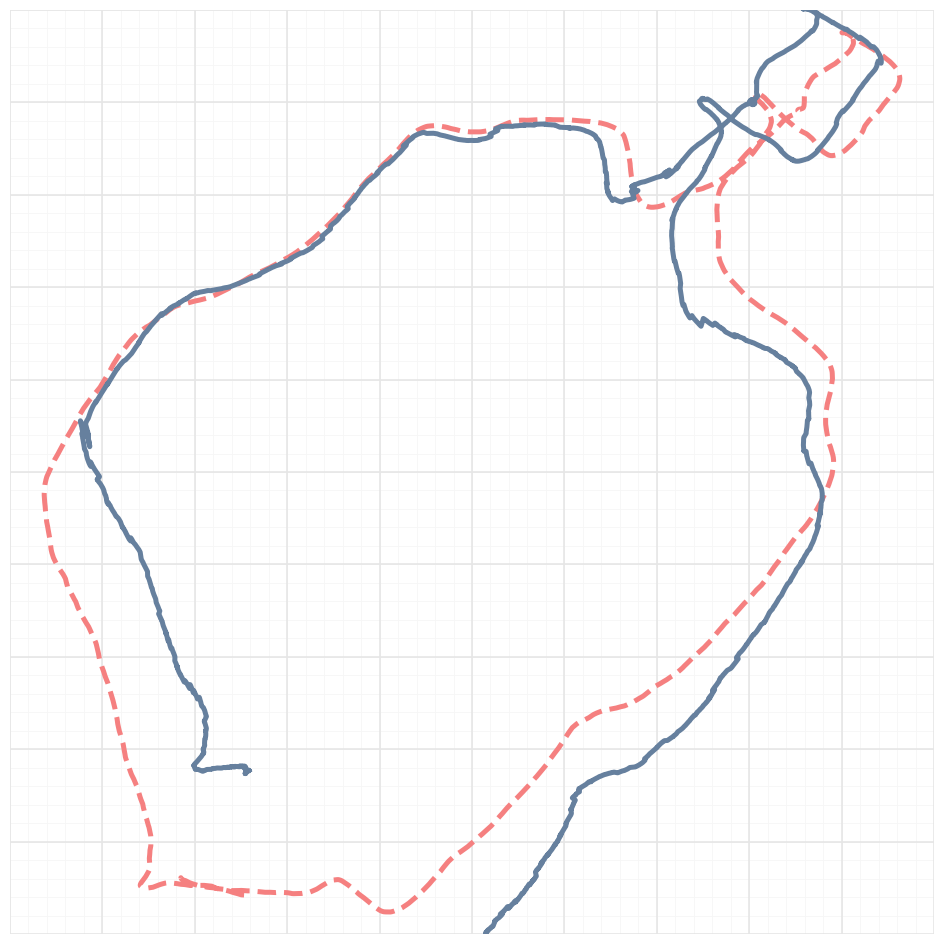} & \includegraphics[width=\linewidth]{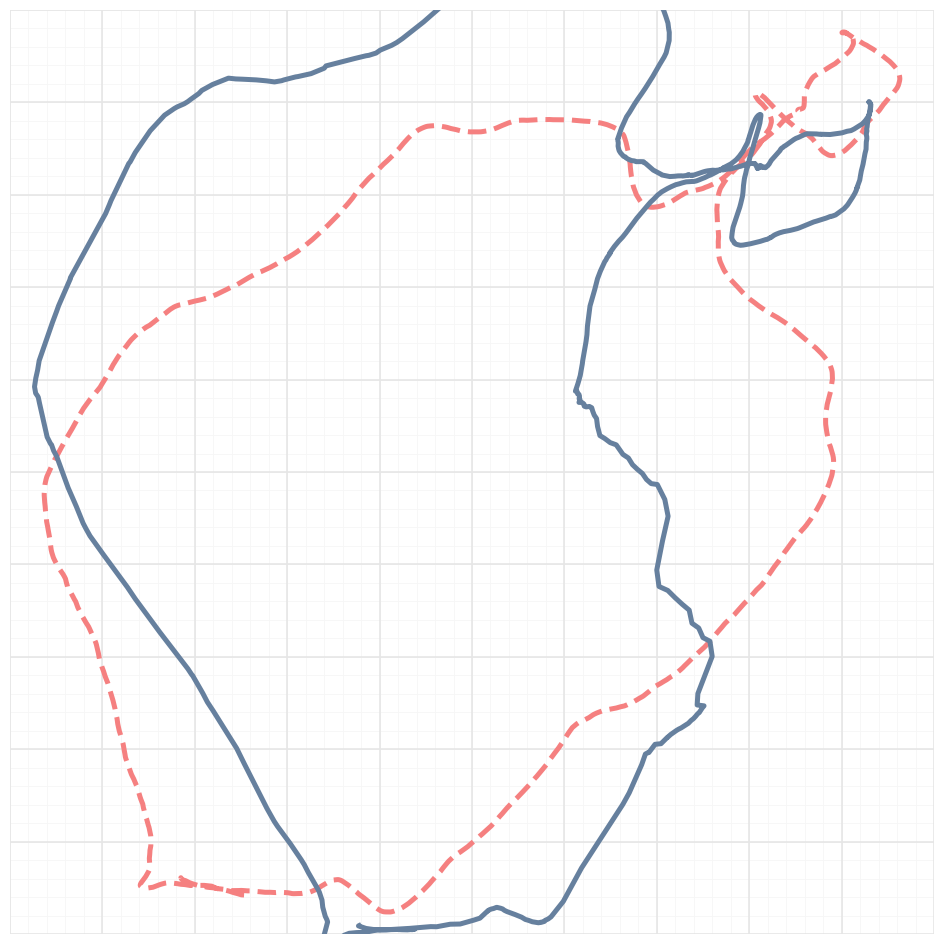} & \includegraphics[width=\linewidth]{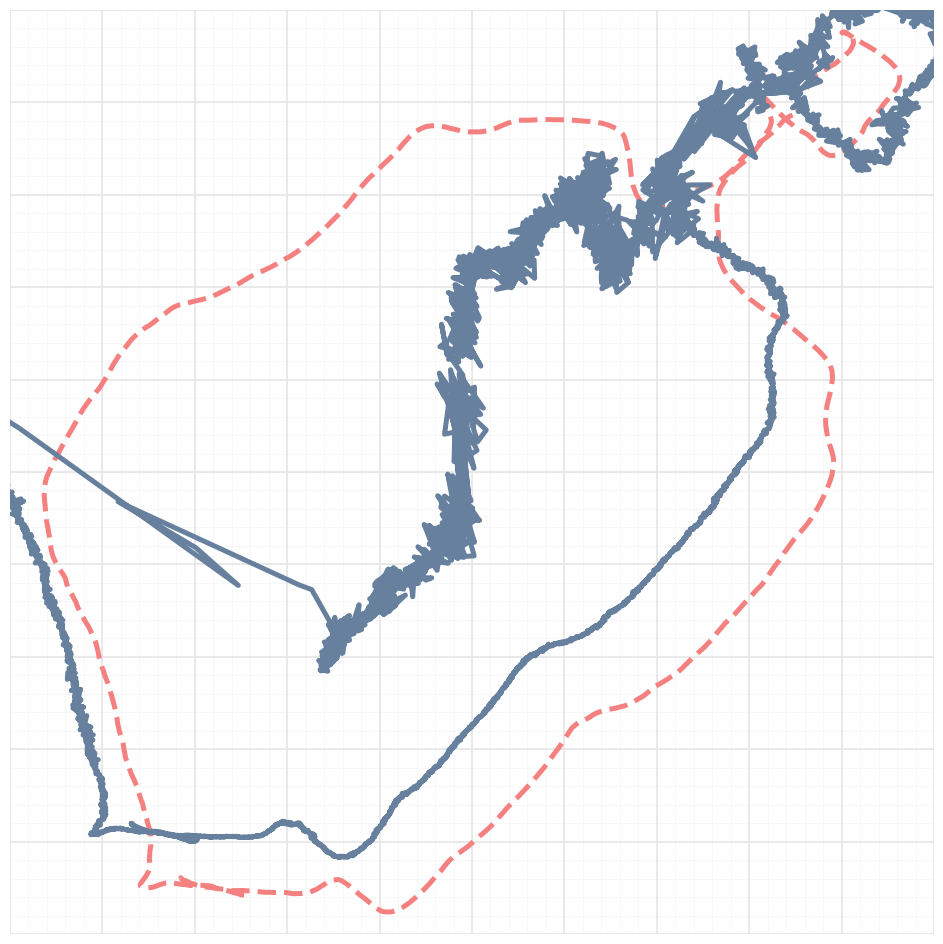} & \includegraphics[width=\linewidth]{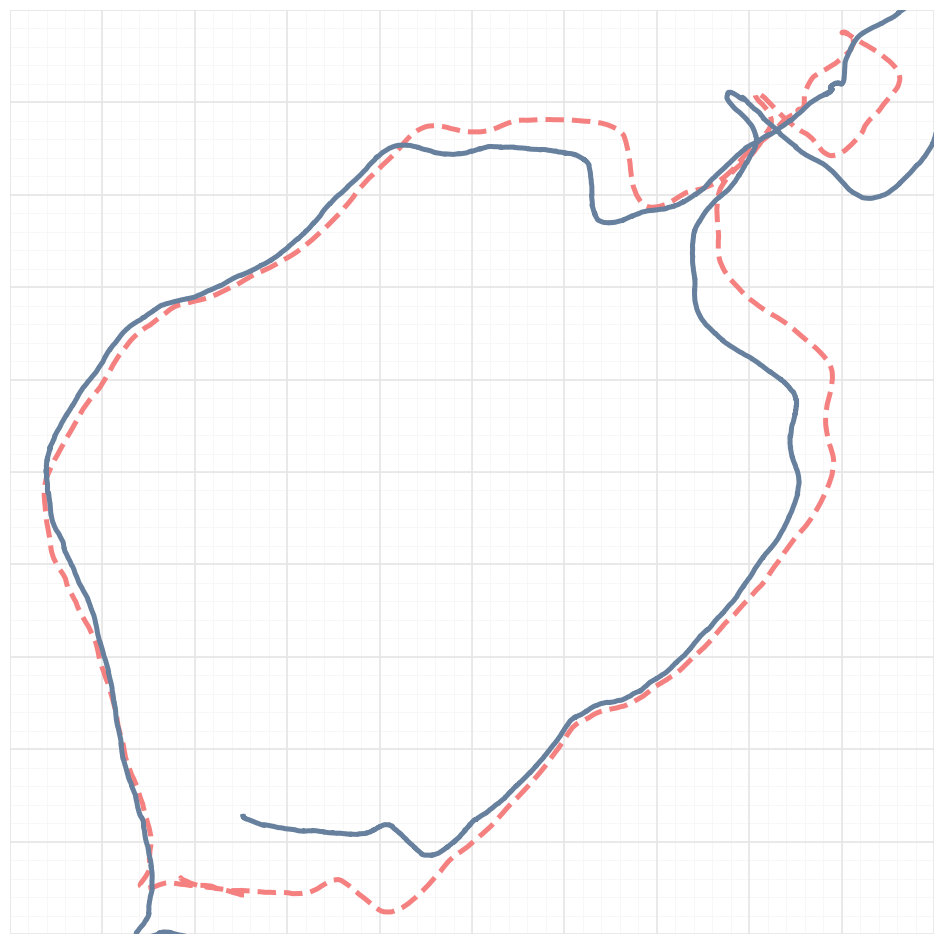} & \includegraphics[width=\linewidth]{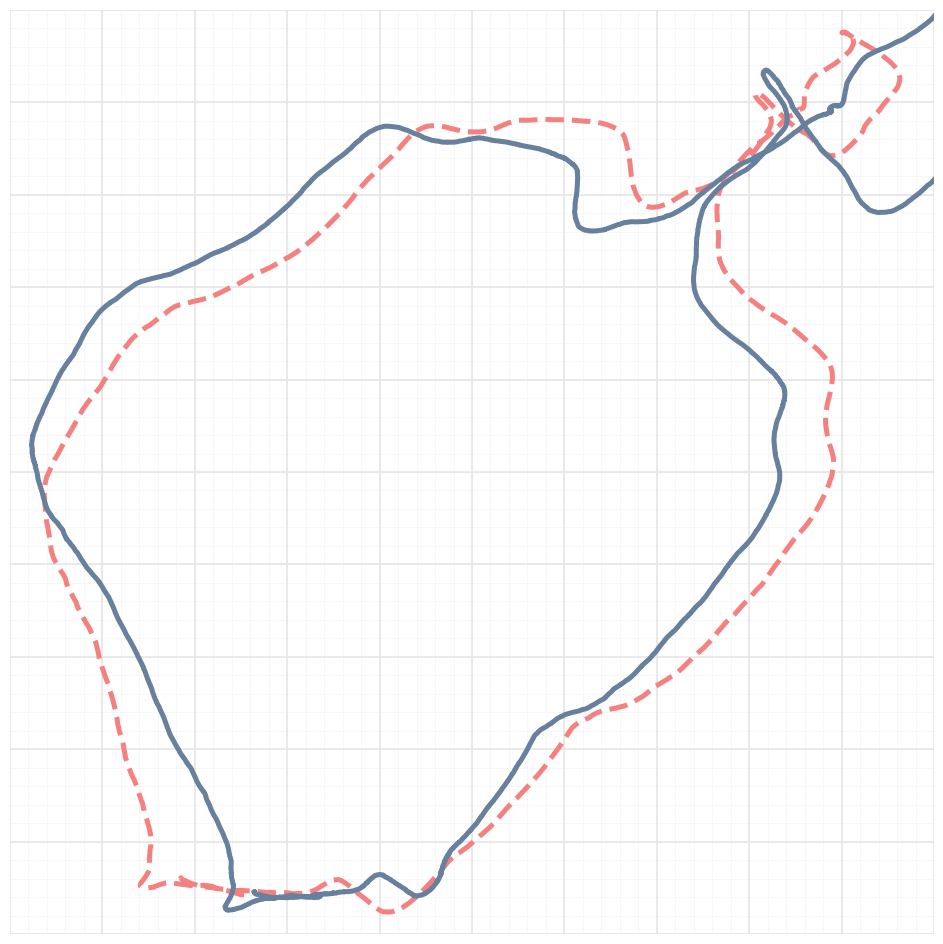} \\

\end{tabular}
\caption{\textbf{Additional trajectory comparison.} Groundtruth trajectories (\textcolor{coral}{coral}, dashed) are overlaid with the estimates from each method (\textcolor{grayblue}{gray-blue}, solid) after Sim(3) Umeyama alignment~\citep{umeyama1991least}. Our approach robustly yields the most accurate trajectories across all sequences.}
\label{fig:trajs_supp}
\end{figure}



\end{document}